\documentclass{article}

\PassOptionsToPackage{numbers, compress}{natbib}

\usepackage[final]{neurips_2025}

\usepackage[utf8]{inputenc} 
\usepackage[T1]{fontenc}    
\usepackage{url}            
\usepackage{booktabs}       
\usepackage{amsfonts}       
\usepackage{nicefrac}       
\usepackage{microtype}      
\usepackage{xcolor}         
\usepackage{bm}
\usepackage{graphicx}
\usepackage{amsmath}
\usepackage{algorithm}
\usepackage{algpseudocode}
\usepackage{multirow}
\usepackage{arydshln}
\usepackage{adjustbox}   
\usepackage{marvosym}
\usepackage{url}
\usepackage{algorithm}
\usepackage{algpseudocode}
\usepackage{enumitem}
\usepackage{booktabs}
\usepackage{multicol, multirow}
\usepackage{graphicx}
\usepackage{xspace}
\usepackage{times}
\usepackage[hyperindex,breaklinks,colorlinks,citecolor=darkblue]{hyperref} 
\usepackage{wrapfig}
\usepackage{textcomp} 
\usepackage{hyperref}
\usepackage{caption}
\usepackage{subcaption}
\usepackage{todonotes}
\usepackage{listings}
\usepackage[utf8]{inputenc} 
\usepackage{amsmath, amssymb, amsthm}
\usepackage{svg}
\usepackage{tabularx}
\usepackage{wrapfig}
\usepackage{microtype}

\definecolor{rliableblue}{HTML}{77AADD}
\definecolor{darkblue}{rgb}{0.0, 0.0, 0.55}
\definecolor{bblue}{rgb}{0,0.45,0.74}
\definecolor{myred}{rgb}{0.85,0.33,0.1}
\definecolor{deepgreen}{RGB}{0, 150, 0}
\definecolor{website_color}{rgb}{0.9333333333333333, 0.10980392156862745, 0.592156862745098} 
\newcommand{\RN}[1]{%
  \textup{\uppercase\expandafter{\romannumeral#1}}%
}
\usepackage{caption}
\usepackage{footnote}
\makesavenoteenv{table}
\usepackage{minitoc}
\usepackage{tocloft}
\usepackage{colortbl}
\usepackage{alltt}
\usepackage{geometry}
\usepackage{multicol}
\usepackage{listings}     
\usepackage{graphicx}    
\usepackage{subcaption}  
\usepackage{csquotes}
\usepackage{lstautogobble}  
\usepackage{color}          
\usepackage{zi4}            
\usepackage[most]{tcolorbox}
\makeatletter 
\usepackage{enumitem}
\setlist[itemize]{leftmargin=*} 
\setlist[itemize]{topsep=0pt, partopsep=0pt, itemsep=0pt, parsep=0pt} 
\makeatother 
\title{Unlocking Multimodal Mathematical Reasoning \\via Process Reward Model}

\definecolor{backblue}{RGB}{210, 230, 250}
\newcommand{\high}{\cellcolor{backblue}}
\definecolor{backred}{RGB}{255, 223, 223}
\newcommand{\highr}{\cellcolor{backred}}
\definecolor{backgreen}{RGB}{220,244,229}
\newcommand{\highg}{\cellcolor{backgreen}}
\definecolor{back_deepblue}{RGB}{180, 210, 240}

\makeatletter
  \newcommand\figcaption{\def\@captype{figure}\caption}
  \newcommand\tabcaption{\def\@captype{table}\caption}
\makeatother
\definecolor{back_deepred}{RGB}{255, 200, 200}

\definecolor{mypink}{HTML}{FF69B4}
\newcommand{\difftext}[1]{\small\textcolor{mypink}{($\uparrow$#1)}}
\definecolor{back_deepgreen}{RGB}{190, 230, 210}

\renewcommand\footnotemark{}
\author{Ruilin Luo$^{12*}$ \quad  Zhuofan Zheng$^{2*}$ \quad Yifan Wang$^{1}$
\quad \textbf{Xinzhe Ni}$^{1}$ \\
\quad \textbf{Zicheng Lin}$^{1}$ \quad \textbf{Songtao Jiang}$^{3}$ \quad \textbf{Yiyao Yu}$^{1}$ \quad \textbf{Chufan Shi}$^{1}$\\
\textbf{Lei Wang}$^{4}$ \quad \textbf{Ruihang Chu}$^{1\dagger}$ \quad \textbf{Jin Zeng}$^{2\dagger}$ \quad \textbf{Yujiu Yang}$^{1}$ \thanks{*\ Equal contribution. Work done during Ruilin's internship at ByteDance. $\dagger$ Corresponding author.\\\text{\ \ \ \ \ \ \ \ \ }ruihangchu@gmail.com, zengjin@bytedance.com}\\\\
$^1$Tsinghua University \quad  $^2$ByteDance \\ $^3$Zhejiang University \quad $^4$ Ping An Technology (Shenzhen) Co., Ltd.
}

\begin{document}

\maketitle

\begin{abstract}
  Process Reward Models (PRMs) have shown promise in enhancing the mathematical reasoning capabilities of Large Language Models (LLMs) through Test-Time Scaling~(TTS). However, their integration into multimodal reasoning remains largely unexplored. In this work, we take the first step toward unlocking the potential of PRMs in multimodal mathematical reasoning. We identify three key challenges: (\romannumeral1) the scarcity of high-quality reasoning data constrains the capabilities of foundation Multimodal Large Language Models~(MLLMs), which imposes further limitations on the upper bounds of TTS and reinforcement learning (RL); (\romannumeral2) a lack of automated methods for process labeling within multimodal contexts persists; (\romannumeral3) 
  the employment of process rewards in unimodal RL faces issues like reward hacking, which may extend to multimodal scenarios.
  To address these issues, we introduce \textbf{URSA}, a three-stage \textbf{U}nfolding multimodal p\textbf{R}ocess-\textbf{S}upervision \textbf{A}ided training framework. We first construct MMathCoT-1M, a high-quality large-scale multimodal Chain-of-Thought (CoT) reasoning dataset, to build a \textit{stronger math reasoning foundation MLLM}, URSA-8B. Subsequently, we go through an automatic process to synthesize process supervision data, which emphasizes both logical correctness and perceptual consistency. We introduce DualMath-1.1M to facilitate the training of URSA-8B-RM. Finally, we propose \textbf{P}rocess-\textbf{S}upervised \textbf{G}roup-\textbf{R}elative-\textbf{P}olicy-\textbf{O}ptimization~(\textbf{PS-GRPO}), pioneering a \textit{multimodal PRM-aided online RL method} that outperforms vanilla GRPO. 
  With PS-GRPO application, URSA-8B-PS-GRPO outperforms Gemma3-12B and GPT-4o by 8.4\% and 2.7\% on average across 6 benchmarks. Code, data and checkpoint can be found at~\href{https://github.com/URSA-MATH}{https://github.com/URSA-MATH}.
\end{abstract}

\section{Introduction}
Following the substantial progress of Large Language Models (LLMs) in math reasoning~\citep{luo2023wizardmath,yang2024qwenmath,ying2024internmath,shao2024deepseekmath,yang2024mathglm,yu2024metamath,ni2024exploring,yu2025chain}, the math reasoning capabilities of Multimodal Large Language Models (MLLMs) have increasingly garnered attention~\citep{yang2024mathglmv,DBLP:journals/corr/abs-2412-18319,zhuang2025math,xiang2024atomthink,huang25visionr1}. Previous work has typically focused on aspects such as math reasoning data curation~\citep{han24infimm,shi2024math,cai2024geogpt4v,deng2024r,DBLP:journals/corr/abs-2312-11370}, training math-intensive vision encoders~\citep{zhang2024mavis,xia24geox}, enhancing vision-language alignment~\citep{zhuang2025math,xia24charx}, or the application of post-training techniques~\citep{zhang24cot,peng2024multimath,zhang2024improve,huang25visionr1}. Given the success of Process Reward Models (PRMs) in improving LLM reasoning through methods like Test-Time Scaling~(TTS)~\citep{liu2025can,zhang2024generative} and Reinforcement Fine-Tuning~(ReFT)~\citep{zhang2024rest,liu2024diving}, the application of PRMs to multimodal reasoning remains unexplored.
\begin{figure*}[!t]
    \centering
    \includegraphics[width=0.95\linewidth]{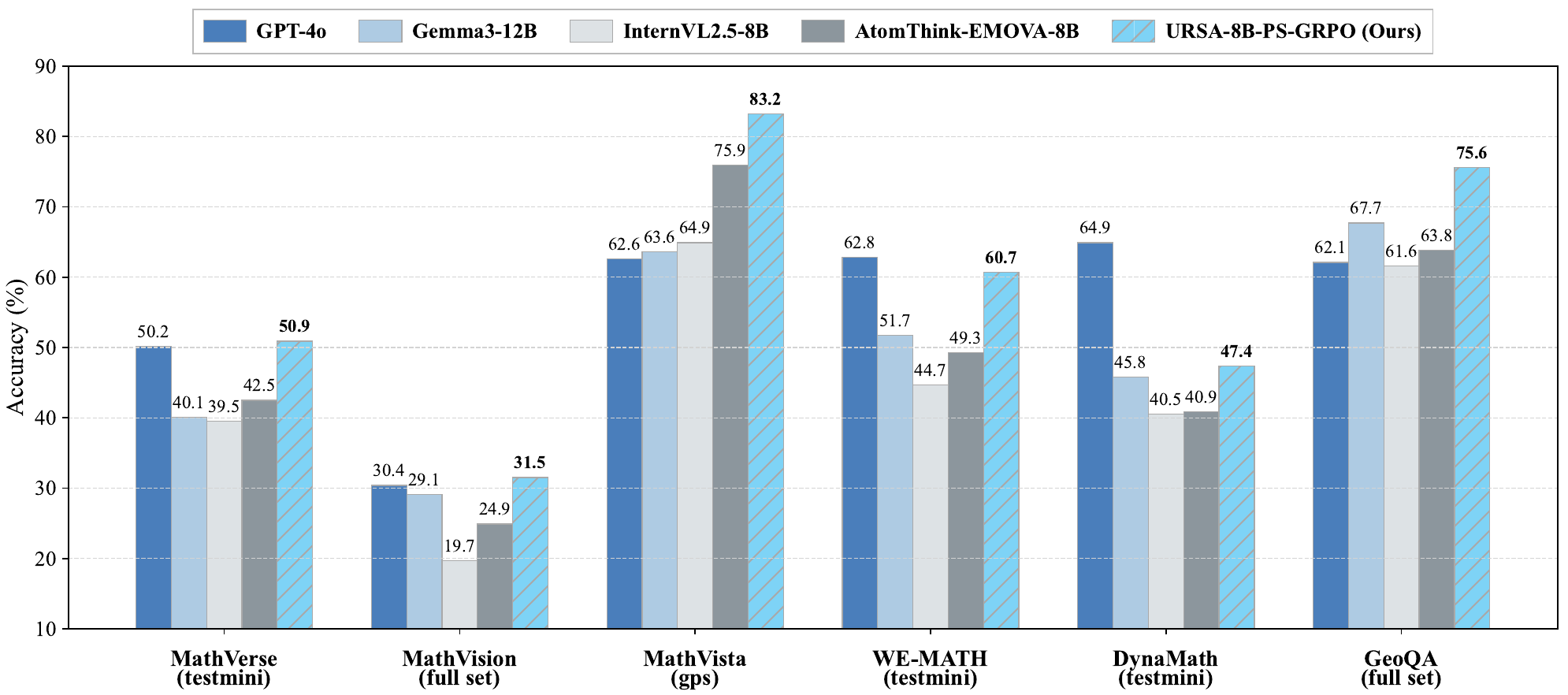}
    \caption{Performance comparison with leading open-source MLLMs and GPT-4o.}
    \label{fig:teaser_graph}
    \vspace{-10pt}
\end{figure*}

In this work, we take the first step toward integrating PRMs into multimodal math reasoning. We identify three key challenges:
(\romannumeral1) Since both TTS and RL are heavily influenced by the strength of foundation models~\citep{yue2025limit-of-rlvr,liu2025can}, the limited availability of large-scale, high-quality reasoning data constrains the upper bounds of current MLLMs and weakens the effectiveness of PRM integration;
(\romannumeral2) There hasn’t yet been adequate automated process labeling techniques merged within multimodal contexts,  where both logical validity and perceptual consistency should be emphasized~\citep{yan2024errorradar,zhang2024critic,ai2025projudge}.
(\romannumeral3) While PRMs can be effectively used in TTS, applying them directly in online RL introduces risks such as reward hacking and length bias in rewarding~\citep{weng2024rewardhacking,fu2025reward}.

To address these challenges, we propose the \textbf{URSA} framework, a three-stage \textbf{U}nfolding multimodal p\textbf{R}ocess-\textbf{S}upervision \textbf{A}ided training pipeline that supports both the construction and application of multimodal PRMs.
In Stage~\RN{1}, we curate \textbf{MMathCoT-1M}, a large-scale, high-quality multimodal Chain-of-Thought dataset synthesized from 1.43 million open-source examples, which enhances the foundation model's reasoning capabilities through targeted instruction tuning.
In Stage~\RN{2}, we construct \textbf{DualMath-1.1M} via a dual-view process supervised data synthesis strategy which combines a binary error locating engine and a misinterpretation insertion engine. It provides complementary signals for logical validity and visual grounding, and is used to train a process reward model.
In Stage~\RN{3}, we analyze the limitations of scalar process reward modeling in online RL and propose \textbf{Process Supervision-GRPO (PS-GRPO)}, which mitigates reward hacking and PRM's length bias in rewarding by implicitly penalizing process-level inconsistencies during policy optimization.

Results on 6 multimodal reasoning benchmarks show that our PRM improves Best-of-N verification, surpassing self-consistency and outcome-based baselines. When used in PS-GRPO, the resulting model achieves state-of-the-art performance among open-source MLLMs of similar size. 
Our contributions are as follows:

\begin{itemize}[left=1em] 
\item We release two large-scale open-source datasets, MMathCoT-1M and DualMath-1.1M, to address the scarcity of high-quality multimodal CoT reasoning and process supervision data. 
\item We propose PS-GRPO, an online reinforcement learning algorithm that incorporates multimodal PRMs by comparing the relative quality of rollouts, rather than relying on scalar reward modeling. It effectively mitigates PRM's reward hacking and length bias in rewarding.
\item Experimental results show that our reward model improves both test-time verification and online training. With PS-GRPO application~(Figure~\ref{fig:teaser_graph}), URSA-8B-PS-GRPO outperforms Gemma3-12B and GPT-4o by 8.4\% and 2.7\% on average across 6 benchmarks.
\end{itemize}
\section{Stage~\RN{1}: Math-Intensive Alignment and Instruction Tuning}\label{sec:align_and_sft}
\subsection{Collection of Vision-Language Alignment Data}
\label{sec:alignment}
We employ a LLaVA-like architecture and first collect vision-language alignment data directly from existing open-source datasets~\citep{li2024llava,zhai2023sigmoid,kirillov2023segment,lu2024deepseek}. As demonstrated in Figure~\ref{fig:statistics_2}, we collect URSA-Alignment-860K from Multimath~\citep{peng2024multimath}, MAVIS~\citep{zhang2024mavis} and Geo170K~\cite{DBLP:journals/corr/abs-2312-11370}. We then filter out samples with overly verbose captions, to form an~860K math-intensive alignment dataset. 
Following the engineering practices of previous work, we only train the MLP projector in the alignment step.
\subsection{CoT Reasoning Data Synthesis}
\begin{figure}[!t]
    \centering
    \includegraphics[width=0.9\linewidth]{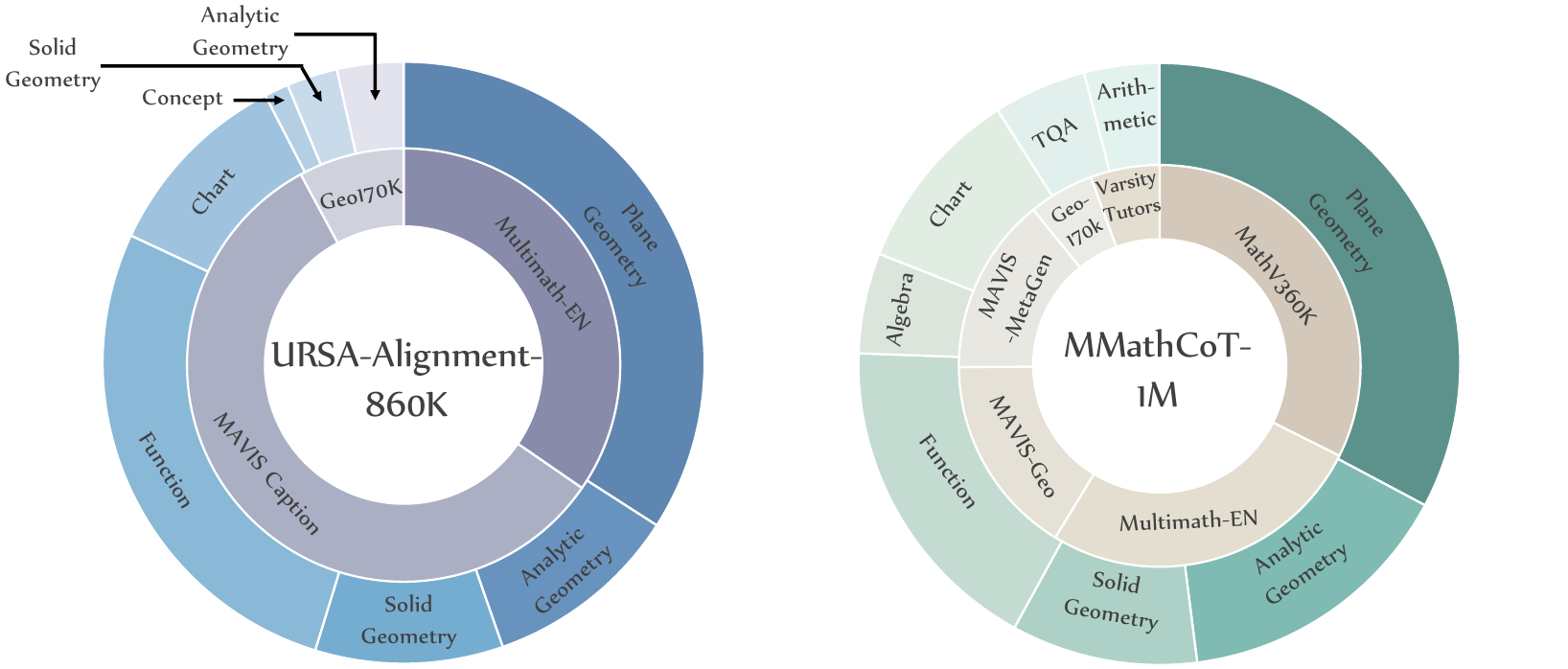}
    \vskip -0.05in
    \caption{Statistics of URSA-Alignment-860K and MMathCoT-1M.}
    \label{fig:statistics_2}
\end{figure}

For a powerful foundation building, we collect 1.43M samples from existing math reasoning datasets to support the construction of large-scale CoT reasoning data. As shown in Figure~\ref{fig:statistics_2}, data is sourced from MathV360K~\citep{shi2024math}, Multimath~\citep{peng2024multimath}, MAVIS~\citep{zhang2024mavis}, Geo170K~\citep{DBLP:journals/corr/abs-2312-11370} and VarsityTutors~\citep{zhuang2025math}. Based on the type of solution, we categorize the data into \textit{answer-only}, \textit{analysis-formatted}, and \textit{CoT-formatted}. We adopt different synthesis strategies for them to curate high-quality CoT reasoning trajectories. We utilize Gemini-1.5-Flash-002~(refer to $\mathcal{G}$ below) as a cost-effective tool for data curation, avoiding expensive large-scale manual annotation. 
\paragraph{CoT Expansion.} For~\textit{answer-only} data $\mathcal{D}_1=\{(x_i,y_i)\}_{i=1}^{N_1}$, such as MathV360K~\citep{shi2024math}, each sample contains a question $x_i$ and a ground-truth answer $y_i$. This type of data is heavily used in previous works for fast thinking reasoning mode~\citep{shi2024math,zhuang2025math,cai2024geogpt4v}. However, answer-only training restricts the model from fully capturing the problem-solving process. It may lead to memory-based reasoning, hindering the model's ability to directly provide answers to more complex reasoning problems~\citep{trinh2024solving}. We expand certain scale CoT reasoning trajectories for this category of data. Given a expansion prompt $\mathcal{P}_{\mathcal{C}}$, we provide $x_i$ and $y_i$, then prompt~$\mathcal{G}$ to output the reasoning trajectory leading to the answer $y_i$, yielding the expanded solutions $\mathcal{S}_{Ao}=\mathcal{G}(\mathcal{P}_{\mathcal{C}};\{x_i,y_i\}_{i=1}^{N_1})$.
\paragraph{Rewriting.} This strategy is designed for \textit{analysis-formatted} samples, denoted as $\mathcal{D}_2=\{(x_i,y_i,a_i)\}_{i=1}^{N_2}$. This includes datasets like MAVIS-Geo, MAVIS-MetaGen~\citep{zhang2024mavis}, VarsityTutors~\citep{zhuang2025math}, and Geo170K-QA~\citep{chen2021geoqa}. Each sample contains a question $x_i$, an answer $y_i$, and textual analysis $a_i$. While this type of data provides walkthroughs, it often suffers from two issues: (\romannumeral1) It lacks strict step-by-step logic, exhibiting jumps in language or reasoning. (\romannumeral2) A significant portion of the answers are relatively brief and cannot provide rich rationale. Given a rewriting prompt $\mathcal{P}_{\mathcal{R}}$, we utilize $\mathcal{G}$ to transcribe these solutions, thereby enhancing their step-by-step reasoning trajectories and linguistic diversity, resulting in the rewritten set $\mathcal{S}_{An}=\mathcal{G}(\mathcal{P}_{\mathcal{R}};\{x_i,y_i,a_i\}_{i=1}^{N_2})$.
\paragraph{Format Unification.} This strategy is used for \textit{CoT-formatted} data, primarily sourced from Multimath-EN-300K~\citep{peng2024multimath}, which is collected from K-12 textbooks and contains mathematical language and symbolic-style reasoning solutions. This portion of the data, $\mathcal{D}_{3}=\{(x_i,y_i,c_i)\}_{i=1}^{N_3}$, consists of a question $x_i$, an answer $y_i$, and a solution $c_i$. We unify the format through natural language stylization using a prompt $\mathcal{P}_{\mathcal{F}}$ with $\mathcal{G}$, producing the unified set $\mathcal{S}_{C}=\mathcal{G}(\mathcal{P}_{\mathcal{F}};\{x_i,y_i,c_i\}_{i=1}^{N_3})$.\begin{figure*}[!t]
    \centering
    \includegraphics[width=1.0\linewidth]{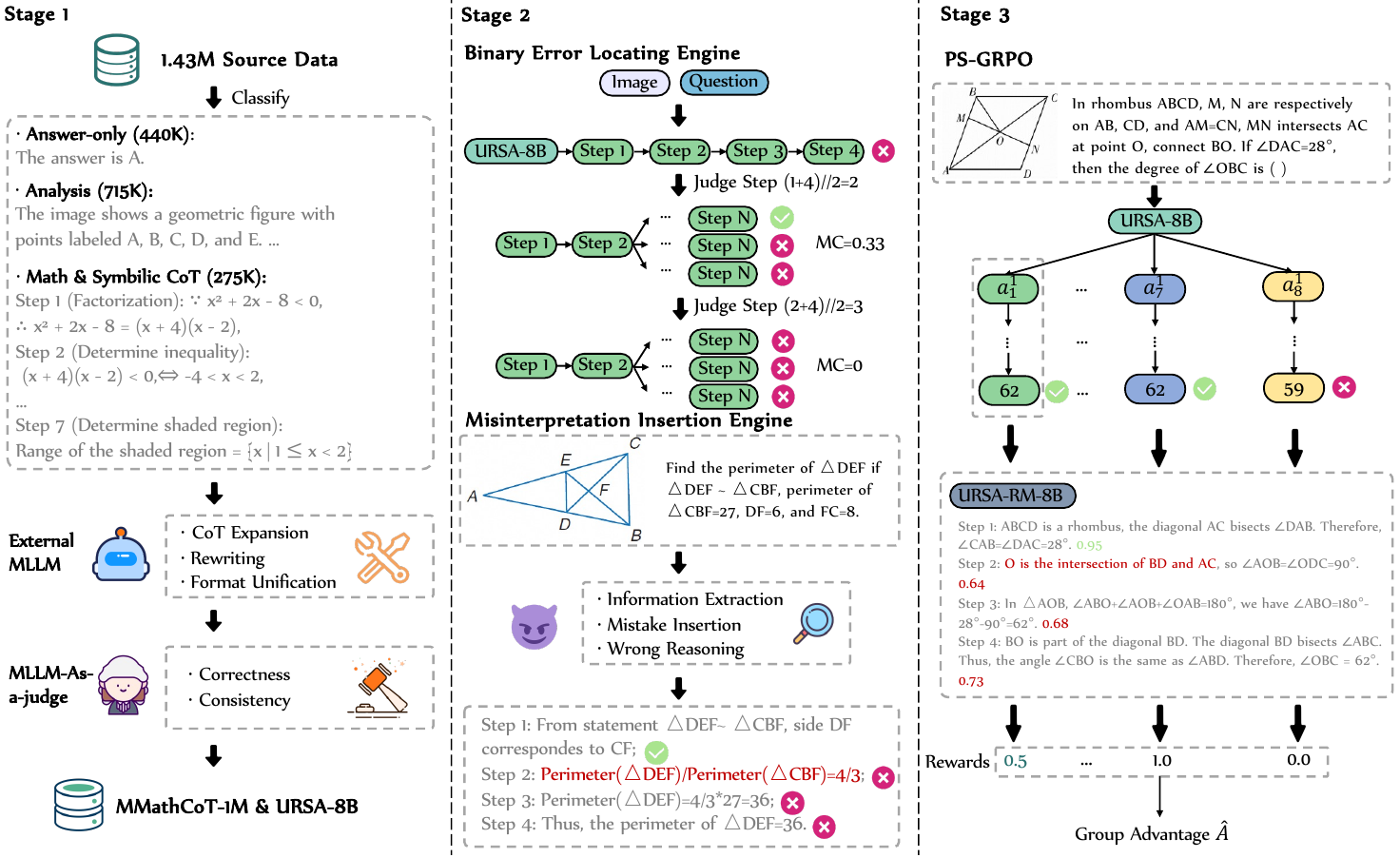}
    \caption{Pipeline of URSA. Stage 1 depicts the workflow of data curation as described in Section~\ref{sec:align_and_sft}. Stage 2 illustrates how binary error locating and misinterpretation insertion facilitate the automation of process supervision data. Stage 3 demonstrates how our PS-GRPO operates by imposing penalties on rollouts that are questioned by the PRM.}
    \label{fig:main_graph}
\end{figure*}\paragraph{MMathCoT-1M.} Finally, we filter out instances where: (\romannumeral1) Correctness is violated: the generated content altered the original answer, or (\romannumeral2) Consistency is problematic: the solution includes text that questions the original answer or makes new assumptions to force the given answer. 
This process yields MMathCoT-1M. The complete prompt designs can be found in Appendix~\ref{app:prompt_design}.

We perform full-parameter instruction fine-tuning with MMathCoT-1M to train URSA-8B, based on the aligned model. The SFT dataset $\mathcal{D}_{SFT}$ is formed by the union of the curated solutions, i.e., $\mathcal{D}_{SFT} = \{(x_i,y_i) \mid (x_i,y_i) \in \mathcal{S}_{Ao} \cup \mathcal{S}_{An} \cup \mathcal{S}_{C}\}$. Training objective is demonstrated in Equation~\ref{equ:sft_loss}.
\begin{equation}\label{equ:sft_loss}
\begin{aligned}
\mathcal{L}_{SFT} &= -\mathbb{E}_{(x,y) \sim \mathcal{D}_{SFT}} \sum_{t=1}^T \log \mathcal{M}(y_t | x, y_{<t}) 
\end{aligned}
\end{equation}
In this phase, we construct a stronger reasoning foundation model, URSA-8B, with the expectation of achieving a higher bound at inference time and to process supervision data of greater diversity.


\section{Stage \RN2: Dual-View Process Supervised Data Synthesis}
\subsection{Binary Error Locating Engine}
Following suggestions by previous work~\citep{lightman2023let,luo2024improve,wang2024math}, we train a PRM for first error step identification. We collect $\sim$553K incorrect solutions from URSA-8B's zero-shot inference on MMathCoT-1M. Erroneous steps in these solutions are labeled using Monte Carlo Tree Search (MCTS). For MCTS, an operation $\mathcal{F}(\{s_1, \ldots, s_i\}, N)$ generates $N$ rollouts from a reasoning prefix $\{s_1, \ldots, s_i\}$. The single step's Monte Carlo estimation value, $mc_i$, is the fraction of these rollouts leading to a correct answer:
\begin{equation}
    mc_i=\frac{|\text{Correct rollouts from $\mathcal{F}(\{s_1, s_2, \ldots, s_i\}, N)$}|}{|\text{Total rollouts from $\mathcal{F}(\{s_1, s_2, \ldots, s_i\}, N)$}|}
\end{equation}
A step $s_i$ is deemed \enquote{potentially correct} if $mc_i > 0$~\citep{wang2024math,luo2024improve}. We optimize the identification of first error step using Binary Error Locating Engine~(BEL): if the middle step has positive $mc$~(i.e. $mc_{mid}>0$), the error is in the latter half; otherwise, in the first (see Algorithm~\ref{alg:binary}). To mitigate step-level label bias and include positive examples, we add $\sim$180K correct solutions (1/3 the number of incorrect ones), with all steps easily marked \enquote{True}. This yields $\mathcal{S}_{BEL}$, a ~773K process annotation dataset based on correctness potential.
\subsection{Misinterpretation Insertion Engine}\label{sec:misinterpretation}
Apart from logical errors, the perception inconsistency between images and text in reasoning steps is a unique problem in multimodal scenarios~\citep{yan2024errorradar,zheng2024thinking,gao2023pal}. We propose a Misinterpretation Insertion Engine~(MIE) to artificially insert hallucinatory information, automatically constructing process supervision data with incorrect reasoning paths starting from the insertion point.
Specifically, MIE includes three steps. First, we prompt $\mathcal{G}$ to perform a captioning task, extracting mathematical paradigm information from the image as much as possible. Second, the model $\mathcal{G}$ is required to focus on potentially confusable conditions within the existing correct solution and modify them using adjacent or similar conditions. Finally, the model $\mathcal{G}$ is prompted to continue reasoning based on the step with the inserted error. We leverage strong instruction-following capability of $\mathcal{G}$, instructing it to automatically assign negative labels to every subsequent step following the erroneous insertion. We generate $\sim$302K samples $\mathcal{S}_{MIE}$ using this strategy. Cases from MIE can be found in the Appendix~\ref{app:sec:mie_case}.
\subsection{PRM Training}
As shown in Equation~\ref{equ:merge}, we merge two types of data, proposing a $\sim$1.1M process supervision data called DualMath-1.1M. During training, we append a special token after each step to indicate its predicted correctness. We model the PRM training as a binary classification task for the correctness of each step, as shown in Equation~\ref{equ:prm}, here $\pi_{p}$ is the trained PRM based on URSA-8B. $e_j$ and $y_j$ represent single step and corresponding label ($y_j\in \{0, 1\}$).
\begin{gather}
    \mathcal{D}_{PRM} = \{ (e, y_{e}) \sim \mathcal{S}_{BEL} \cup \mathcal{S}_{MIE} \} \label{equ:merge} \\
    \mathcal{L}_{PRM} = -\mathbb{E}_{(e, y) \sim \mathcal{D}_{PRM}} \sum_{j=1}^{|e|} \Big[ y_j \log \pi_p(e_j) + (1-y_j) \log (1-\pi_p(e_j)) \Big] \label{equ:prm}
\end{gather}
Thus, Stage \RN{2} delivers URSA-8B-RM, a strong PRM trained on DualMath-1.1M---the \textbf{first} large-scale, automatically labeled dataset for multimodal reasoning process supervision. While BoN evaluation demonstrates PRM's value in TTS, a critical question emerges: how can its guidance be directly integrated into MLLM post-training? This remains largely uncharted. Stage \RN{3} draws a lesson about why previous scalar process reward modeling tends to fail, and then we achieve effective progress through process-as-outcome reward modeling.

\section{Stage \RN3: Integrating multimodal PRM into RL}\label{sec:stage_3}
Inspired by successes like DeepSeek-R1~\citep{guo2025deepseek}, several recent studies have tried to adapt outcome reward-based GRPO for multimodal reasoning, demonstrating notable progress~\citep{pan2025medvlm,zhan2025visionr1,huang2025visionr1,liu2025noisyrollout}. Outcome reward-based GRPO computes the $i$-th response's advantage through normalizing in-group rewards. However, outcome reward-based GRPO ignores the quality of reasoning processes~\citep{lightman2023let,li2024qvalue,amrith2024reward}.

\begin{figure}[!t]
    \centering
    \includegraphics[width=1.0\linewidth]{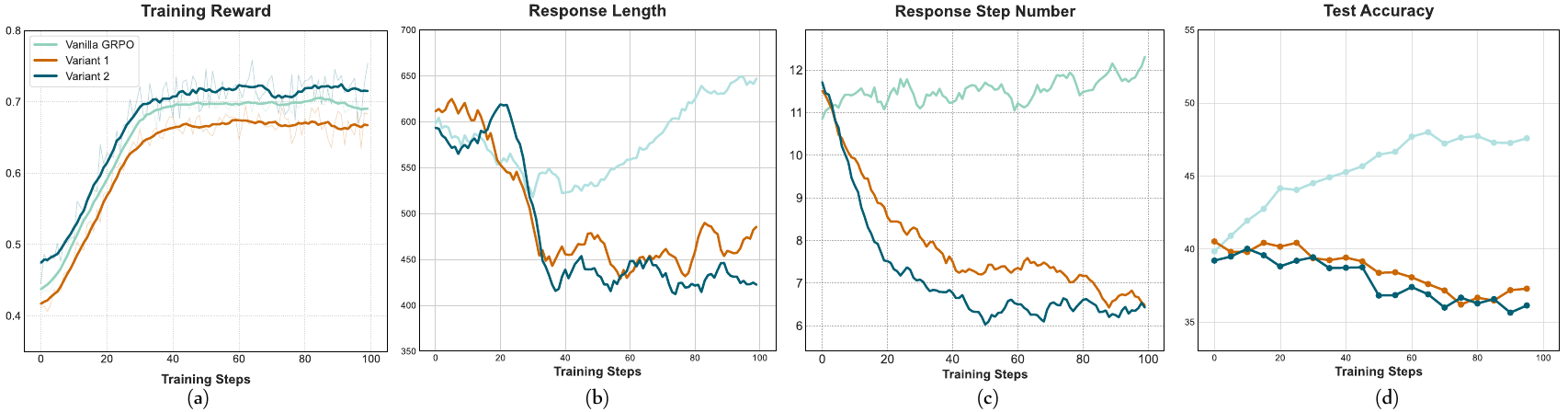}
    \vskip -0.1in
    \caption{Figure (a)-(d) respectively illustrate training rewards, response length, response step number and test set accuracy of vanilla GRPO and two variants proposed in Section~\ref{sec:stage_3}. Test set is randomly selected 500 examples from MMathCoT-1M for an in-domain evaluation.}
    \label{fig:sft_abl_total}
\end{figure}
Following most standard response-level and step-level reward modeling in RL~\citep{wang2024math,liu2024diving,guo2025deepseek,huang25visionr1,ma2025slm}, we examine two simple variants of GRPO with integrated scalar process rewards to reveal the failure patterns during the training process~\citep{gao2024ondesign}. 
\textit{Variant 1}: For $i$-th rollout, the reward is the sum of the outcome reward and the average process reward, i.e. $r^i=r_o^i+\bar{r_s^i}$. \textit{Variant 2}: Despite the outcome reward, a scalar process reward $r_{s,t}^i$ is assigned to the $i$-th rollout's $t$-th step.
\begin{figure}[!t]
    \centering
    \includegraphics[width=1.0\linewidth]{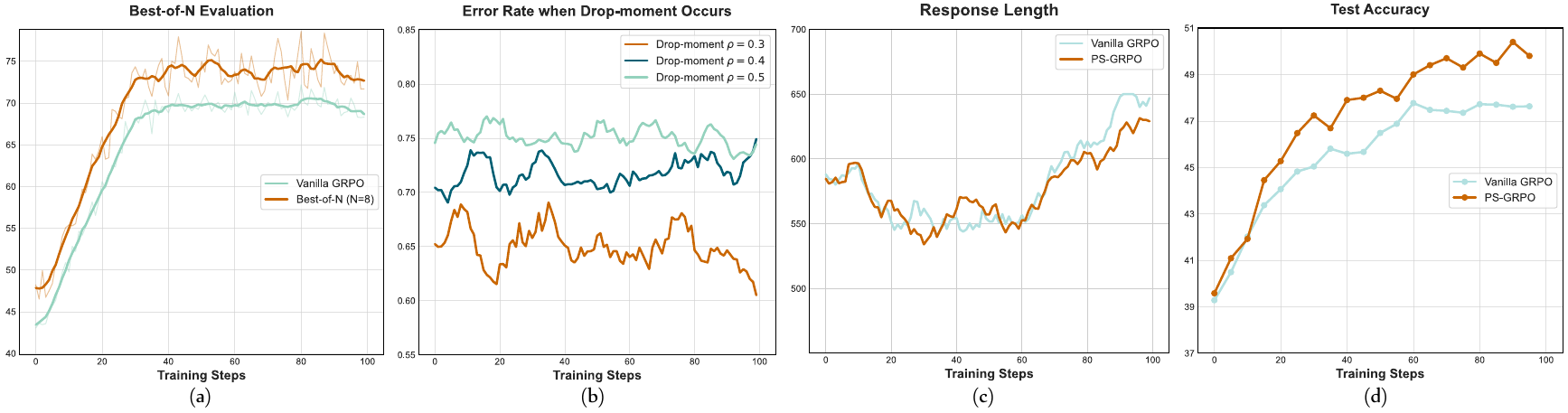}
    \vskip -0.1in
    \caption{Figure (a) shows the BoN evaluation during GRPO training. We select the best rollout using the mean value of process rewards. Figure (b) illustrates the proportion of rollouts where URSA-8B-RM identifies \enquote{drop-moment} and the final results are indeed incorrect. Figures (c) and (d) display the response length and test accuracy during PS-GRPO training.}
    \vspace{-1em}
    \label{fig:ps_grpo}
\end{figure}
We observe two highly significant conclusions from Figure~\ref{fig:sft_abl_total}: \textbf{(\romannumeral1)}~\textit{High susceptibility to reward hacking}. The test accuracy of both variants is lower than vanilla GRPO. 
This indicates that when process scalar rewards are employed as learning objectives, the model quickly learns strategies that cater to process correctness. However, correctness in the process does not necessarily correlate fully with the heuristics leading to the ground-truth.
\textbf{(\romannumeral2)}~\textit{PRM's length bias in rewarding}. We observe a trend where increased training leads to shorter model responses and fewer reasoning steps. This phenomenon stems from an inherent length bias in the PRM's training labels; for examples with incorrect answers, steps taken after the first error are unlikely to yield a correct solution. This results in the PRM conservatively rewards the later stages of a reasoning rollout, thereby encouraging the MLLM towards more passive reasoning and a reliance on pattern recognition from existing conditions or simpler heuristics.
\paragraph{PS-GRPO} The findings above confirm the consideration that flaws in the reward function are amplified when scalar process rewards serve as the optimization target~\citep{amodei2016concrete,weng2024rewardhacking}. 
We ask \enquote{Which internal signals of PRM can be trusted?}
We employ two views to investigate the reliable region of the PRM: first, the BoN performance during online learning, and second, the PRM's error identification capability. 
Regarding the latter, we introduce the concept of a \textit{\enquote{drop-moment}} within the PRM's reward sequence, which signifies that the PRM questions the validity of the preceding steps. Specifically, for a given solution's PRM reward sequence $\{r_{p1}^i, r_{p2}^i, \cdots, r_{pN}^i \}$, a significant decrease in reward between consecutive steps indicates the occurrence of such a drop-moment.
\begin{equation}\label{equ:drop_moment}
    \begin{aligned} 
         \delta_p^i = \max \left\{ \frac{r_{p,j}^i - r_{p,j+1}^i}{r_{p,j}^i} \middle| j=0, 1, \dots, N-1 \right\} > \rho
    \end{aligned}
\end{equation}
Here, $\rho$ represents PRM's drop-moment threshold. As illustrated in Figure~\ref{fig:ps_grpo}, the PRM's ability for BoN selection and error identification remains largely unimpaired during the online RL process, exhibiting stable performance. This suggests that \textit{although the scalar reward from the PRM in online RL might be unreliable, the relative quality of solutions it reveals is comparatively trustworthy}.
We leverage this beneficial property to address the reward sparsity problem in GRPO~\citep{zhang2025grpo,zhang2025r1,yu2025dapo}, aiming to make online RL focus more on learning from rollouts that have accurate results and rigorous processes. We use $\rho$ from Equation~\ref{equ:drop_moment} as the occurrence threshold for a \enquote{drop-moment}; when it occurs, we apply a reward penalty~$\gamma$ to rollouts with correct results. This both differentiates the learning value of outcome-correct rollouts and, due to its focus on relative drops in reward sequences, circumvents the impact of PRM's length bias in rewarding.
\begin{equation}\label{equ:reward_modeling} 
    \begin{aligned}
        R^i = 
\begin{cases} 
1, & o^i\text{ is correct and } \delta_p^i < \rho\\
1-\gamma, & o^i\text{ is correct and } \delta_p^i \geq \rho \\
0, & \text{otherwise}
\end{cases}
    \end{aligned}
\end{equation}
We utilize reward modeling in Equation~\ref{equ:reward_modeling} to conduct a process-supervised GRPO, which facilitates the computation of in-group advantages in Equation~\ref{equ:adv}.
%
\begin{table*}[!t]
\centering
\caption{Performance Comparison on 6 math reasoning benchmarks. We use accuracy for MathVerse, MathVision, MathVista and GeoQA. We use Score~(Loose) on WE-MATH. And average-case accuracy is employed on DYNAMATH. Best results of Closed-source MLLMs are highlighted in \textcolor{back_deepgreen}{green}. Best and runner-up results of Open-source MLLMs are highlighted in \textcolor{back_deepred}{red} and \textcolor{back_deepblue}{blue}.}
\label{tab:performance_opensource_highlight}
\resizebox{1\textwidth}{!}{%
\begin{tabular}{lcccccccc}
\toprule
 & \multirow{2}{*}{\textbf{Size}} & \multirow{2}{*}{\textbf{Avg}} & \textbf{MathVerse} & \textbf{MathVision} & \textbf{MathVista} & \textbf{WE-MATH} & \textbf{DYNAMATH} & \textbf{GeoQA} \\
 & & & testmini & full set & gps & testmini & testmini & full set \\
\midrule
\multicolumn{9}{c}{ \textit{Closed-Source MLLMs} } \\ \midrule
GPT-4o~\citep{openai2024gpt4o} & -& \highg{55.5} & \highg{50.2} & \highg{30.4}& 64.7 & 62.8 & \highg{64.9} & \highg{62.1} \\ 
GPT-4o-mini~\citep{openai2024gpt4o} & -& 49.2 & 42.3 & 22.8 & 59.9 & 56.3 & 53.5 & 60.1 \\ 
Gemini-1.5-pro~\citep{team2023gemini} & - & 53.2 & 35.3 & 19.2 & \highg{81.7} & \highg{66.9} & 60.5 & 55.5 \\ 
\midrule
\multicolumn{9}{c}{ \textit{Open-Source General MLLMs} } \\
 \midrule
InternVL-Chat-V1.5~\citep{chen2024far} & 26B & 33.6 & 26.1 & 15.4 & 56.9 & 32.7 & 36.7 & 33.5 \\
Llama-3.2-11B-Vision-Instruct~\citep{metaLlama32} & 11B & 28.0 & 28.9 & 16.9 & 40.9 & 12.0 & 32.2 & 36.9 \\
Qwen2-VL~\citep{wang2024qwen2} & 8B & 40.2 & 33.6 & 19.2 & 51.0 & 43.0 & 42.1 & 52.2 \\
InternVL2-8B~\citep{chen2024internvl} & 8B & 41.8 & 37.0 & 18.4 & 57.7 & 44.9 & 39.7 & 52.8 \\
InternVL2-8B-MPO~\citep{wang2024enhancing} & 8B & 45.1 & 38.2 & 22.3 & 69.2 & 44.4 & 40.5 & 55.9 \\
InternVL2.5-8B~\citep{chen2024expanding} & 8B & 45.2 & 39.5 & 19.7 & 64.9 & 44.7 & 40.5 & 61.6 \\
LLaVA-OneVision~\citep{li2024llava} & 8B & 40.9 & 28.9 & 18.3 & 71.6 & 44.9 & 37.5 & 43.9 \\
Points-Qwen2.5-Instruct~\citep{liu2024points} & 8B & 49.8 & 41.1 & 23.9 & 76.0 & 51.0 & 42.8 & 63.8 \\
Gemma3-12B~\citep{team2025gemma}& 12B & 49.8 & 40.1 & \high{29.1} & 63.6 & 51.7 & \high{45.8} & 67.7 \\
\midrule
\multicolumn{9}{c}{\textit{Open-Source Reasoning MLLMs}} \\ \midrule
Math-LLaVA~\citep{shi2024math} & 13B & 35.2 & 22.9 & 15.7 & 57.7 & 31.3 & 35.5 & 48.1 \\
MathPUMA-Qwen2-7B~\citep{zhuang2025math} & 8B & 39.6 & 33.6 & 14.0 & 48.1 & 41.0 & 37.3 & 63.6 \\
MultiMath~\citep{peng2024multimath} & 7B & 43.1 & 27.7 & 16.3 & 66.8 & 42.2 & 37.9 & 67.7 \\
MAVIS~\citep{zhang2024mavis} & 7B & 44.4 & 35.2 & 18.5 & 64.1 & 44.3 & 36.2 & 68.3 \\
InfiMM-Math~\citep{han24infimm} & 7B & 48.6 & 40.5 & 18.8 & 77.3 & 48.3 & 38.2 & 68.3 \\
AtomThink-EMOVA~\citep{xiang2024atomthink} & 8B & 49.5 & 42.5 & 24.9 & 75.9 & 49.3& 40.9 & 63.8 \\
MathGLM-Vision~\citep{yang2024mathglmv} & 9B & 47.6 & 44.2 & 19.2 & 64.2  & 45.2 & 42.2 & 70.4 \\
LlamaV-o1~\citep{DBLP:journals/corr/abs-2501-06186} & 11B & 38.4 & 33.9 & 17.9 & 53.3 & 42.6 & 34.7 & 43.1 \\
OpenVLThinker~\citep{deng25openvlthiner} & 7B & - & \high{47.9} & 25.3 & 76.4 & - & - & - \\
R1-Onevision~\citep{yang2025r1onevision} & 7B & - & 47.4 & 26.9 & 72.4 & 51.4 & - & - \\
\midrule
URSA-8B & 8B & \high{54.7} & 45.7 & 28.7 & \high{81.7} & \high{53.6} & 44.7 & \high{73.5} \\
URSA-8B-PS-GRPO & 8B & \highr{58.2} & \highr{50.9} & \highr{31.5} & \highr{83.2} & \highr{60.7} & \highr{47.4} & \highr{75.6} \\
\bottomrule
\end{tabular}%
}
\end{table*}
\begin{table*}[!t]
  \vspace{-0.5em}
  \caption{Comparison of TTS on URSA-8B and AtomThink-EMOVA using BoN performance.}
  \centering
  \label{tab:prm_comparison_final_updated_gps} 
  \resizebox{1\textwidth}{!}{%
  \begin{tabular}{llcccccccccccc}
    \toprule
    \multirow{2}{*}{\textbf{Model}} & \multirow{2}{*}{\textbf{Method}} & \multicolumn{4}{c}{\textbf{MathVerse}} & \multicolumn{4}{c}{\textbf{MathVista-GPS}} & \multicolumn{4}{c}{\textbf{MathVision}} \\
    \cmidrule(lr){3-6} \cmidrule(lr){7-10} \cmidrule(lr){11-14}
    & & N=4 & N=8 & N=16 & N=32 & N=4 & N=8 & N=16 & N=32 & N=4 & N=8 & N=16 & N=32 \\
    \midrule
    \multirow{3}{*}{URSA-8B}
    & Self-Consistency & 49.3 & 50.1 & 50.7 & 50.7 & 82.7 & 83.9 & 84.8 & 85.4 & 29.4 & 31.9 & 32.8 & 33.1 \\ 
    & InternVL2.5-8B ORM & 48.6 & 50.9 & 51.8 & 51.3 & 82.5 & 83.3 & 84.3 & 85.1 & 29.9 & 32.1 & 32.8 & 33.5 \\ 
    & URSA-8B-RM & \textbf{53.3} & \textbf{54.2} & \textbf{54.7} & \textbf{55.0} & \textbf{83.2} & \textbf{85.5} & \textbf{86.5} & \textbf{87.2} & \textbf{31.6} & \textbf{33.1} & \textbf{34.0} & \textbf{35.1} \\ 
    \midrule
    \multirow{3}{*}{AtomThink-EMOVA}
    & Self-Consistency & 45.9 & 46.7 & 47.1 & 47.3 & 76.8 & 77.9 & 78.6 & 79.0 & 25.3 & 26.8 & 27.6 & 28.0 \\
    & InternVL2.5-8B ORM & 45.7 & 45.6 & 46.4 & 46.1 & 76.6 & 77.7 & 78.3 & 79.2 & 26.0 & 26.6 & 27.2 & 27.8 \\ 
    & URSA-8B-RM & \textbf{48.0} & \textbf{48.8} & \textbf{49.3} & \textbf{49.6} & \textbf{78.0} & \textbf{79.6} & \textbf{80.5} & \textbf{81.0} & \textbf{27.5} & \textbf{29.0} & \textbf{30.2} & \textbf{31.0} \\ 
    \bottomrule
  \end{tabular}%
  }
\end{table*}
\section{Experiments}\label{sec:exp}
\subsection{Experimental Setup}\label{sec:exp_setup}
\paragraph{Benchmarks} We evaluate our URSA-series models on 6 widely used reasoning benchmarks, including MathVerse~\citep{zhang2025mathverse}, DYNAMATH~\citep{zou2024dynamath}, MathVista~\citep{lu2023mathvista}, WE-MATH~\citep{qiao2024we}, GeoQA~\citep{chen2021geoqa} and MathVision~\citep{wang2024math}. Detailed description and evaluation criteria can be found in Appendix~\ref{app:sec:benchmark}. We consistently employ zero-shot inference for comparison.
\paragraph{Baselines} We include some leading proprietary MLLMs, such as GPT-4o and GPT-4o-mini~\citep{openai2024gpt4o}. For open-source MLLMs with comparable size, we select InternVL-series~\citep{chen2024internvl,dong2024internlm}, LLaVA-OneVision~\citep{li2024llava}, Gemma3-12B~\citep{team2025gemma}, Qwen2-VL~\citep{wang2024qwen2}, and so on. For MLLMs intended for math reasoning purposes, we select AtomThink~\citep{xiang2024atomthink}, InfiMM-Math~\citep{han24infimm}, MAVIS~\citep{zhang2024mavis}, MathGLM-Vision~\citep{yang2024mathglmv}, LlamaV-o1~\citep{DBLP:journals/corr/abs-2501-06186}. This kind of work focuses on the synthesis of STEM reasoning data or o1-like slow thinking. We do not select baselines that use MathVision as training set for fairness, such as Mulberry-Qwen2-VL-7B~\citep{yao2024mulberry} and MAmooTH-VL~\citep{guo2024mammoth}. For PRM's TTS performance, we select Self-Consistency~\citep{wang2022self} and open-source MLLM as ORM for comparison, such as InternVL2.5-8B~\citep{chen2024internvl}.
\paragraph{Implementation Details} URSA uses SAM-B+SigLIP-L as the hybrid vision encoder and Qwen2.5-Math-Instruct as the LLM backbone. We employ a two-layer MLP connection for vision-language alignment training. We select 15K data in MMathCoT-1M for PS-GRPO. $\gamma$ and $\rho$ in Equation~\ref{equ:reward_modeling} are set to 0.5 and 0.3, respectively. Details on module selection, data selection, hyperparameters, and time cost are placed in the Appendix~\ref{app:selection_criteria} and~\ref{app:implementation_details}.
\subsection{Main Results}
\begin{figure}[!t]
    \centering
    \includegraphics[width=1.0\linewidth]{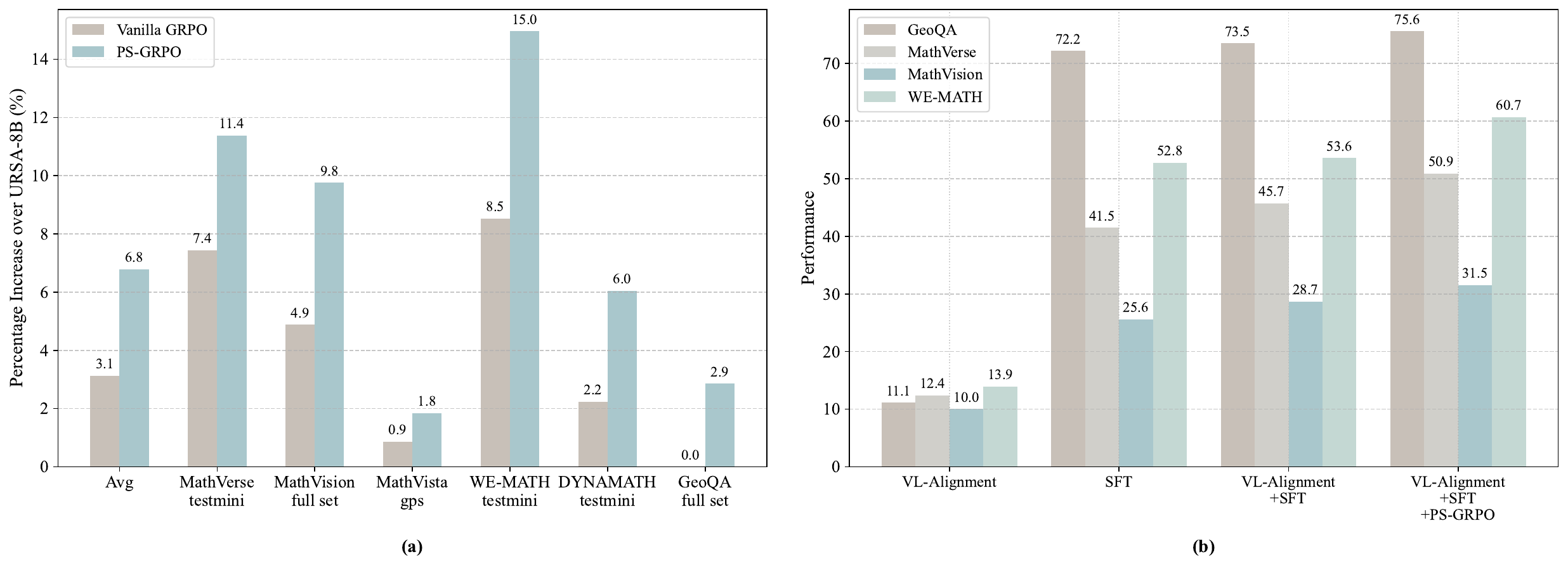}
    \vskip -0.1in
    \caption{Figure(a) represents the comparison of relative improvements on URSA-8B; Figure(b) illustrates how each training stage contributes to the total performance.}
    \label{fig:ps_grpo_data}
\end{figure}

\paragraph{SoTA Performance} In Table~\ref{tab:performance_opensource_highlight}, we present the performance of URSA-8B and URSA-8B-PS-GRPO. First, URSA-8B provides a stronger reasoning foundation model. It demonstrates a 5.2 point advantage over AtomThink-EMOVA which focuses on \enquote{slow thinking} training. It also outperforms leading general-purpose MLLMs of comparable size, such as Gemma3-12B and InternVL2.5-8B.
URSA-8B-PS-GRPO outperforms GPT-4o across 6 benchmarks on average and shows significant advantages on MathVista-GPS (83.2 vs 62.6), GeoQA~(73.5 vs 62.1), and achieves the first surpassing performance on MathVision~(31.5 vs 30.4).
However, a significant performance gap on DynaMath suggests that smaller-scale MLLMs still lack more robust problem-solving capabilities.
Compared to the leading math reasoning MLLM AtomThink-EMOVA-8B and general-purpose MLLM Gemma3-12B in terms of average performance, our model shows advantages of \textbf{8.5\%} and \textbf{8.2\%}, respectively.
Compared with recent R1-inspired method OpenVLThinker~\citep{deng25openvlthiner} and R1-Onevision~\citep{yang2025r1onevision}, we still show significant advantage on MathVision and WE-MATH. 
\paragraph{Effective Best-of-N Evaluation} 
In Table~\ref{tab:prm_comparison_final_updated_gps}, we demonstrate the advantages of URSA-8B-RM compared to self-consistency and the ORM baseline on serving TTS~\citep{wang2024math,luo2024improve}. We find that self-consistency remains a strong baseline, which InternVL2.5-8B (serving as the ORM) does not consistently surpasses. However, URSA-8B-RM exhibits more effective BoN evaluation and demonstrates its generalization on AtomThink-EMOVA-8B. In addition, using URSA-8B-RM as the verifier, only 4 samplings can achieve a huge improvement based on URSA-8B. Specifically, it provides a 16.6\% and 10.1\% relative improvement on MathVerse and MathVision. In Best-of-32 setting, URSA-8B achieve 35.1 and 55.0 in MathVision and MathVerse, showing clear advantage with GPT-4o.
\paragraph{PS-GRPO vs Vanilla GRPO} As shown in Figure~\ref{fig:ps_grpo_data}~(a), given the same training data, hyperparameters, and rollout number PS-GRPO achieves a higher improvement on average performance~(6.8\% vs 3.1\%). PS-GRPO demonstrates an improvement that is nearly double that of vanilla GRPO in WE-MATH and more challenging MathVision, suggesting its effectiveness. We notice that the improvement of RL on MathVista-GPS and GeoQA is relatively small. This is because URSA-8B's inherent abilities have already achieved an effect close to the upper bound on these two benchmarks. However, PS-GRPO still has advantages over vanilla GRPO.
\begin{table*}[!t]
  \centering
  \caption{Ablation study on DualMath-1.1M (BoN evaluation). w/o $\mathcal{S}_{MIE}$ and w/o $\mathcal{S}_{BEL}$ represents dropping one part of DualMath-1.1M to train the PRM.}
  \label{tab:prm_comparison_abl} 
  \resizebox{1\textwidth}{!}{%
  \begin{tabular}{llcccccccccccc}
    \toprule
    \multirow{2}{*}{\textbf{Model}} & \multirow{2}{*}{\textbf{Dataset}} & \multicolumn{4}{c}{\textbf{MathVerse}} & \multicolumn{4}{c}{\textbf{MathVista-GPS}} & \multicolumn{4}{c}{\textbf{MathVision}} \\
    \cmidrule(lr){3-6} \cmidrule(lr){7-10} \cmidrule(lr){11-14}
    & & N=4 & N=8 & N=16 & N=32 & N=4 & N=8 & N=16 & N=32 & N=4 & N=8 & N=16 & N=32 \\
    \midrule
    \multirow{3}{*}{URSA-8B} 
    & DualMath-1.1M & \textbf{53.3} & \textbf{54.2} & \textbf{54.7} & \textbf{55.0} & \textbf{83.2} & \textbf{85.5} & \textbf{86.5} & \textbf{87.2} & \textbf{31.6} & \textbf{33.1} & \textbf{34.0} & \textbf{35.1} \\
    & w/o $\mathcal{S}_{MIE}$ & 52.8 & 52.6 & 52.4 & 53.9 & 81.3 & 83.8 & 83.1 & 83.2 & 29.9 & 30.5 & 33.1 & 34.5 \\ 
    & w/o $\mathcal{S}_{BEL}$ & 50.3 & 51.4 & 51.8 & 53.0 & 80.1 & 83.1 & 82.2 & 83.0 & 28.7 & 29.8 & 32.3 & 34.2 \\ 
    \midrule
    \multirow{3}{*}{AtomThink-EMOVA} 
    & DualMath-1.1M
    & \textbf{48.0} & \textbf{48.8} & \textbf{49.3} & \textbf{49.6} & \textbf{78.0} & \textbf{79.6} & \textbf{80.5} & \textbf{81.0} & \textbf{27.5} & \textbf{29.0} & \textbf{30.2} & \textbf{31.0} \\ 
    & w/o $\mathcal{S}_{MIE}$ & 47.5 & 48.2 & 47.8 & 48.0 & 76.8 & 78.3 & 79.1 & 79.5 & 26.0 & 27.4 & 28.5 & 29.2 \\ 
    & w/o $\mathcal{S}_{BEL}$ & 46.8 & 47.5 & 47.9 & 47.3 & 76.0 & 77.5 & 78.3 & 78.7 & 25.4 & 26.7 & 27.8 & 28.5 \\ 
    \bottomrule
  \end{tabular}%
  }
\end{table*}

\section{Analysis}
\subsection{How Each Stage Contributes the Performance}
In this section, we demonstrate how each stage contributes to the performance. As demonstrated in Figure~\ref{fig:ps_grpo_data}~(b), all stages make a performance contribution. MMathCoT-1M contributes the highest absolute performance gain.
The effect of Alignment-860K is more evident on MathVerse and MathVision, likely because the question images in these two datasets contain richer textual modality information, allowing alignment resources (such as textual images) to better supplement this comprehension capability. PS-GRPO, on the other hand, is dedicated to breaking the bottleneck after large-scale SFT, performing more prominently on WE-MATH and MathVerse with relative improvements of 13.2\% and 11.4\% respectively, compared to URSA-8B. We provide a generalization validation on InternVL2.5-8B and Multimath in Appendix~\ref{app:generalization}.
\begin{table}[!t]
\centering
\caption{Sensitivity analysis on reward penalty and PRM's \enquote{drop-moment} judgment.}
\label{tab:sensitivity_analysis}
\resizebox{0.95\textwidth}{!}{
\small
\begin{tabular}{@{}lcccccccc@{}}
\toprule
\multirow{2}{*}{\textbf{$\gamma$}} & \multirow{2}{*}{\textbf{$\rho$}} & \multicolumn{1}{c}{\textbf{MathVerse}} & \multicolumn{1}{c}{\textbf{MathVision}} & \multicolumn{1}{c}{\textbf{MathVista}} & \multicolumn{1}{c}{\textbf{WE-MATH}} & \multicolumn{1}{c}{\textbf{DYNAMATH}} & \multicolumn{1}{c}{\textbf{GeoQA}} & \multirow{2}{*}{\textbf{Avg}}\\
 & & \multicolumn{1}{c}{\small testmini} & \multicolumn{1}{c}{\small full set} & \multicolumn{1}{c}{\small gps} & \multicolumn{1}{c}{\small testmini} & \multicolumn{1}{c}{\small testmini} & \multicolumn{1}{c}{\small full set} & \\
\midrule
0.5           & 0.3 & 50.9      & 31.5        & \textbf{83.2}       & 60.7     & \textbf{47.4}        & \textbf{75.6}    & \textbf{58.2}\\ 
0.5           & 0.4 & 49.9      & 30.8        & 81.2       & 59.9     & 46.9        & 75.0    & 57.3\\
0.5           & 0.2 & 49.6      & 30.5        & 80.9       & 59.6     & 46.6        & 74.7    & 57.0\\
1.0           & 0.3 & 49.0      & 29.4        & 79.8       & 58.8     & 45.3        & 72.5    & 56.3\\ 
0.7 & 0.3 & \textbf{52.0} & 31.1 & 81.7 & 59.6 & 47.0 & 73.8 & 57.5 \\
0.3           & 0.3 & 51.5      & \textbf{32.0}        & 82.1       & \textbf{61.0}     & 46.3        & 74.6    & 57.9\\ 
\bottomrule
\end{tabular}
}
\end{table}
\subsection{Ablation Studies on Automatic Process Labeling}
We give an ablation study on how two parts of DualMath-1.1M contribute to URSA-8B-RM. As shown in Table~\ref{tab:prm_comparison_abl}, we can see that the method based on BEL, which focuses on the potential to correctness, and the method based on MIE, which focuses on the perception consistency, both contribute positively to the outcome. This further illustrates that in the process of multimodal math reasoning, image-text inconsistency is widespread and needs to be mitigated. We address this issue by augmenting the process supervision training data through the enforced imposition of common hallucination categories. Specifically, the data generated by BEL demonstrates a more significant impact, indicating that the quality of synthesized data can still be improved.
\subsection{Sensitivity Analysis on Reward Penalty and Drop-moment}
In this section, we conduct a sensitivity analysis on two hyperparameters of PS-GRPO,~$\gamma$ and~$\rho$. These respectively define the magnitude of the reward penalty for rollouts exhibiting a \enquote{drop-moment} and the tolerance threshold for identifying such \enquote{drop-moments}. As shown in Table~\ref{tab:sensitivity_analysis}, our core findings are twofold: (\romannumeral1) The value of $\gamma$ should not be set too high, as this implies excessive trust in the PRM, which may cause the rewards of a group to vanish and lead to training instability. When fixing $\rho$ at 0.3, we find that setting $\gamma$ to a value within a certain appropriate range~(we test 0.3-0.7) is generally beneficial for average performance. (\romannumeral2) An excessively large~$\rho$ diminishes reward differentiation, causing the RL behavior to approximate that of vanilla GRPO. Conversely, an excessively small~$\rho$ is unreasonable by design, as it is overly sensitive to process reward changes and tends to result in an overly broad range of penalties. In an extreme case where all correct rollouts are penalized, PS-GRPO degenerates back to vanilla GRPO.
\section{Conclusion}
In this study, we take the first step to thoroughly explore the application of PRM in multimodal math reasoning. We introduce a three-stage training pipeline URSA designed to address three major challenges. Initially, we provide a large-scale CoT reasoning dataset MMathCoT-1M. This dataset forms the basis for developing URSA-8B, a MLLM with enhanced reasoning capabilities, and paves the way for further TTS or RL scenarios. Next, we present a dual-view automated process supervision annotation method, covering logical validity and perceptual consistency in multimodal scenarios. We introduce the first large-scale process supervision dataset in multimodal reasoning, DualMath-1.1M. Finally, we address reward hacking and rewarding length bias through process-as-outcome modeling, and put forward PS-GRPO, which is a PRM-aided online RL method that surpasses GRPO. The resulting URSA-8B-PS-GRPO model demonstrates superior average performance over leading open-source MLLM such as Gemma3-12B~(8.4\%) and proprietary GPT-4o~(2.7\%). 

{
\small
\bibliographystyle{unsrtnat}
\bibliography{references}

\begin{thebibliography}{107}
\providecommand{\natexlab}[1]{#1}
\providecommand{\url}[1]{\texttt{#1}}
\expandafter\ifx\csname urlstyle\endcsname\relax
  \providecommand{\doi}[1]{doi: #1}\else
  \providecommand{\doi}{doi: \begingroup \urlstyle{rm}\Url}\fi

\bibitem[Luo et~al.(2023)Luo, Sun, Xu, Zhao, Lou, Tao, Geng, Lin, Chen, and Zhang]{luo2023wizardmath}
Haipeng Luo, Qingfeng Sun, Can Xu, Pu~Zhao, Jianguang Lou, Chongyang Tao, Xiubo Geng, Qingwei Lin, Shifeng Chen, and Dongmei Zhang.
\newblock Wizardmath: Empowering mathematical reasoning for large language models via reinforced evol-instruct.
\newblock \emph{arXiv preprint arXiv:2308.09583}, 2023.

\bibitem[Yang et~al.(2024{\natexlab{a}})Yang, Zhang, Hui, Gao, Yu, Li, Liu, Tu, Zhou, Lin, Lu, Xue, Lin, Liu, Ren, and Zhang]{yang2024qwenmath}
An~Yang, Beichen Zhang, Binyuan Hui, Bofei Gao, Bowen Yu, Chengpeng Li, Dayiheng Liu, Jianhong Tu, Jingren Zhou, Junyang Lin, Keming Lu, Mingfeng Xue, Runji Lin, Tianyu Liu, Xingzhang Ren, and Zhenru Zhang.
\newblock Qwen2.5-math technical report: Toward mathematical expert model via self-improvement.
\newblock \emph{CoRR}, abs/2409.12122, 2024{\natexlab{a}}.
\newblock \doi{10.48550/ARXIV.2409.12122}.
\newblock URL \url{https://doi.org/10.48550/arXiv.2409.12122}.

\bibitem[Ying et~al.(2024)Ying, Zhang, Li, Zhou, Shao, Fei, Ma, Hong, Liu, Wang, Wang, Wu, Li, Zhou, Liu, Zhang, Zhang, Yan, Qiu, Wang, Chen, and Lin]{ying2024internmath}
Huaiyuan Ying, Shuo Zhang, Linyang Li, Zhejian Zhou, Yunfan Shao, Zhaoye Fei, Yichuan Ma, Jiawei Hong, Kuikun Liu, Ziyi Wang, Yudong Wang, Zijian Wu, Shuaibin Li, Fengzhe Zhou, Hongwei Liu, Songyang Zhang, Wenwei Zhang, Hang Yan, Xipeng Qiu, Jiayu Wang, Kai Chen, and Dahua Lin.
\newblock Internlm-math: Open math large language models toward verifiable reasoning.
\newblock \emph{CoRR}, abs/2402.06332, 2024.
\newblock \doi{10.48550/ARXIV.2402.06332}.
\newblock URL \url{https://doi.org/10.48550/arXiv.2402.06332}.

\bibitem[Shao et~al.(2024)Shao, Wang, Zhu, Xu, Song, Bi, Zhang, Zhang, Li, Wu, et~al.]{shao2024deepseekmath}
Zhihong Shao, Peiyi Wang, Qihao Zhu, Runxin Xu, Junxiao Song, Xiao Bi, Haowei Zhang, Mingchuan Zhang, YK~Li, Y~Wu, et~al.
\newblock Deepseekmath: Pushing the limits of mathematical reasoning in open language models.
\newblock \emph{arXiv preprint arXiv:2402.03300}, 2024.

\bibitem[Yang et~al.(2024{\natexlab{b}})Yang, Chen, Du, Yu, Wang, Hong, Jiang, Xu, Dong, and Tang]{yang2024mathglm}
Zhen Yang, Jinhao Chen, Zhengxiao Du, Wenmeng Yu, Weihan Wang, Wenyi Hong, Zhihuan Jiang, Bin Xu, Yuxiao Dong, and Jie Tang.
\newblock Mathglm-vision: Solving mathematical problems with multi-modal large language model.
\newblock \emph{arXiv preprint arXiv:2409.13729}, 2024{\natexlab{b}}.

\bibitem[Yu et~al.(2024)Yu, Jiang, Shi, Yu, Liu, Zhang, Kwok, Li, Weller, and Liu]{yu2024metamath}
Longhui Yu, Weisen Jiang, Han Shi, Jincheng Yu, Zhengying Liu, Yu~Zhang, James~T. Kwok, Zhenguo Li, Adrian Weller, and Weiyang Liu.
\newblock Metamath: Bootstrap your own mathematical questions for large language models.
\newblock In \emph{The Twelfth International Conference on Learning Representations, {ICLR} 2024, Vienna, Austria, May 7-11, 2024}. OpenReview.net, 2024.
\newblock URL \url{https://openreview.net/forum?id=N8N0hgNDRt}.

\bibitem[Ni et~al.(2024)Ni, Gong, Gou, Shen, Yang, Duan, and Chen]{ni2024exploring}
Xinzhe Ni, Yeyun Gong, Zhibin Gou, Yelong Shen, Yujiu Yang, Nan Duan, and Weizhu Chen.
\newblock Exploring the mystery of influential data for mathematical reasoning.
\newblock \emph{CoRR}, abs/2404.01067, 2024.
\newblock \doi{10.48550/ARXIV.2404.01067}.
\newblock URL \url{https://doi.org/10.48550/arXiv.2404.01067}.

\bibitem[Yu et~al.(2025{\natexlab{a}})Yu, Zhang, Zhang, Liang, Zhang, Zhang, Yang, Khademi, Awadalla, Wang, Yang, and Wei]{yu2025chain}
Yiyao Yu, Yuxiang Zhang, Dongdong Zhang, Xiao Liang, Hengyuan Zhang, Xingxing Zhang, Ziyi Yang, Mahmoud Khademi, Hany Awadalla, Junjie Wang, Yujiu Yang, and Furu Wei.
\newblock Chain-of-reasoning: Towards unified mathematical reasoning in large language models via a multi-paradigm perspective.
\newblock \emph{CoRR}, abs/2501.11110, 2025{\natexlab{a}}.
\newblock \doi{10.48550/ARXIV.2501.11110}.
\newblock URL \url{https://doi.org/10.48550/arXiv.2501.11110}.

\bibitem[Yang et~al.(2024{\natexlab{c}})Yang, Chen, Du, Yu, Wang, Hong, Jiang, Xu, Dong, and Tang]{yang2024mathglmv}
Zhen Yang, Jinhao Chen, Zhengxiao Du, Wenmeng Yu, Weihan Wang, Wenyi Hong, Zhihuan Jiang, Bin Xu, Yuxiao Dong, and Jie Tang.
\newblock Mathglm-vision: Solving mathematical problems with multi-modal large language model.
\newblock \emph{CoRR}, abs/2409.13729, 2024{\natexlab{c}}.
\newblock \doi{10.48550/ARXIV.2409.13729}.
\newblock URL \url{https://doi.org/10.48550/arXiv.2409.13729}.

\bibitem[Yao et~al.(2024{\natexlab{a}})Yao, Huang, Wu, Zhang, Wang, Liu, Wang, Song, Feng, Shen, and Tao]{DBLP:journals/corr/abs-2412-18319}
Huanjin Yao, Jiaxing Huang, Wenhao Wu, Jingyi Zhang, Yibo Wang, Shunyu Liu, Yingjie Wang, Yuxin Song, Haocheng Feng, Li~Shen, and Dacheng Tao.
\newblock Mulberry: Empowering {MLLM} with o1-like reasoning and reflection via collective monte carlo tree search.
\newblock \emph{CoRR}, abs/2412.18319, 2024{\natexlab{a}}.
\newblock \doi{10.48550/ARXIV.2412.18319}.
\newblock URL \url{https://doi.org/10.48550/arXiv.2412.18319}.

\bibitem[Zhuang et~al.(2025)Zhuang, Huang, Zhang, and Zeng]{zhuang2025math}
Wenwen Zhuang, Xin Huang, Xiantao Zhang, and Jin Zeng.
\newblock Math-puma: Progressive upward multimodal alignment to enhance mathematical reasoning.
\newblock In \emph{Proceedings of the AAAI Conference on Artificial Intelligence}, volume~39, pages 26183--26191, 2025.

\bibitem[Xiang et~al.(2024)Xiang, Liu, Jiang, Nie, Huang, Fan, Li, Huang, Zeng, Han, et~al.]{xiang2024atomthink}
Kun Xiang, Zhili Liu, Zihao Jiang, Yunshuang Nie, Runhui Huang, Haoxiang Fan, Hanhui Li, Weiran Huang, Yihan Zeng, Jianhua Han, et~al.
\newblock Atomthink: A slow thinking framework for multimodal mathematical reasoning.
\newblock \emph{arXiv preprint arXiv:2411.11930}, 2024.

\bibitem[Huang et~al.(2025{\natexlab{a}})Huang, Jia, Zhai, Cao, Ye, Zhao, Xu, Hu, and Lin]{huang25visionr1}
Wenxuan Huang, Bohan Jia, Zijie Zhai, Shaosheng Cao, Zheyu Ye, Fei Zhao, Zhe Xu, Yao Hu, and Shaohui Lin.
\newblock Vision-r1: Incentivizing reasoning capability in multimodal large language models.
\newblock \emph{CoRR}, abs/2503.06749, 2025{\natexlab{a}}.
\newblock \doi{10.48550/ARXIV.2503.06749}.
\newblock URL \url{https://doi.org/10.48550/arXiv.2503.06749}.

\bibitem[Han et~al.(2024)Han, Jian, Hu, Liu, Wang, Fan, Ai, Huang, He, Yang, et~al.]{han24infimm}
Xiaotian Han, Yiren Jian, Xuefeng Hu, Haogeng Liu, Yiqi Wang, Qihang Fan, Yuang Ai, Huaibo Huang, Ran He, Zhenheng Yang, et~al.
\newblock Infimm-webmath-40b: Advancing multimodal pre-training for enhanced mathematical reasoning.
\newblock In \emph{The 4th Workshop on Mathematical Reasoning and AI at NeurIPS'24}, 2024.

\bibitem[Shi et~al.(2024)Shi, Hu, Bin, Liu, Yang, Ng, Bing, and Lee]{shi2024math}
Wenhao Shi, Zhiqiang Hu, Yi~Bin, Junhua Liu, Yang Yang, See-Kiong Ng, Lidong Bing, and Roy Ka-Wei Lee.
\newblock Math-llava: Bootstrapping mathematical reasoning for multimodal large language models.
\newblock \emph{arXiv preprint arXiv:2406.17294}, 2024.

\bibitem[Cai et~al.(2024)Cai, Bao, Guo, Zhang, Song, and Zheng]{cai2024geogpt4v}
Shihao Cai, Keqin Bao, Hangyu Guo, Jizhi Zhang, Jun Song, and Bo~Zheng.
\newblock Geogpt4v: Towards geometric multi-modal large language models with geometric image generation.
\newblock \emph{arXiv preprint arXiv:2406.11503}, 2024.

\bibitem[Deng et~al.(2024)Deng, Liu, Li, Luo, Wu, Zhang, Lyu, Zhang, Zhang, Ding, et~al.]{deng2024r}
Linger Deng, Yuliang Liu, Bohan Li, Dongliang Luo, Liang Wu, Chengquan Zhang, Pengyuan Lyu, Ziyang Zhang, Gang Zhang, Errui Ding, et~al.
\newblock R-cot: Reverse chain-of-thought problem generation for geometric reasoning in large multimodal models.
\newblock \emph{arXiv preprint arXiv:2410.17885}, 2024.

\bibitem[Gao et~al.(2023{\natexlab{a}})Gao, Pi, Zhang, Ye, Zhong, Wang, Hong, Han, Xu, Li, and Kong]{DBLP:journals/corr/abs-2312-11370}
Jiahui Gao, Renjie Pi, Jipeng Zhang, Jiacheng Ye, Wanjun Zhong, Yufei Wang, Lanqing Hong, Jianhua Han, Hang Xu, Zhenguo Li, and Lingpeng Kong.
\newblock G-llava: Solving geometric problem with multi-modal large language model.
\newblock \emph{CoRR}, abs/2312.11370, 2023{\natexlab{a}}.
\newblock \doi{10.48550/ARXIV.2312.11370}.
\newblock URL \url{https://doi.org/10.48550/arXiv.2312.11370}.

\bibitem[Zhang et~al.(2024{\natexlab{a}})Zhang, Wei, Jiang, Guo, Li, Zhang, Tong, Liu, Zhou, Wei, et~al.]{zhang2024mavis}
Renrui Zhang, Xinyu Wei, Dongzhi Jiang, Ziyu Guo, Shicheng Li, Yichi Zhang, Chengzhuo Tong, Jiaming Liu, Aojun Zhou, Bin Wei, et~al.
\newblock Mavis: Mathematical visual instruction tuning with an automatic data engine.
\newblock \emph{arXiv preprint arXiv:2407.08739}, 2024{\natexlab{a}}.

\bibitem[Xia et~al.(2024{\natexlab{a}})Xia, Li, Ye, Wu, Zhou, Yuan, Peng, Cai, Yan, Wang, He, Shi, Chen, Yan, and Zhang]{xia24geox}
Renqiu Xia, Mingsheng Li, Hancheng Ye, Wenjie Wu, Hongbin Zhou, Jiakang Yuan, Tianshuo Peng, Xinyu Cai, Xiangchao Yan, Bin Wang, Conghui He, Botian Shi, Tao Chen, Junchi Yan, and Bo~Zhang.
\newblock Geox: Geometric problem solving through unified formalized vision-language pre-training.
\newblock \emph{CoRR}, abs/2412.11863, 2024{\natexlab{a}}.
\newblock \doi{10.48550/ARXIV.2412.11863}.
\newblock URL \url{https://doi.org/10.48550/arXiv.2412.11863}.

\bibitem[Xia et~al.(2024{\natexlab{b}})Xia, Zhang, Ye, Yan, Liu, Zhou, Chen, Dou, Shi, Yan, and Qiao]{xia24charx}
Renqiu Xia, Bo~Zhang, Hancheng Ye, Xiangchao Yan, Qi~Liu, Hongbin Zhou, Zijun Chen, Min Dou, Botian Shi, Junchi Yan, and Yu~Qiao.
\newblock Chartx {\&} chartvlm: {A} versatile benchmark and foundation model for complicated chart reasoning.
\newblock \emph{CoRR}, abs/2402.12185, 2024{\natexlab{b}}.
\newblock \doi{10.48550/ARXIV.2402.12185}.
\newblock URL \url{https://doi.org/10.48550/arXiv.2402.12185}.

\bibitem[Zhang et~al.(2024{\natexlab{b}})Zhang, Zhang, Li, Zhao, Karypis, and Smola]{zhang24cot}
Zhuosheng Zhang, Aston Zhang, Mu~Li, Hai Zhao, George Karypis, and Alex Smola.
\newblock Multimodal chain-of-thought reasoning in language models.
\newblock \emph{Trans. Mach. Learn. Res.}, 2024, 2024{\natexlab{b}}.
\newblock URL \url{https://openreview.net/forum?id=y1pPWFVfvR}.

\bibitem[Peng et~al.(2024)Peng, Fu, Gao, Zhong, Fu, and Tang]{peng2024multimath}
Shuai Peng, Di~Fu, Liangcai Gao, Xiuqin Zhong, Hongguang Fu, and Zhi Tang.
\newblock Multimath: Bridging visual and mathematical reasoning for large language models.
\newblock \emph{arXiv preprint arXiv:2409.00147}, 2024.

\bibitem[Zhang et~al.(2024{\natexlab{c}})Zhang, Zhang, Li, Zhang, Sun, Gan, Yang, Pang, and Yang]{zhang2024improve}
Ruohong Zhang, Bowen Zhang, Yanghao Li, Haotian Zhang, Zhiqing Sun, Zhe Gan, Yinfei Yang, Ruoming Pang, and Yiming Yang.
\newblock Improve vision language model chain-of-thought reasoning.
\newblock \emph{arXiv preprint arXiv:2410.16198}, 2024{\natexlab{c}}.

\bibitem[Liu et~al.(2025{\natexlab{a}})Liu, Gao, Zhao, Zhang, Li, Qi, Ouyang, and Zhou]{liu2025can}
Runze Liu, Junqi Gao, Jian Zhao, Kaiyan Zhang, Xiu Li, Biqing Qi, Wanli Ouyang, and Bowen Zhou.
\newblock Can 1b {LLM} surpass 405b llm? rethinking compute-optimal test-time scaling.
\newblock \emph{CoRR}, abs/2502.06703, 2025{\natexlab{a}}.
\newblock \doi{10.48550/ARXIV.2502.06703}.
\newblock URL \url{https://doi.org/10.48550/arXiv.2502.06703}.

\bibitem[Zhang et~al.(2024{\natexlab{d}})Zhang, Hosseini, Bansal, Kazemi, Kumar, and Agarwal]{zhang2024generative}
Lunjun Zhang, Arian Hosseini, Hritik Bansal, Mehran Kazemi, Aviral Kumar, and Rishabh Agarwal.
\newblock Generative verifiers: Reward modeling as next-token prediction.
\newblock \emph{arXiv preprint arXiv:2408.15240}, 2024{\natexlab{d}}.

\bibitem[Zhang et~al.(2024{\natexlab{e}})Zhang, Zhoubian, Hu, Yue, Dong, and Tang]{zhang2024rest}
Dan Zhang, Sining Zhoubian, Ziniu Hu, Yisong Yue, Yuxiao Dong, and Jie Tang.
\newblock Rest-mcts*: {LLM} self-training via process reward guided tree search.
\newblock In Amir Globersons, Lester Mackey, Danielle Belgrave, Angela Fan, Ulrich Paquet, Jakub~M. Tomczak, and Cheng Zhang, editors, \emph{Advances in Neural Information Processing Systems 38: Annual Conference on Neural Information Processing Systems 2024, NeurIPS 2024, Vancouver, BC, Canada, December 10 - 15, 2024}, 2024{\natexlab{e}}.
\newblock URL \url{http://papers.nips.cc/paper\_files/paper/2024/hash/76ec4dc30e9faaf0e4b6093eaa377218-Abstract-Conference.html}.

\bibitem[Liu et~al.(2024{\natexlab{a}})Liu, Li, Zhang, Zhou, Cheng, and He]{liu2024diving}
Wei Liu, Junlong Li, Xiwen Zhang, Fan Zhou, Yu~Cheng, and Junxian He.
\newblock Diving into self-evolving training for multimodal reasoning.
\newblock \emph{CoRR}, abs/2412.17451, 2024{\natexlab{a}}.
\newblock \doi{10.48550/ARXIV.2412.17451}.
\newblock URL \url{https://doi.org/10.48550/arXiv.2412.17451}.

\bibitem[Yue et~al.(2025)Yue, Chen, Lu, Zhao, Wang, Yue, Song, and Huang]{yue2025limit-of-rlvr}
Yang Yue, Zhiqi Chen, Rui Lu, Andrew Zhao, Zhaokai Wang, Yang Yue, Shiji Song, and Gao Huang.
\newblock Does reinforcement learning really incentivize reasoning capacity in llms beyond the base model?
\newblock \emph{arXiv preprint arXiv:2504.13837}, 2025.

\bibitem[Yan et~al.(2024)Yan, Wang, Huo, Li, Li, Su, Gao, Zhang, Xu, Chu, et~al.]{yan2024errorradar}
Yibo Yan, Shen Wang, Jiahao Huo, Hang Li, Boyan Li, Jiamin Su, Xiong Gao, Yi-Fan Zhang, Tianlong Xu, Zhendong Chu, et~al.
\newblock Errorradar: Benchmarking complex mathematical reasoning of multimodal large language models via error detection.
\newblock \emph{arXiv preprint arXiv:2410.04509}, 2024.

\bibitem[Zhang et~al.(2024{\natexlab{f}})Zhang, Lei, Li, Wang, Liu, Yang, Li, Wang, Yang, Wu, et~al.]{zhang2024critic}
Di~Zhang, Jingdi Lei, Junxian Li, Xunzhi Wang, Yujie Liu, Zonglin Yang, Jiatong Li, Weida Wang, Suorong Yang, Jianbo Wu, et~al.
\newblock Critic-v: Vlm critics help catch vlm errors in multimodal reasoning.
\newblock \emph{arXiv preprint arXiv:2411.18203}, 2024{\natexlab{f}}.

\bibitem[Ai et~al.(2025)Ai, Zhou, Xu, Li, Zhang, Li, Sun, Feng, Huang, Wang, and Zhang]{ai2025projudge}
Jiaxin Ai, Pengfei Zhou, Zhaopan Xu, Ming Li, Fanrui Zhang, Zizhen Li, Jianwen Sun, Yukang Feng, Baojin Huang, Zhongyuan Wang, and Kaipeng Zhang.
\newblock Projudge: {A} multi-modal multi-discipline benchmark and instruction-tuning dataset for mllm-based process judges.
\newblock \emph{CoRR}, abs/2503.06553, 2025.
\newblock \doi{10.48550/ARXIV.2503.06553}.
\newblock URL \url{https://doi.org/10.48550/arXiv.2503.06553}.

\bibitem[Weng(2024)]{weng2024rewardhacking}
Lilian Weng.
\newblock Reward hacking and how to mitigate it.
\newblock \url{https://lilianweng.github.io/posts/2024-11-28-reward-hacking/}, November 2024.
\newblock [Accessed 11-28-2024].

\bibitem[Fu et~al.(2025)Fu, Zhao, Yao, Wang, Han, and Xiao]{fu2025reward}
Jiayi Fu, Xuandong Zhao, Chengyuan Yao, Heng Wang, Qi~Han, and Yanghua Xiao.
\newblock Reward shaping to mitigate reward hacking in {RLHF}.
\newblock \emph{CoRR}, abs/2502.18770, 2025.
\newblock \doi{10.48550/ARXIV.2502.18770}.
\newblock URL \url{https://doi.org/10.48550/arXiv.2502.18770}.

\bibitem[Li et~al.(2024)Li, Zhang, Guo, Zhang, Li, Zhang, Zhang, Zhang, Li, Liu, et~al.]{li2024llava}
Bo~Li, Yuanhan Zhang, Dong Guo, Renrui Zhang, Feng Li, Hao Zhang, Kaichen Zhang, Peiyuan Zhang, Yanwei Li, Ziwei Liu, et~al.
\newblock Llava-onevision: Easy visual task transfer.
\newblock \emph{arXiv preprint arXiv:2408.03326}, 2024.

\bibitem[Zhai et~al.(2023)Zhai, Mustafa, Kolesnikov, and Beyer]{zhai2023sigmoid}
Xiaohua Zhai, Basil Mustafa, Alexander Kolesnikov, and Lucas Beyer.
\newblock Sigmoid loss for language image pre-training.
\newblock In \emph{Proceedings of the IEEE/CVF International Conference on Computer Vision}, pages 11975--11986, 2023.

\bibitem[Kirillov et~al.(2023)Kirillov, Mintun, Ravi, Mao, Rolland, Gustafson, Xiao, Whitehead, Berg, Lo, et~al.]{kirillov2023segment}
Alexander Kirillov, Eric Mintun, Nikhila Ravi, Hanzi Mao, Chloe Rolland, Laura Gustafson, Tete Xiao, Spencer Whitehead, Alexander~C Berg, Wan-Yen Lo, et~al.
\newblock Segment anything.
\newblock In \emph{Proceedings of the IEEE/CVF International Conference on Computer Vision}, pages 4015--4026, 2023.

\bibitem[Lu et~al.(2024)Lu, Liu, Zhang, Wang, Dong, Liu, Sun, Ren, Li, Yang, et~al.]{lu2024deepseek}
Haoyu Lu, Wen Liu, Bo~Zhang, Bingxuan Wang, Kai Dong, Bo~Liu, Jingxiang Sun, Tongzheng Ren, Zhuoshu Li, Hao Yang, et~al.
\newblock Deepseek-vl: towards real-world vision-language understanding.
\newblock \emph{arXiv preprint arXiv:2403.05525}, 2024.

\bibitem[Trinh et~al.(2024)Trinh, Wu, Le, He, and Luong]{trinh2024solving}
Trieu~H Trinh, Yuhuai Wu, Quoc~V Le, He~He, and Thang Luong.
\newblock Solving olympiad geometry without human demonstrations.
\newblock \emph{Nature}, 625\penalty0 (7995):\penalty0 476--482, 2024.

\bibitem[Chen et~al.(2021)Chen, Tang, Qin, Liang, Liu, Xing, and Lin]{chen2021geoqa}
Jiaqi Chen, Jianheng Tang, Jinghui Qin, Xiaodan Liang, Lingbo Liu, Eric Xing, and Liang Lin.
\newblock Geoqa: A geometric question answering benchmark towards multimodal numerical reasoning.
\newblock In \emph{Findings of the Association for Computational Linguistics: ACL-IJCNLP 2021}, pages 513--523, 2021.

\bibitem[Lightman et~al.(2023)Lightman, Kosaraju, Burda, Edwards, Baker, Lee, Leike, Schulman, Sutskever, and Cobbe]{lightman2023let}
Hunter Lightman, Vineet Kosaraju, Yura Burda, Harri Edwards, Bowen Baker, Teddy Lee, Jan Leike, John Schulman, Ilya Sutskever, and Karl Cobbe.
\newblock Let's verify step by step.
\newblock \emph{arXiv preprint arXiv:2305.20050}, 2023.

\bibitem[Luo et~al.(2024)Luo, Liu, Liu, Phatale, Lara, Li, Shu, Zhu, Meng, Sun, et~al.]{luo2024improve}
Liangchen Luo, Yinxiao Liu, Rosanne Liu, Samrat Phatale, Harsh Lara, Yunxuan Li, Lei Shu, Yun Zhu, Lei Meng, Jiao Sun, et~al.
\newblock Improve mathematical reasoning in language models by automated process supervision.
\newblock \emph{arXiv preprint arXiv:2406.06592}, 2024.

\bibitem[Wang et~al.(2024{\natexlab{a}})Wang, Li, Shao, Xu, Dai, Li, Chen, Wu, and Sui]{wang2024math}
Peiyi Wang, Lei Li, Zhihong Shao, Runxin Xu, Damai Dai, Yifei Li, Deli Chen, Yu~Wu, and Zhifang Sui.
\newblock Math-shepherd: Verify and reinforce llms step-by-step without human annotations.
\newblock In \emph{Proceedings of the 62nd Annual Meeting of the Association for Computational Linguistics (Volume 1: Long Papers)}, pages 9426--9439, 2024{\natexlab{a}}.

\bibitem[Zheng et~al.(2024)Zheng, Xu, Sun, Pu, Chen, and Sun]{zheng2024thinking}
Haojie Zheng, Tianyang Xu, Hanchi Sun, Shu Pu, Ruoxi Chen, and Lichao Sun.
\newblock Thinking before looking: Improving multimodal llm reasoning via mitigating visual hallucination.
\newblock \emph{arXiv preprint arXiv:2411.12591}, 2024.

\bibitem[Gao et~al.(2023{\natexlab{b}})Gao, Madaan, Zhou, Alon, Liu, Yang, Callan, and Neubig]{gao2023pal}
Luyu Gao, Aman Madaan, Shuyan Zhou, Uri Alon, Pengfei Liu, Yiming Yang, Jamie Callan, and Graham Neubig.
\newblock Pal: Program-aided language models.
\newblock In \emph{International Conference on Machine Learning}, pages 10764--10799. PMLR, 2023{\natexlab{b}}.

\bibitem[Guo et~al.(2025)Guo, Yang, Zhang, Song, Zhang, Xu, Zhu, Ma, Wang, Bi, et~al.]{guo2025deepseek}
Daya Guo, Dejian Yang, Haowei Zhang, Junxiao Song, Ruoyu Zhang, Runxin Xu, Qihao Zhu, Shirong Ma, Peiyi Wang, Xiao Bi, et~al.
\newblock Deepseek-r1: Incentivizing reasoning capability in llms via reinforcement learning.
\newblock \emph{arXiv preprint arXiv:2501.12948}, 2025.

\bibitem[Pan et~al.(2025)Pan, Liu, Wu, Liu, Zhu, Li, Chen, Ouyang, and Rueckert]{pan2025medvlm}
Jiazhen Pan, Che Liu, Junde Wu, Fenglin Liu, Jiayuan Zhu, Hongwei~Bran Li, Chen Chen, Cheng Ouyang, and Daniel Rueckert.
\newblock Medvlm-r1: Incentivizing medical reasoning capability of vision-language models (vlms) via reinforcement learning.
\newblock \emph{CoRR}, abs/2502.19634, 2025.
\newblock \doi{10.48550/ARXIV.2502.19634}.
\newblock URL \url{https://doi.org/10.48550/arXiv.2502.19634}.

\bibitem[Zhan et~al.(2025)Zhan, Zhu, Zheng, Zhao, Yang, Tang, and Wang]{zhan2025visionr1}
Yufei Zhan, Yousong Zhu, Shurong Zheng, Hongyin Zhao, Fan Yang, Ming Tang, and Jinqiao Wang.
\newblock Vision-r1: Evolving human-free alignment in large vision-language models via vision-guided reinforcement learning.
\newblock \emph{CoRR}, abs/2503.18013, 2025.
\newblock \doi{10.48550/ARXIV.2503.18013}.
\newblock URL \url{https://doi.org/10.48550/arXiv.2503.18013}.

\bibitem[Huang et~al.(2025{\natexlab{b}})Huang, Jia, Zhai, Cao, Ye, Zhao, Xu, Hu, and Lin]{huang2025visionr1}
Wenxuan Huang, Bohan Jia, Zijie Zhai, Shaosheng Cao, Zheyu Ye, Fei Zhao, Zhe Xu, Yao Hu, and Shaohui Lin.
\newblock Vision-r1: Incentivizing reasoning capability in multimodal large language models.
\newblock \emph{CoRR}, abs/2503.06749, 2025{\natexlab{b}}.
\newblock \doi{10.48550/ARXIV.2503.06749}.
\newblock URL \url{https://doi.org/10.48550/arXiv.2503.06749}.

\bibitem[Liu et~al.(2025{\natexlab{b}})Liu, Ni, Wu, Du, Dou, Wang, Pang, and Shieh]{liu2025noisyrollout}
Xiangyan Liu, Jinjie Ni, Zijian Wu, Chao Du, Longxu Dou, Haonan Wang, Tianyu Pang, and Michael~Qizhe Shieh.
\newblock Noisyrollout: Reinforcing visual reasoning with data augmentation, 2025{\natexlab{b}}.
\newblock URL \url{https://arxiv.org/abs/2504.13055}.

\bibitem[Li and Li(2024)]{li2024qvalue}
Wendi Li and Yixuan Li.
\newblock Process reward model with q-value rankings.
\newblock \emph{CoRR}, abs/2410.11287, 2024.
\newblock \doi{10.48550/ARXIV.2410.11287}.
\newblock URL \url{https://doi.org/10.48550/arXiv.2410.11287}.

\bibitem[Setlur et~al.(2024)Setlur, Nagpal, Fisch, Geng, Eisenstein, Agarwal, Agarwal, Berant, and Kumar]{amrith2024reward}
Amrith Setlur, Chirag Nagpal, Adam Fisch, Xinyang Geng, Jacob Eisenstein, Rishabh Agarwal, Alekh Agarwal, Jonathan Berant, and Aviral Kumar.
\newblock Rewarding progress: Scaling automated process verifiers for {LLM} reasoning.
\newblock \emph{CoRR}, abs/2410.08146, 2024.
\newblock \doi{10.48550/ARXIV.2410.08146}.
\newblock URL \url{https://doi.org/10.48550/arXiv.2410.08146}.

\bibitem[Ma et~al.(2025)Ma, Chen, Liu, Tian, Liu, Liu, and Luo]{ma2025slm}
Yiran Ma, Zui Chen, Tianqiao Liu, Mi~Tian, Zhuo Liu, Zitao Liu, and Weiqi Luo.
\newblock What are step-level reward models rewarding? counterintuitive findings from mcts-boosted mathematical reasoning.
\newblock In Toby Walsh, Julie Shah, and Zico Kolter, editors, \emph{AAAI-25, Sponsored by the Association for the Advancement of Artificial Intelligence, February 25 - March 4, 2025, Philadelphia, PA, {USA}}, pages 24812--24820. {AAAI} Press, 2025.
\newblock \doi{10.1609/AAAI.V39I23.34663}.
\newblock URL \url{https://doi.org/10.1609/aaai.v39i23.34663}.

\bibitem[Gao et~al.(2024{\natexlab{a}})Gao, Xu, Ye, Liu, He, Fu, Mei, Wang, and Wu]{gao2024ondesign}
Jiaxuan Gao, Shusheng Xu, Wenjie Ye, Weilin Liu, Chuyi He, Wei Fu, Zhiyu Mei, Guangju Wang, and Yi~Wu.
\newblock On designing effective {RL} reward at training time for {LLM} reasoning.
\newblock \emph{CoRR}, abs/2410.15115, 2024{\natexlab{a}}.
\newblock \doi{10.48550/ARXIV.2410.15115}.
\newblock URL \url{https://doi.org/10.48550/arXiv.2410.15115}.

\bibitem[Amodei et~al.(2016)Amodei, Olah, Steinhardt, Christiano, Schulman, and Man{\'e}]{amodei2016concrete}
Dario Amodei, Chris Olah, Jacob Steinhardt, Paul Christiano, John Schulman, and Dan Man{\'e}.
\newblock Concrete problems in ai safety.
\newblock \emph{arXiv preprint arXiv:1606.06565}, 2016.

\bibitem[Zhang and Zuo(2025)]{zhang2025grpo}
Jixiao Zhang and Chunsheng Zuo.
\newblock Grpo-lead: A difficulty-aware reinforcement learning approach for concise mathematical reasoning in language models.
\newblock \emph{arXiv preprint arXiv:2504.09696}, 2025.

\bibitem[Zhang et~al.(2025{\natexlab{a}})Zhang, Huang, Yao, Liu, Zhang, Lu, and Tao]{zhang2025r1}
Jingyi Zhang, Jiaxing Huang, Huanjin Yao, Shunyu Liu, Xikun Zhang, Shijian Lu, and Dacheng Tao.
\newblock R1-vl: Learning to reason with multimodal large language models via step-wise group relative policy optimization.
\newblock \emph{arXiv preprint arXiv:2503.12937}, 2025{\natexlab{a}}.

\bibitem[Yu et~al.(2025{\natexlab{b}})Yu, Zhang, Zhu, Yuan, Zuo, Yue, Fan, Liu, Liu, Liu, et~al.]{yu2025dapo}
Qiying Yu, Zheng Zhang, Ruofei Zhu, Yufeng Yuan, Xiaochen Zuo, Yu~Yue, Tiantian Fan, Gaohong Liu, Lingjun Liu, Xin Liu, et~al.
\newblock Dapo: An open-source llm reinforcement learning system at scale.
\newblock \emph{arXiv preprint arXiv:2503.14476}, 2025{\natexlab{b}}.

\bibitem[OpenAI(2024)]{openai2024gpt4o}
OpenAI.
\newblock {GPT-4o} system card, 2024.
\newblock URL \url{https://openai.com/research/gpt-4o-system-card}.

\bibitem[Team et~al.(2023)Team, Anil, Borgeaud, Alayrac, Yu, Soricut, Schalkwyk, Dai, Hauth, Millican, et~al.]{team2023gemini}
Gemini Team, Rohan Anil, Sebastian Borgeaud, Jean-Baptiste Alayrac, Jiahui Yu, Radu Soricut, Johan Schalkwyk, Andrew~M Dai, Anja Hauth, Katie Millican, et~al.
\newblock Gemini: a family of highly capable multimodal models.
\newblock \emph{arXiv preprint arXiv:2312.11805}, 2023.

\bibitem[Chen et~al.(2024{\natexlab{a}})Chen, Wang, Tian, Ye, Gao, Cui, Tong, Hu, Luo, Ma, et~al.]{chen2024far}
Zhe Chen, Weiyun Wang, Hao Tian, Shenglong Ye, Zhangwei Gao, Erfei Cui, Wenwen Tong, Kongzhi Hu, Jiapeng Luo, Zheng Ma, et~al.
\newblock How far are we to gpt-4v? closing the gap to commercial multimodal models with open-source suites.
\newblock \emph{arXiv preprint arXiv:2404.16821}, 2024{\natexlab{a}}.

\bibitem[Meta(2024)]{metaLlama32}
Meta.
\newblock {L}lama 3.2: {R}evolutionizing edge {A}{I} and vision with open, customizable models --- ai.meta.com.
\newblock \url{https://ai.meta.com/blog/llama-3-2-connect-2024-vision-edge-mobile-devices/}, 2024.
\newblock [Accessed 17-04-2025].

\bibitem[Wang et~al.(2024{\natexlab{b}})Wang, Bai, Tan, Wang, Fan, Bai, Chen, Liu, Wang, Ge, et~al.]{wang2024qwen2}
Peng Wang, Shuai Bai, Sinan Tan, Shijie Wang, Zhihao Fan, Jinze Bai, Keqin Chen, Xuejing Liu, Jialin Wang, Wenbin Ge, et~al.
\newblock Qwen2-vl: Enhancing vision-language model's perception of the world at any resolution.
\newblock \emph{arXiv preprint arXiv:2409.12191}, 2024{\natexlab{b}}.

\bibitem[Chen et~al.(2024{\natexlab{b}})Chen, Wu, Wang, Su, Chen, Xing, Zhong, Zhang, Zhu, Lu, et~al.]{chen2024internvl}
Zhe Chen, Jiannan Wu, Wenhai Wang, Weijie Su, Guo Chen, Sen Xing, Muyan Zhong, Qinglong Zhang, Xizhou Zhu, Lewei Lu, et~al.
\newblock Internvl: Scaling up vision foundation models and aligning for generic visual-linguistic tasks.
\newblock In \emph{Proceedings of the IEEE/CVF Conference on Computer Vision and Pattern Recognition}, pages 24185--24198, 2024{\natexlab{b}}.

\bibitem[Wang et~al.(2024{\natexlab{c}})Wang, Chen, Wang, Cao, Liu, Gao, Zhu, Zhu, Lu, Qiao, et~al.]{wang2024enhancing}
Weiyun Wang, Zhe Chen, Wenhai Wang, Yue Cao, Yangzhou Liu, Zhangwei Gao, Jinguo Zhu, Xizhou Zhu, Lewei Lu, Yu~Qiao, et~al.
\newblock Enhancing the reasoning ability of multimodal large language models via mixed preference optimization.
\newblock \emph{arXiv preprint arXiv:2411.10442}, 2024{\natexlab{c}}.

\bibitem[Chen et~al.(2024{\natexlab{c}})Chen, Wang, Cao, Liu, Gao, Cui, Zhu, Ye, Tian, Liu, et~al.]{chen2024expanding}
Zhe Chen, Weiyun Wang, Yue Cao, Yangzhou Liu, Zhangwei Gao, Erfei Cui, Jinguo Zhu, Shenglong Ye, Hao Tian, Zhaoyang Liu, et~al.
\newblock Expanding performance boundaries of open-source multimodal models with model, data, and test-time scaling.
\newblock \emph{arXiv preprint arXiv:2412.05271}, 2024{\natexlab{c}}.

\bibitem[Liu et~al.(2024{\natexlab{b}})Liu, Zhao, Zhuang, Tian, Zhou, and Zhou]{liu2024points}
Yuan Liu, Zhongyin Zhao, Ziyuan Zhuang, Le~Tian, Xiao Zhou, and Jie Zhou.
\newblock Points: Improving your vision-language model with affordable strategies.
\newblock \emph{arXiv preprint arXiv:2409.04828}, 2024{\natexlab{b}}.

\bibitem[Team et~al.(2025)Team, Kamath, Ferret, Pathak, Vieillard, Merhej, Perrin, Matejovicova, Ram{\'e}, Rivi{\`e}re, et~al.]{team2025gemma}
Gemma Team, Aishwarya Kamath, Johan Ferret, Shreya Pathak, Nino Vieillard, Ramona Merhej, Sarah Perrin, Tatiana Matejovicova, Alexandre Ram{\'e}, Morgane Rivi{\`e}re, et~al.
\newblock Gemma 3 technical report.
\newblock \emph{arXiv preprint arXiv:2503.19786}, 2025.

\bibitem[Thawakar et~al.(2025)Thawakar, Dissanayake, More, Thawkar, Heakl, Ahsan, Li, Zumri, Lahoud, Anwer, Cholakkal, Laptev, Shah, Khan, and Khan]{DBLP:journals/corr/abs-2501-06186}
Omkar Thawakar, Dinura Dissanayake, Ketan More, Ritesh Thawkar, Ahmed Heakl, Noor Ahsan, Yuhao Li, Mohammed Zumri, Jean Lahoud, Rao~Muhammad Anwer, Hisham Cholakkal, Ivan Laptev, Mubarak Shah, Fahad~Shahbaz Khan, and Salman~H. Khan.
\newblock Llamav-o1: Rethinking step-by-step visual reasoning in llms.
\newblock \emph{CoRR}, abs/2501.06186, 2025.
\newblock \doi{10.48550/ARXIV.2501.06186}.
\newblock URL \url{https://doi.org/10.48550/arXiv.2501.06186}.

\bibitem[Deng et~al.(2025)Deng, Bansal, Yin, Peng, Wang, and Chang]{deng25openvlthiner}
Yihe Deng, Hritik Bansal, Fan Yin, Nanyun Peng, Wei Wang, and Kai{-}Wei Chang.
\newblock Openvlthinker: An early exploration to complex vision-language reasoning via iterative self-improvement.
\newblock \emph{CoRR}, abs/2503.17352, 2025.
\newblock \doi{10.48550/ARXIV.2503.17352}.
\newblock URL \url{https://doi.org/10.48550/arXiv.2503.17352}.

\bibitem[Yang et~al.(2025)Yang, He, Pan, Jiang, Deng, Yang, Lu, Yin, Rao, Zhu, Zhang, and Chen]{yang2025r1onevision}
Yi~Yang, Xiaoxuan He, Hongkun Pan, Xiyan Jiang, Yan Deng, Xingtao Yang, Haoyu Lu, Dacheng Yin, Fengyun Rao, Minfeng Zhu, Bo~Zhang, and Wei Chen.
\newblock R1-onevision: Advancing generalized multimodal reasoning through cross-modal formalization.
\newblock \emph{arXiv preprint arXiv:2503.10615}, 2025.

\bibitem[Zhang et~al.(2025{\natexlab{b}})Zhang, Jiang, Zhang, Lin, Guo, Qiu, Zhou, Lu, Chang, Qiao, et~al.]{zhang2025mathverse}
Renrui Zhang, Dongzhi Jiang, Yichi Zhang, Haokun Lin, Ziyu Guo, Pengshuo Qiu, Aojun Zhou, Pan Lu, Kai-Wei Chang, Yu~Qiao, et~al.
\newblock Mathverse: Does your multi-modal llm truly see the diagrams in visual math problems?
\newblock In \emph{European Conference on Computer Vision}, pages 169--186. Springer, 2025{\natexlab{b}}.

\bibitem[Zou et~al.(2024)Zou, Guo, Yang, Zhang, Hu, and Zhang]{zou2024dynamath}
Chengke Zou, Xingang Guo, Rui Yang, Junyu Zhang, Bin Hu, and Huan Zhang.
\newblock Dynamath: A dynamic visual benchmark for evaluating mathematical reasoning robustness of vision language models.
\newblock \emph{arXiv preprint arXiv:2411.00836}, 2024.

\bibitem[Lu et~al.(2023{\natexlab{a}})Lu, Bansal, Xia, Liu, Li, Hajishirzi, Cheng, Chang, Galley, and Gao]{lu2023mathvista}
Pan Lu, Hritik Bansal, Tony Xia, Jiacheng Liu, Chunyuan Li, Hannaneh Hajishirzi, Hao Cheng, Kai-Wei Chang, Michel Galley, and Jianfeng Gao.
\newblock Mathvista: Evaluating mathematical reasoning of foundation models in visual contexts.
\newblock \emph{arXiv preprint arXiv:2310.02255}, 2023{\natexlab{a}}.

\bibitem[Qiao et~al.(2024)Qiao, Tan, Dong, Wu, Sun, Song, GongQue, Lei, Wei, Zhang, et~al.]{qiao2024we}
Runqi Qiao, Qiuna Tan, Guanting Dong, Minhui Wu, Chong Sun, Xiaoshuai Song, Zhuoma GongQue, Shanglin Lei, Zhe Wei, Miaoxuan Zhang, et~al.
\newblock We-math: Does your large multimodal model achieve human-like mathematical reasoning?
\newblock \emph{arXiv preprint arXiv:2407.01284}, 2024.

\bibitem[Dong et~al.(2024{\natexlab{a}})Dong, Zhang, Zang, Cao, Wang, Ouyang, Wei, Zhang, Duan, Cao, et~al.]{dong2024internlm}
Xiaoyi Dong, Pan Zhang, Yuhang Zang, Yuhang Cao, Bin Wang, Linke Ouyang, Xilin Wei, Songyang Zhang, Haodong Duan, Maosong Cao, et~al.
\newblock Internlm-xcomposer2: Mastering free-form text-image composition and comprehension in vision-language large model.
\newblock \emph{arXiv preprint arXiv:2401.16420}, 2024{\natexlab{a}}.

\bibitem[Yao et~al.(2024{\natexlab{b}})Yao, Huang, Wu, Zhang, Wang, Liu, Wang, Song, Feng, Shen, and Tao]{yao2024mulberry}
Huanjin Yao, Jiaxing Huang, Wenhao Wu, Jingyi Zhang, Yibo Wang, Shunyu Liu, Yingjie Wang, Yuxin Song, Haocheng Feng, Li~Shen, and Dacheng Tao.
\newblock Mulberry: Empowering {MLLM} with o1-like reasoning and reflection via collective monte carlo tree search.
\newblock \emph{CoRR}, abs/2412.18319, 2024{\natexlab{b}}.
\newblock \doi{10.48550/ARXIV.2412.18319}.
\newblock URL \url{https://doi.org/10.48550/arXiv.2412.18319}.

\bibitem[Guo et~al.(2024)Guo, Zheng, Bai, Li, Wang, Zhu, Li, Neubig, Chen, and Yue]{guo2024mammoth}
Jarvis Guo, Tuney Zheng, Yuelin Bai, Bo~Li, Yubo Wang, King Zhu, Yizhi Li, Graham Neubig, Wenhu Chen, and Xiang Yue.
\newblock Mammoth-vl: Eliciting multimodal reasoning with instruction tuning at scale.
\newblock \emph{arXiv preprint arXiv:2412.05237}, 2024.

\bibitem[Wang et~al.(2022)Wang, Wei, Schuurmans, Le, Chi, Narang, Chowdhery, and Zhou]{wang2022self}
Xuezhi Wang, Jason Wei, Dale Schuurmans, Quoc Le, Ed~Chi, Sharan Narang, Aakanksha Chowdhery, and Denny Zhou.
\newblock Self-consistency improves chain of thought reasoning in language models.
\newblock \emph{arXiv preprint arXiv:2203.11171}, 2022.

\bibitem[Gao et~al.(2023{\natexlab{c}})Gao, Pi, Zhang, Ye, Zhong, Wang, Hong, Han, Xu, Li, et~al.]{gao2023g}
Jiahui Gao, Renjie Pi, Jipeng Zhang, Jiacheng Ye, Wanjun Zhong, Yufei Wang, Lanqing Hong, Jianhua Han, Hang Xu, Zhenguo Li, et~al.
\newblock G-llava: Solving geometric problem with multi-modal large language model.
\newblock \emph{arXiv preprint arXiv:2312.11370}, 2023{\natexlab{c}}.

\bibitem[Dong et~al.(2024{\natexlab{b}})Dong, Liu, Sun, Yang, Hu, Rao, and Liu]{dong2024insight}
Yuhao Dong, Zuyan Liu, Hai-Long Sun, Jingkang Yang, Winston Hu, Yongming Rao, and Ziwei Liu.
\newblock Insight-v: Exploring long-chain visual reasoning with multimodal large language models.
\newblock \emph{arXiv preprint arXiv:2411.14432}, 2024{\natexlab{b}}.

\bibitem[Hu et~al.(2024)Hu, Shi, Fu, Roth, Ostendorf, Zettlemoyer, Smith, and Krishna]{hu2024visual}
Yushi Hu, Weijia Shi, Xingyu Fu, Dan Roth, Mari Ostendorf, Luke Zettlemoyer, Noah~A Smith, and Ranjay Krishna.
\newblock Visual sketchpad: Sketching as a visual chain of thought for multimodal language models.
\newblock \emph{arXiv preprint arXiv:2406.09403}, 2024.

\bibitem[Yu et~al.(2023)Yu, Jiang, Shi, Yu, Liu, Zhang, Kwok, Li, Weller, and Liu]{yu2023metamath}
Longhui Yu, Weisen Jiang, Han Shi, Jincheng Yu, Zhengying Liu, Yu~Zhang, James~T Kwok, Zhenguo Li, Adrian Weller, and Weiyang Liu.
\newblock Metamath: Bootstrap your own mathematical questions for large language models.
\newblock \emph{arXiv preprint arXiv:2309.12284}, 2023.

\bibitem[Liu et~al.(2024{\natexlab{c}})Liu, Zhang, Qiu, Huang, Lin, Zhao, Geng, Lin, Jin, Zhang, et~al.]{liu2024sphinx}
Dongyang Liu, Renrui Zhang, Longtian Qiu, Siyuan Huang, Weifeng Lin, Shitian Zhao, Shijie Geng, Ziyi Lin, Peng Jin, Kaipeng Zhang, et~al.
\newblock Sphinx-x: Scaling data and parameters for a family of multi-modal large language models.
\newblock \emph{arXiv preprint arXiv:2402.05935}, 2024{\natexlab{c}}.

\bibitem[Sprague et~al.(2024)Sprague, Yin, Rodriguez, Jiang, Wadhwa, Singhal, Zhao, Ye, Mahowald, and Durrett]{sprague2024cot}
Zayne Sprague, Fangcong Yin, Juan~Diego Rodriguez, Dongwei Jiang, Manya Wadhwa, Prasann Singhal, Xinyu Zhao, Xi~Ye, Kyle Mahowald, and Greg Durrett.
\newblock To cot or not to cot? chain-of-thought helps mainly on math and symbolic reasoning.
\newblock \emph{arXiv preprint arXiv:2409.12183}, 2024.

\bibitem[Lu et~al.(2023{\natexlab{b}})Lu, Shen, Wang, Wang, van Rechem, Fu, and Wei]{lu2023machine}
Yingzhou Lu, Minjie Shen, Huazheng Wang, Xiao Wang, Capucine van Rechem, Tianfan Fu, and Wenqi Wei.
\newblock Machine learning for synthetic data generation: a review.
\newblock \emph{arXiv preprint arXiv:2302.04062}, 2023{\natexlab{b}}.

\bibitem[Huang et~al.(2024)Huang, Liu, Gong, Gou, Shen, Duan, and Chen]{huang2024key}
Yiming Huang, Xiao Liu, Yeyun Gong, Zhibin Gou, Yelong Shen, Nan Duan, and Weizhu Chen.
\newblock Key-point-driven data synthesis with its enhancement on mathematical reasoning.
\newblock \emph{arXiv preprint arXiv:2403.02333}, 2024.

\bibitem[Fu et~al.(2024)Fu, Lin, Long, Shen, Zhao, Zhang, Dong, Wang, Yin, Ma, et~al.]{fu2024vita}
Chaoyou Fu, Haojia Lin, Zuwei Long, Yunhang Shen, Meng Zhao, Yifan Zhang, Shaoqi Dong, Xiong Wang, Di~Yin, Long Ma, et~al.
\newblock Vita: Towards open-source interactive omni multimodal llm.
\newblock \emph{arXiv preprint arXiv:2408.05211}, 2024.

\bibitem[Gou et~al.(2023)Gou, Shao, Gong, Shen, Yang, Duan, and Chen]{gou2023critic}
Zhibin Gou, Zhihong Shao, Yeyun Gong, Yelong Shen, Yujiu Yang, Nan Duan, and Weizhu Chen.
\newblock Critic: Large language models can self-correct with tool-interactive critiquing.
\newblock \emph{arXiv preprint arXiv:2305.11738}, 2023.

\bibitem[Gao et~al.(2024{\natexlab{b}})Gao, Cai, Xu, Wang, Zheng, Lin, Lu, Lin, Zhou, Xiao, et~al.]{gao2024llm}
Bofei Gao, Zefan Cai, Runxin Xu, Peiyi Wang, Ce~Zheng, Runji Lin, Keming Lu, Junyang Lin, Chang Zhou, Wen Xiao, et~al.
\newblock Llm critics help catch bugs in mathematics: Towards a better mathematical verifier with natural language feedback.
\newblock \emph{CoRR}, 2024{\natexlab{b}}.

\bibitem[Lin et~al.(2024)Lin, Gou, Liang, Luo, Liu, and Yang]{lin-etal-2024-criticbench}
Zicheng Lin, Zhibin Gou, Tian Liang, Ruilin Luo, Haowei Liu, and Yujiu Yang.
\newblock {C}ritic{B}ench: Benchmarking {LLM}s for critique-correct reasoning.
\newblock In Lun-Wei Ku, Andre Martins, and Vivek Srikumar, editors, \emph{Findings of the Association for Computational Linguistics: ACL 2024}, pages 1552--1587, Bangkok, Thailand, August 2024. Association for Computational Linguistics.
\newblock \doi{10.18653/v1/2024.findings-acl.91}.
\newblock URL \url{https://aclanthology.org/2024.findings-acl.91}.

\bibitem[Kumar et~al.(2024)Kumar, Zhuang, Agarwal, Su, Co-Reyes, Singh, Baumli, Iqbal, Bishop, Roelofs, et~al.]{kumar2024training}
Aviral Kumar, Vincent Zhuang, Rishabh Agarwal, Yi~Su, John~D Co-Reyes, Avi Singh, Kate Baumli, Shariq Iqbal, Colton Bishop, Rebecca Roelofs, et~al.
\newblock Training language models to self-correct via reinforcement learning.
\newblock \emph{arXiv preprint arXiv:2409.12917}, 2024.

\bibitem[Snell et~al.(2024)Snell, Lee, Xu, and Kumar]{snell2024scaling}
Charlie Snell, Jaehoon Lee, Kelvin Xu, and Aviral Kumar.
\newblock Scaling llm test-time compute optimally can be more effective than scaling model parameters.
\newblock \emph{arXiv preprint arXiv:2408.03314}, 2024.

\bibitem[Tu et~al.(2025)Tu, Feng, Chen, Liu, Tang, and Xie]{tu2025vilbench}
Haoqin Tu, Weitao Feng, Hardy Chen, Hui Liu, Xianfeng Tang, and Cihang Xie.
\newblock Vilbench: A suite for vision-language process reward modeling.
\newblock \emph{arXiv preprint arXiv:2503.20271}, 2025.

\bibitem[Wang et~al.(2025)Wang, Gao, Chen, Chen, Zhu, Zhao, Liu, Cao, Ye, Zhu, et~al.]{wang2025visualprm}
Weiyun Wang, Zhangwei Gao, Lianjie Chen, Zhe Chen, Jinguo Zhu, Xiangyu Zhao, Yangzhou Liu, Yue Cao, Shenglong Ye, Xizhou Zhu, et~al.
\newblock Visualprm: An effective process reward model for multimodal reasoning.
\newblock \emph{arXiv preprint arXiv:2503.10291}, 2025.

\bibitem[Sun et~al.(2025)Sun, Liang, Wei, Yu, Li, Yang, Zhou, and Zhang]{sun2025mm}
Linzhuang Sun, Hao Liang, Jingxuan Wei, Bihui Yu, Tianpeng Li, Fan Yang, Zenan Zhou, and Wentao Zhang.
\newblock Mm-verify: Enhancing multimodal reasoning with chain-of-thought verification.
\newblock \emph{arXiv preprint arXiv:2502.13383}, 2025.

\bibitem[Gao et~al.(2024{\natexlab{c}})Gao, Cai, Xu, Wang, Zheng, Lin, Lu, Lin, Zhou, Xiao, Hu, Liu, and Chang]{gao2024critic}
Bofei Gao, Zefan Cai, Runxin Xu, Peiyi Wang, Ce~Zheng, Runji Lin, Keming Lu, Junyang Lin, Chang Zhou, Wen Xiao, Junjie Hu, Tianyu Liu, and Baobao Chang.
\newblock {LLM} critics help catch bugs in mathematics: Towards a better mathematical verifier with natural language feedback.
\newblock \emph{CoRR}, abs/2406.14024, 2024{\natexlab{c}}.
\newblock \doi{10.48550/ARXIV.2406.14024}.
\newblock URL \url{https://doi.org/10.48550/arXiv.2406.14024}.

\bibitem[Zeng et~al.(2024)Zeng, Huang, Zhao, Wang, Shan, and He]{zeng2024b}
Weihao Zeng, Yuzhen Huang, Lulu Zhao, Yijun Wang, Zifei Shan, and Junxian He.
\newblock B-star: Monitoring and balancing exploration and exploitation in self-taught reasoners.
\newblock \emph{arXiv preprint arXiv:2412.17256}, 2024.

\bibitem[Meng et~al.(2025)Meng, Du, Liu, Zhou, Lu, Fu, Shi, Wang, He, Zhang, et~al.]{meng2025mm}
Fanqing Meng, Lingxiao Du, Zongkai Liu, Zhixiang Zhou, Quanfeng Lu, Daocheng Fu, Botian Shi, Wenhai Wang, Junjun He, Kaipeng Zhang, et~al.
\newblock Mm-eureka: Exploring visual aha moment with rule-based large-scale reinforcement learning.
\newblock \emph{arXiv preprint arXiv:2503.07365}, 2025.

\bibitem[von Werra et~al.(2020)von Werra, Belkada, Tunstall, Beeching, Thrush, Lambert, Huang, Rasul, and Gallouédec]{vonwerra2022trl}
Leandro von Werra, Younes Belkada, Lewis Tunstall, Edward Beeching, Tristan Thrush, Nathan Lambert, Shengyi Huang, Kashif Rasul, and Quentin Gallouédec.
\newblock Trl: Transformer reinforcement learning.
\newblock \url{https://github.com/huggingface/trl}, 2020.

\bibitem[Bai et~al.(2023)Bai, Bai, Yang, Wang, Tan, Wang, Lin, Zhou, and Zhou]{bai2023qwen}
Jinze Bai, Shuai Bai, Shusheng Yang, Shijie Wang, Sinan Tan, Peng Wang, Junyang Lin, Chang Zhou, and Jingren Zhou.
\newblock Qwen-vl: A versatile vision-language model for understanding, localization, text reading, and beyond.
\newblock \emph{arXiv preprint arXiv:2308.12966}, 1\penalty0 (2):\penalty0 3, 2023.

\bibitem[Face(2025)]{openr1}
Hugging Face.
\newblock Open r1: A fully open reproduction of deepseek-r1, January 2025.
\newblock URL \url{https://github.com/huggingface/open-r1}.

\bibitem[Diederik(2014)]{diederik2014adam}
P~Kingma Diederik.
\newblock Adam: A method for stochastic optimization.
\newblock \emph{(No Title)}, 2014.

\bibitem[Zhao et~al.(2023)Zhao, Gu, Varma, Luo, Huang, Xu, Wright, Shojanazeri, Ott, Shleifer, Desmaison, Balioglu, Damania, Nguyen, Chauhan, Hao, Mathews, and Li]{DBLP:journals/pvldb/ZhaoGVLHXWSOSDB23}
Yanli Zhao, Andrew Gu, Rohan Varma, Liang Luo, Chien{-}Chin Huang, Min Xu, Less Wright, Hamid Shojanazeri, Myle Ott, Sam Shleifer, Alban Desmaison, Can Balioglu, Pritam Damania, Bernard Nguyen, Geeta Chauhan, Yuchen Hao, Ajit Mathews, and Shen Li.
\newblock Pytorch {FSDP:} experiences on scaling fully sharded data parallel.
\newblock \emph{Proc. {VLDB} Endow.}, 16\penalty0 (12):\penalty0 3848--3860, 2023.
\newblock \doi{10.14778/3611540.3611569}.
\newblock URL \url{https://www.vldb.org/pvldb/vol16/p3848-huang.pdf}.

\bibitem[Kwon et~al.(2023)Kwon, Li, Zhuang, Sheng, Zheng, Yu, Gonzalez, Zhang, and Stoica]{kwon2023efficient}
Woosuk Kwon, Zhuohan Li, Siyuan Zhuang, Ying Sheng, Lianmin Zheng, Cody~Hao Yu, Joseph Gonzalez, Hao Zhang, and Ion Stoica.
\newblock Efficient memory management for large language model serving with pagedattention.
\newblock In \emph{Proceedings of the 29th Symposium on Operating Systems Principles}, pages 611--626, 2023.

\bibitem[Hu et~al.(2025)Hu, Zhang, Han, Jiang, Zhang, and Shum]{hu2025reasoner}
Jingcheng Hu, Yinmin Zhang, Qi~Han, Daxin Jiang, Xiangyu Zhang, and Heung{-}Yeung Shum.
\newblock Open-reasoner-zero: An open source approach to scaling up reinforcement learning on the base model.
\newblock \emph{CoRR}, abs/2503.24290, 2025.
\newblock \doi{10.48550/ARXIV.2503.24290}.
\newblock URL \url{https://doi.org/10.48550/arXiv.2503.24290}.

\bibitem[Wang et~al.(2024{\natexlab{d}})Wang, Pan, Shi, Lu, Zhan, and Li]{wang2024measuring}
Ke~Wang, Junting Pan, Weikang Shi, Zimu Lu, Mingjie Zhan, and Hongsheng Li.
\newblock Measuring multimodal mathematical reasoning with math-vision dataset.
\newblock \emph{arXiv preprint arXiv:2402.14804}, 2024{\natexlab{d}}.

\end{thebibliography}
}



\appendix
\newpage
\setcounter{tocdepth}{0}
\addtocontents{toc}{\protect\setcounter{tocdepth}{3}}

\renewcommand{\contentsname}{Appendices Content}  
\hypersetup{linkcolor=teal}
\renewcommand{\cftsecleader}{\cftdotfill{1}} 
\renewcommand{\cftsecfont}{\normalfont} 
\renewcommand{\cftsecpagefont}{\normalfont} 
\renewcommand{\cftsecdotsep}{4.5} 
\renewcommand{\cftsubsecdotsep}{4.5} 
\renewcommand{\cftsubsecnumwidth}{2em} 
\renewcommand{\cftsubsecindent}{1.5em} 
\tableofcontents
\hypersetup{linkcolor=blue}
\thispagestyle{empty}
\newpage
\section{Related Work}
\paragraph{Multimodal Math Reasoning}
The mathematical reasoning capabilities of MLLMs have recently attracted significant attention~\cite{zhuang2025math,gao2023g,li2024llava,dong2024insight,hu2024visual,yang2024mathglm,han24infimm,guo2024mammoth}. Unlike traditional mathematical reasoning tasks in Language Models (LLMs)~\cite{luo2023wizardmath,yu2023metamath}, multimodal mathematical reasoning requires MLLMs to interpret visual information and perform cross-modal reasoning between images and text. Tasks such as solving geometric problems and analyzing graphs are particularly challenging~\cite{chen2021geoqa}. Recent advances have focused on improving visual mathematical input specialized encoders in specific scenarios~\cite{zhang2024mavis,liu2024sphinx,chen2024far}. A significant emphasis has also been placed on synthesizing diverse and complex training data. For instance, Math-LLaVA~\cite{shi2024math} introduces the MathV360K dataset, which categorizes images by complexity and enhances associated questions. Multimath~\cite{peng2024multimath} curates high-quality reasoning data from K-12 textbooks and employs GPT-4 for CoT data generation and validation. R-CoT~\cite{deng2024r} further diversifies problems through a two-stage reverse question-answer generation process. These data synthesis methods are widely adopted in academia and industry due to their demonstrated efficiency~\cite{sprague2024cot,lu2023machine,huang2024key,fu2024vita}.
\paragraph{Process Reward Model} Recent studies have explored test-time scaling laws in LLMs, aiming to identify optimal reasoning ecectories from diverse thinking trajectories~\cite{zhang2024generative,gou2023critic,gao2024llm,lin-etal-2024-criticbench,zhang2024critic,kumar2024training,snell2024scaling}. Initial efforts, such as self-consistency~\cite{wang2022self}, have laid the groundwork for test-time scaling. OpenAI has introduced verifiers to supervise and select reasoning paths during inference~\cite{lightman2023let}. Math-Shepherd~\cite{wang2024math} evaluates intermediate reasoning steps based on their likelihood of leading to correct answers, while OmegaPRM~\cite{luo2024improve} constructs PRM training data and employs MCTS for training. Despite these advancements, the lack of models with robust CoT reasoning capabilities and limited exploration into diverse reward model training data remain significant bottlenecks in multimodal mathematical reasoning. Some concurrent work also begins to pay attention to PRM-assisted visual reasoning, such as construction and benchmarking~\citep{tu2025vilbench,wang2025visualprm,sun2025mm}.
\section{Preliminary}
\subsection{Group Relative Policy Optimization}
Vanilla GRPO eliminates value function in PPO and estimates the advantages within online rollout group. Given a question with image $q$ and ground-truth $y$, policy model $\pi_{\theta_{old}}$ samples a group of $G$ responses $\{o^i\}_{i=1}^{G}$. GRPO compute the $i$-th response's advantage through normalizing in-group rewards $\{r^j\}_{j=1}^{G}$, and employs PPO's clipped objective and KL penalty term:
\begin{equation}\label{equ:adv}
    \begin{aligned}
        A^i=\frac{r^i - \text{mean}(\{r^j\}_{j=1}^{G})}{ \text{std}(\{r^j\}_{j=1}^{G})} 
    \end{aligned}
\end{equation}
\begin{equation}\label{eq:grpo_objective_superscript} 
    \begin{aligned}
        \mathcal{J}_{GRPO}(\theta)&=\mathbb{E}_{(q,y)\sim \mathcal{D}, \{o^i\}_{i=1}^G\sim\pi_{\theta_{old}}(\cdot |q)}\\ 
        &[\frac{1}{G}\sum\limits_{i=1}^G \frac{1}{|o^i|}\sum\limits_{t=1}^{|o^i|}(\min (r_{t}^i(\theta) A^{i}, \text{clip}(r_{t}^i(\theta),1-\epsilon,1+\epsilon) A^i) 
        -\beta D_{KL}^{i,t}(\pi_{\theta}||\pi_{ref}))]
    \end{aligned}
\end{equation}
We introduce the two variants of PRM-integrated GRPO discussed in Section~\ref{sec:stage_3}. 
Given PRM $\mathcal{M}_p$ and process reward sequences $r_s=\mathcal{M}_p(\{s_1,s_2,\cdots,s_{N}\})$ (\romannumeral1) \textit{Variant 1}: Given verifiable outcome reward $r_o^i$, 
we set a single rollout's reward as $r^i=r_o^i+\bar{r_s^i}$. (\romannumeral2) \textit{Variant 2}: 
We utilize a step-level reward and a multiple relative advantage calculated by the mean value of process rewards from each rollout:
\begin{equation}\label{equ:adv_var2}
    \begin{aligned}
        A_{t}^i = \underbrace{ r_{s,t}^i \frac{\bar{r_{s}^i}- \text{mean}({\bar{\{r_{s}^j\}}_{j=1}^{G}})}{ \text{std}({\bar{\{r_{s}^j\}}_{j=1}^{G}}) } }_{\text{GRPO with process rewards}}
        +
        \underbrace{ \frac{r_{o}^i - \text{mean}(\{r_{o}^j\}_{j=1}^{G})}{ \text{std}(\{r_{o}^j\}_{j=1}^{G})} }_{\text{GRPO with outcome rewards}} 
    \end{aligned}
\end{equation}
in which $\bar{r_s^i}=\text{mean}(\mathcal{M}_p(\{s_1^i,s_2^i,\cdots,s_{T_i}^i\}))$.
\subsection{Test-Time Scaling by Best-of-N evaluation}
Following previous works~\cite{lightman2023let,wang2024math}, we adopt BoN evaluation for TTS. Given $N$ response samplings for a question $q$. The PRM is used to give process reward for each sampling. We use mean value of process rewards to select the best single sampling:
\begin{equation}\label{equ:bon}
    \begin{aligned}
        a_{\text{prm}} = \arg\max\limits_{s_i} \text{mean}\{\mathcal{M}_{p}(q, s_i)\}
    \end{aligned}
\end{equation}
Some other works~\citep{gao2024critic} merge self-consistency and PRM to employ a voting-based score cumulation. But we don't select this method for a simpler evaluation manner.
\section{Supplementary Results}
\subsection{Fine-grained Comparison on Used benchmarks}

In this section, we provide some fine-grained results for a clearer comparison. As demonstrated in Table~\ref{tab:mathvision},  our proposed methods demonstrate significant advantages.
Compared to closed-source models like GPT-4o and GPT-4V, our URSA-8B and URSA-8B-PS-GRPO show strong competitiveness. 
Among open-source models, the performance improvements are even more evident. Our URSA-8B model outperforms other open-source models such as InternLM-XComposer2-VL and Ovis1.6-Gemma2-9B in most subtasks. When combined with PS-GRPO, the URSA-8B-PS-GRPO model achieves even better results, showing significant improvements in subtasks like Alg, AnaG, CombG, and others.
Our methods particularly excel in complex mathematical reasoning tasks, demonstrating their powerful mathematical reasoning capabilities. These results highlight the effectiveness of our proposed MMathCoT-1M and PS-GRPO methods in enhancing the mathematical reasoning abilities of models, especially in visual mathematical problems.

In Dynamath~(Table~\ref{tab:dynamath_merged_results}), compared to open-source MLLMs, the URSA series has obvious advantages in plane geometry and algebra. Surprisingly, from the knowledge level classification, the URSA series model performs excellently at the undergraduate level, which is partly attributable to its math-intensive alignment and large-scale instruction fine-tuning.

In MathVerse~(Table~\ref{tab:results_mathverse_filled}), we can see that URSA series model marginally surpass GPT-4o on average. Besides, compared with other open-source MLLMs, URSA-8B-PS-GRPO outperforms leading AtomThink-EMOVA-8B and InternVL2.5-8B with \textbf{8.4} and \textbf{11.4} points.
\begin{table*}[htbp]
\centering
\caption{Performance comparison of different MLLMs on MathVision.}
\label{tab:mathvision}
\resizebox{1.0\linewidth}{!}{
\begin{tabular}{l|c|c|c|c|c|c|c|c|c|c|c|c|c|c|c|c|c|c}
\toprule
Model & Size & ALL & Alg & AnaG & Ari & CombG & Comb & Cnt & DescG & GrphT & Log & Angle & Area & Len & SoIG & Stat & Topo & TransG\\
\midrule
\multicolumn{19}{c}{\textit{Baselines}}\\ \midrule
Human & -&68.8 & 55.1 &78.6 & 99.6 & 98.4 & 43.5 & 98.5 & 91.3 & 62.2 & 61.3 & 33.5 & 47.2 & 73.5 & 87.3& 93.1 & 99.8 & 69.0  \\
\midrule
\multicolumn{19}{c}{\textit{Closed-source MLLMs}}\\ \midrule
GPT-4o & - & \highg{30.4} & \highg{42.0} & \highg{39.3} & \highg{49.3} & \highg{28.9} & \highg{25.6} & \highg{22.4} & 24.0 & 23.3 & \highg{29.4} & 17.3 & \highg{29.8} & \highg{30.1} & \highg{29.1} & \highg{44.8} & \highg{34.8} & 17.9 \\
GPT-4V&-& 22.8 & 27.3 & 32.1 & 35.7 & 21.1 & 16.7 & 13.4 & 22.1 & 14.4 & 16.8 & \highg{22.0} & 22.2 & 20.9 & 23.8 & 24.1 & 21.7 & \highg{25.6} \\
CoT GPT-4V & -&24.0 & 26.7 & 26.2 & 38.6 & 22.1 & 24.4 & 19.4 & \highg{27.9} & 23.3 & 25.2 & 17.3 & 21.4 & 23.4 & 23.8 & 25.9 & 4.4 & \highg{25.6}\\
Gemini-1.5-Pro &-& 19.2 & 20.3 & 35.7 & 34.3 & 19.8 & 15.5 & 20.9 & 26.0 & \highg{26.7} & 22.7 & 14.5 & 14.4 & 16.5 & 18.9 & 10.3 & 26.1 & 17.3 \\
\midrule
\multicolumn{19}{c}{\textit{Open-source MLLMs}}\\ \midrule
LLaVA-1.5 & 7B & 8.5 & 7.0 & 7.1 & 10.7 & 7.1 & 4.8 & 10.5 & 7.7 & 10.0 & 9.2 & 15.6 & 10.2 & 9.8 & 5.3 & 8.6 & 4.4 & 4.8 \\
LLaVA-1.5 & 13B & 11.1 & 7.0 & 14.3 & 14.3 & 9.1 & 6.6 & 6.0 & 13.5 & 5.6 & 13.5 & 10.4 & 12.6 & 14.7 & 11.5 & 13.8 & 13.0 & 10.7 \\
InternLM-XComposer2-VL & 7B &14.5 & 9.3 & 15.5 & 12.1 & 15.3 & 11.3 & 10.5 & 14.4 & \high{22.2} & 19.3 & 19.7 & 15.6 & 15.0 & 11.9 & 15.5 & \high{26.1} & 15.5 \\
Ovis1.6-Gemma2-9B & 9B & 18.8 & 13.3 & 15.5 & 22.1& 17.9 & 11.3 & \highr{22.4} & \high{23.1} & 20.0 & 20.2 & 20.8 & 18.0 & \high{24.7} & 15.6 & 20.7 & 17.4 & 20.8 \\
MiniCPM-v2.6 & 8B & 18.4 & 9.9 & 19.0 & 18.6 & 21.8 & 13.1 & 13.4 & 17.3 & 20.0 & 16.0 & 25.4 & 19.4 & 20.7 & 15.2 & \high{27.6} & \highr{30.4} & 22.0 \\
LLaVA-OneVision& 8B & 18.3 & 11.6 & 16.7 & 20.7 & 18.5 & 11.9 & 14.9 & 19.2 & 13.3 & 20.2 & 17.9 & 21.6 & 23.4 & 12.3 & 22.4 & 13.0 & \high{24.4}  \\
Qwen2-VL & 8B & 19.2 & 15.4 & 20.2 & 19.3 & 16.9 & \high{16.7} & 17.9 & 22.1 & \high{22.2} & 16.0 & 19.1 & 22.4 & 22.5 & 14.8 & 19.0 & 4.3 & 23.8\\
InternVL2-8B &8B & 18.4& 18.6 & 22.6 & 28.6 & \high{22.1} & 13.7 & 10.4 & 11.5 & 13.3 & 21.0 & 20.8 & 22.4 & 20.5 & 16.8 & 17.2 & \high{26.1} & 24.2 \\
InternVL2.5-8B & 8B & 19.7 & 15.1 & 23.8 & 29.3 & 16.2 & 8.9 & 11.9 & 10.6 & 8.9 & 18.5 & 22.0 & 19.4 & 15.4 & 13.9 & 22.4 & 21.7 & 19.6 \\
\midrule
\multicolumn{19}{c}{\textit{Open-source Math MLLMs}}\\ \midrule
Math-LLaVA & 13B & 15.7 & 9.0 & 20.2 & 15.7 & 18.2 & 10.1 & 10.5 & 16.4 & 14.4 & 16.0 & 20.2 & 18.4 & 17.6 & 9.4 & 24.1 & 21.7 & 17.9 \\
Multimath & 7B & 16.3 & 11.3 & 21.1 & 15.5 & 15.9 & 11.3 & 12.1 & 15.5 & 15.9 & 18.5 & 20.1 & 16.4 & 21.3 & 13.3 & 14.6 & 13.3 & 20.8\\
Math-PUMA-Qwen2-7B & 8B &14.0 & 5.0 & 21.1& 21.1 & 21.1 & 11.0 & 5.6 & 15.7 & 10.5 & 13.8 & 11.7 & 15.8 & 12.2 & 17.8 & 19.2 & 15.8 & 12.2\\
MAVIS & 7B & 18.5 & 17.5 & 19.5 & 21.5 & 19.0 & 12.0 & 14.0 & 18.0 & 16.0 & 19.0 & 21.0 & 18.5 & 19.5 & 15.0 & 19.0 & 20.0 & 20.0 \\ 
AtomThink-EMOVA & 8B & 24.9 & 23.5 & 25.5 & \high{32.0} & 21.0 & 15.8 & 19.5 & 21.5 & \highr{22.5} & 21.5 & 26.5 & 25.5 & \high{26.5} & 27.5 & \highr{28.0} & 23.0 & 22.5 \\
\midrule
URSA-8B & 8B & \high{28.7} & \high{28.1} & \high{26.2} & \highr{35.0} & \high{22.1} & 15.5 & 19.4 & 18.3 & \high{22.2} & \high{21.8} & \highr{37.0} & \high{27.0} & \high{26.5} & \high{31.1} & \high{27.6} & 17.4 & 23.8 \\
URSA-8B-PS-GRPO & 8B & \highr{31.5} & \highr{30.1} & \highr{28.6} & 29.3 & \highr{31.5} & \highr{20.8} & \high{20.9} & \highr{26.9} & 17.8 & \highr{24.4} & \high{35.8} & \highr{33.6} & \highr{37.2} & \highr{37.7} & 25.9 & \high{26.1} & \highr{35.1} \\
\bottomrule
\end{tabular}
}
\end{table*}
In WE-MATH~\ref{tab:results_on_WE-MATH_acc}, URSA-series outperforms leading general-purpose and math reasoning MLLMs in three-stage accuracy. Also, the URSA series has remarkable strengths in solid figures, transformations, positions, and directions. This is mainly due to large-scale alignment and instruction tuning, which builds its foundation in understanding mathematical elements.
\begin{table*}[!t] 
\centering
\caption{Detailed performance comparison of MLLMs on \textsc{\textbf{DYNAMATH}} \textit{testmini} dataset, broken down by subject area and knowledge level.}
\label{tab:dynamath_merged_results} 
\resizebox{0.85\linewidth}{!}{
\begin{tabular}{l|c|c|ccccccc|ccc} 
\toprule
Model & Size & ALL & PG & SG & AG & AL & PT & GT & AR & Elem. & High & Undergrad.\\
\midrule 
\multicolumn{13}{c}{\textit{Closed-source MLLMs}}\\ \midrule
GPT-4o & - & \textbf{64.9} & \highg{56.8} & \highg{52.0} & 61.0 & 76.9 & \highg{51.8} & 58.1 & \highg{61.5} & \highg{68.6} & 61.8 & 36.8 \\
Claude-3.5-Sonnet & - &64.8 & 49.9 & 49.3 & 55.3 & \highg{81.0} & 44.1 & \highg{69.4} & 61.2 & 66.7 & \highg{62.6} & 33.3 \\
Gemini-1.5-Pro & - & 60.5 & 52.7 & 42.7 & \highg{61.6}& 70.8 & 20.6 & 65.2 & 54.2 & 62.9 & 59.2 & \highg{37.1} \\ \midrule \midrule
\multicolumn{13}{c}{\textit{Open-source MLLMs}}\\ \midrule 
Llava-v1.5-7B & 7B & 16.6 & 10.5 & 7.3 & 19.5 & 6.5 & 8.2 & 32.3 & 10.8 & 18.9 & 13.3 & 11.7 \\
Llava-v1.6-34B & 34B & 27.1 & 21.4 & 25.3 & 27.6 & 14.9 & 7.6 & 32.7 & 23.1 & 35.9 & 23.8 & 16.6 \\
Deepseek-VL-7B-Chat & 7B & 21.5 & 16.0 & 13.3 & 26.5 & 12.9 & 4.7 & 32.3 & 12.7 & 28.3 & 19.0 & 16.0 \\
InternVL2-8B & 8B & 39.7 & 33.9 & 37.3 & 32.5 & 46.9 & 15.9 & 42.1 & 37.3 & 51.1 & 37.4 & 19.6 \\
Qwen2-VL & 8B & 42.1 & 40.3 & \high{38.7} & \highr{39.9} & 37.1 & 8.2 & \high{44.8} & \high{39.2} & 47.6 & 42.2 & 24.4 \\
AtomThink-EMOVA & 8B & 40.9 & 42.0 & 37.9 & 33.6 & 58.0 & \high{23.0} & 44.0 & 38.4 & \high{52.5} & 43.5 & 32.0 \\ 
\midrule
URSA-8B & 8B & \high{44.7} & \high{48.1} & 38.0 & 33.7 & \highr{66.9} & \highr{24.7} & 39.2 & 38.5 & \highr{53.5} & \high{44.3} & \high{41.8} \\
URSA-8B-PS-GRPO & 8B & \highr{47.4} & \highr{49.7} & \highr{40.1} & \high{35.2} & \high{65.7} & \highr{24.7} & \highr{45.2} & \highr{41.1} & \highr{53.5} & \highr{46.7} & \highr{43.2} \\
\bottomrule
\end{tabular}
}
\end{table*}

\begin{table*}[htbp]
\centering

\caption{Comparison with closed-source MLLMs and open-source MLLMs on \textsc{\textbf{MATHVERSE}} \textit{testmini}. The best results of Closed-source MLLMs are highlighted. The best and second-best results of Open-source MLLMs are highlighted.} 
\label{tab:results_mathverse_filled} 
\resizebox{0.75\linewidth}{!}{ 
\begin{tabular}{l|c|c|c|c|c|c|c|c} 
\toprule
Model & \#Params & ALL & TD & TL & TO & VI & VD & VO \\ 
\midrule
\multicolumn{9}{c}{\textit{Baselines}}\\ \midrule
Random & - & 12.4 & 12.4 & 12.4 & 12.4 & 12.4 & 12.4 & 12.4 \\
Human & - & 64.9 & 71.2 & 70.9 & 41.7 & 61.4 & 68.3 & 66.7 \\
\midrule
 \multicolumn{9}{c}{ \textit{Closed-Source MLLMs} } \\
 \midrule
 GPT-4o & - & \highg{50.8} & \highg{59.8} & 50.3 & \highg{52.4} & \highg{48.0} & \highg{46.5} & \highg{47.6} \\
GPT-4V & - & 39.4 & 54.7 & 41.4 & 48.7 & 34.9 & 34.4 & 31.6 \\
Gemini-1.5-Flash-002 &- & 49.4 & 57.2 & \highg{50.5} & 50.3 & 47.6 & 45.1 & 45.4\\
Gemini-1.5-Pro &- & 35.3 & 39.8 & 34.7  & 44.5 & 32.0 & 36.8 & 33.3 \\
Claude-3.5-Sonnet &- & - & - & - & - & - & - & - \\
Qwen-VL-Plus &- &  21.3 & 26.0 & 21.2 & 25.2 & 18.5 & 19.1 & 21.8  \\
\midrule
\multicolumn{9}{c}{ \textit{Open-Source General MLLMs} } \\ \midrule
mPLUG-Owl2-7B &7B& 10.3 & 11.6 & 11.4 & 13.8 & 11.1 & 9.4 & 8.0 \\
MiniGPT4-7B & 7B & 12.2 & 12.3 & 12.9 & 13.4 & 12.5 & 14.8 & 8.7 \\
LLaVA-1.5-13B & 13B & 12.7 & 17.1 & 12.0 & 22.6 & 12.6 & 12.7 & 9.0  \\
SPHINX-V2-13B & 13B & 16.1 & 20.8 & 14.1 & 14.0 & 16.4 & 15.6 & 16.2 \\
LLaVA-NeXT-34B & 34B & 34.6 & 49.0 & 37.6 & 30.1 & 35.2 & 28.9 & 22.4 \\
InternLM-XComposer2-VL & 7B & 25.9 & 36.9 & 28.3 & 42.5 & 20.1 & 24.4 & 19.8 \\
Deepseek-VL & 8B & 19.3 & 23.0 & 23.2 & 23.1 & 20.2 & 18.4 & 11.8 \\
LLaVA-OneVision (SI) & 8B & 28.9 & 29.0 & 31.5 & 34.5 & 30.1 & 29.5 & 26.9 \\
Qwen2-VL & 8B & 33.6 & 37.4 & 33.5 & 35.0 & 31.3 & 30.3 & 28.1 \\
InternVL2-8B & 8B & 35.9 & 39.0 & 33.8 & 36.0 & 32.2 & 30.9 & 27.7 \\
InternVL2.5-8B & 8B & 39.5 & 43.0 & 43.0 & 43.0 & 43.0 & 42.2 & 22.8\\ 
\midrule
\multicolumn{9}{c}{ \textit{Open-Source Math MLLMs} } \\ \midrule
G-LLaVA-7B & 7B & 16.6 & 20.9 & 20.7 & 21.1 & 17.2 & 14.6 & 9.4 \\
Math-LLaVA-13B &13B & 22.9 & 27.3 & 24.9 & 27.0 & 24.5 & 21.7 & 16.1 \\
Math-PUMA-Qwen2-7B & 8B & 33.6 & 42.1 & 35.0 & 39.8 &  33.4 & 31.6 & 26.0 \\
Math-PUMA-DeepSeek-Math & 7B & 31.8 & 43.4 & 35.4 & 47.5 & 33.6 & 31.6 & 14.7 \\
MAVIS-7B & 7B & 35.2 & 43.2 & 37.2 & 35.2 &  34.1 & 29.7 & \highr{31.8} \\ 
InfiMM-Math & 7B & 40.5 & 46.7 & 39.4 & 41.6 & 38.1 & 40.4 & 27.8 \\ 
Multimath-7B & 7B & 27.7 & 34.8 & 30.8 & 35.3 & 28.1 & 25.9 & 15.0 \\
AtomThink-EMOVA & 8B & 42.5 & 48.1 & 47.7 & 45.7 & 44.0 & \high{44.2} & 26.8 \\  
\midrule 
URSA-8B &8B& \high{45.7} & \high{55.3} & \high{48.3} & \highr{51.8} & \high{46.4} & 43.9 & 28.6\\
URSA-8B-PS-GRPO & 8B & \highr{50.9} & \highr{57.3} & \highr{52.2} & \high{50.2} & \highr{48.7} & \highr{47.6} & \high{31.5} \\
\bottomrule
\end{tabular}
}
\end{table*}

\begin{table}[!t]

\caption{Accuracy comparison with closed-source MLLMs and open-source MLLMs on \textsc{\textbf{WE-MATH}} \textit{testmini} subset. First 3 columns show the overall performance on one-step, two-step and three-step problems. The other columns are used to demonstrate the performance in different problem strategies. \highr{Red} indicates the best performance and \high{Blue} indicates the second best performance among open-source models.}

\label{tab:results_on_WE-MATH_acc}

\centering

\renewcommand{\arraystretch}{1} 

\resizebox{\textwidth}{!}{%

\begin{tabular}{l|c|c|c|c|c|c|c|c|c|c|c|c|c|c|c|c}

\toprule

\multirow{2}{*}{\textbf{Model}} &\multirow{2}{*}{\textbf{\#Params}}&\multirow{2}{*}{\textbf{S1}} & \multirow{2}{*}{\textbf{S2}} &\multirow{2}{*}{\textbf{S3}} & \multicolumn{2}{|c|}{\textbf{Mem}} & \multicolumn{2}{|c|}{\textbf{PF}} & \multicolumn{2}{|c|}{\textbf{SF}} & \multicolumn{2}{|c|}{\centering{\textbf{TMF}}} & \multicolumn{4}{|c}{\textbf{PD}} \\

\cmidrule{6-17}

& & & & & UCU & AL & CPF & UPF & CSF & USF & BTF & CCF & Dir & Pos & RoM & CCP \\

\midrule
\multicolumn{17}{c}{\textit{Closed-source MLLMs}}\\ \midrule
GPT-4o & - &\highg{72.8} & \highg{58.1} & \highg{43.6} & \highg{86.6} & \highg{39.1} & \highg{77.4} & \highg{71.6} & \highg{84.5} & \highg{62.3} & \highg{58.7} &\highg{69.4} & \highg{93.1} & \highg{72.7} & 47.5 & \highg{73.3} \\
GPT-4V & -&65.5 & 49.2 & 38.2 & 82.5 & 38.4 & 70.7 & 60.2 & 76.6 & 56.3 & 57.8 & 67.7 & 79.3 & 57.5 & \highg{47.8} & 63.3 \\
Gemini-1.5-Pro &-& 56.1 & 51.4 & 33.9 & 51.0 & 31.2 & 61.8 & 45.0 & 70.0 & 57.5 & 39.2 & 62.7 & 68.8 & 54.1 & 40.7 & 60.0 \\
Qwen-VL-Max &-& 40.8 & 30.3 & 20.6 & 19.4 & 25.3 & 39.8 & 41.4 & 43.6 & 48.0 & 43.8 & 43.4 & 41.4 & 35.1 & 40.7 & 26.7 \\

\midrule
\multicolumn{17}{c}{\textit{Open-source General MLLMs}}\\ \midrule

LLaVA-1.6 &7B& 23.0 & 20.8 & 15.8 & 18.5 & 20.5 & 16.9 & 29.6 & 15.6 & 18.6 & 42.7 & 24.1 & 17.6 & 43.3 & 28.9 & 26.7 \\
LLaVA-1.6 & 13B&29.4 & 25.3 & 32.7 & 21.7 & 23.2 & 23.4 & 34.7 & 25.3 & 26.4 & 37.5 & 41.7 & 26.9 & 28.9 & 37.1 & 30.0 \\
GLM-4V-9B & 9B&47.3 & 37.2 & 38.2 & 53.4 & 37.0 & 51.3 & 46.5 & 50.6 & 38.2 & 44.1 & 45.2 & 41.0 & 49.3 & 36.8 & 53.3 \\
MiniCPM-LLaMA3-V2.5 &8B& 39.8 & 31.1 & 29.7 & 28.6 & 37.0 & 40.8 & 39.8 & 41.0 & 38.6 & 32.0 & 42.7 & 41.0 & 42.7 & 44.0 & 43.3 \\
LongVA &7B& 43.5 & 30.6 & 28.5 & 24.5 & 39.8 & 45.1 & 40.8 & 51.9 & 42.5 & 45.6 & 44.6 & 44.5 & 40.7 & 47.5 & 20.0 \\
InternLM-XComposer2-VL&7B & 47.0 & 33.1 & 33.3 & 31.3 & \high{46.5} & 47.7 & 42.6 & 51.4 & 43.9 & 41.1 & 50.6 & 65.5 & 53.9 & 55.2 & 40.0 \\
Phi3-Vision&4.2B & 42.1 & 34.2 & 27.9 & 28.7 & 16.0 & 47.2 & 38.8 & 50.0 & 44.4 & 28.8 & 31.2 & 48.6 & 49.2 & 26.4 & 50.0 \\
DeepSeek-VL &7B& 32.6 & 26.7 & 25.5 & 16.6 & 35.1 & 27.3 & 38.0 & 24.2 & 38.7 & 50.0 & 23.3 & 24.5 & 41.0 & 51.7 & 23.3  \\
InternVL2-8B & 8B & 59.4 & 43.6 & 35.2 & \high{71.4} & 20.5 & 62.0 & 55.5 & 67.1 & 57.3 & \high{54.0} & \high{60.5} & 58.6 & 63.6 & 44.5 & 50.0 \\
InternVL2.5-8B & 8B & 58.7 & 43.1 & 38.8 & 48.7 & 35.8 & 65.5 & 54.5 & 62.3 & 61.5 & 47.8 &  60.3& \highr{79.0}&  64.0& 51.1 & \highr{63.3}\\ 
Qwen2-VL & 8B & 59.1 & 43.6 & 26.7 & 62.7 & 37.2 & 62.6 & \high{60.8} & 65.7 & 49.2 & 52.5 & 49.2 & 48.1 & \high{68.2} & 55.0 & \high{56.7} \\
Gemma3-12B & 12B & \high{64.3} & 47.2 & \high{42.1} & \highr{83.1} & 33.9 & \high{70.2} & 58.2 & \highr{77.5} & 61.1 & 50.1 & 63.7 & 82.6 & 58.4 & 36.8 & 60.0\\
\midrule
\multicolumn{17}{c}{\textit{Open-source Math MLLMs}}\\ \midrule

G-LLaVA & 7B&32.4 & 30.6 & 32.7 & 33.3 & 29.1 & 32.0 & 37.9 & 19.6 & 33.5 & 37.1 & 32.8 & 31.2 & 33.2 & 25.6 & 40.0 \\
Math-LLaVA&13B& 38.7 & 34.2 & 34.6 & 30.3 & 17.9 & 39.2 & 40.4 & 37.1 & 37.7 & 53.0 & 51.3 & 30.8 & 30.8&40.9 & 46.7 \\
Math-PUMA-Qwen2-7B&8B & 53.3 & 39.4 & 36.4 & 63.5 & 42.5 & 60.2 & 45.9 & 66.2 & 48.6 & 42.3 & 53.5 & 31.2 & 37.7 & 40.4 & 46.7 \\
MAVIS w/o DPO & 7B & 56.9 & 37.1 & 33.2 & - & - & - & - & - & - & - & - & - & - & - & -\\
MAVIS & 7B & 57.2 & 37.9 & 34.6 & - & - & - & - & - & - & - & - & - & - & - & -\\ \midrule
URSA-8B &8B & 63.1 & \high{56.4} & 41.8 & 59.1 & 32.5 & \highr{72.3} & 60.3 & \high{70.9} & \high{66.0} & 51.4 & 59.8 & 58.3 & 39.5 & \high{58.8} & 53.3 \\
URSA-8B-PS-GRPO & 8B & \highr{68.6} & \highr{64.2} & \highr{52.7} & 52.6 & \highr{63.5} & 68.5 & \highr{64.1} & 68.8 & \highr{73.6} & \highr{69.4} & \highr{75.8} & \high{72.1} & \highr{72.6} & \highr{73.6} & \highr{63.3} \\
\bottomrule
\end{tabular}
}
\centering
\end{table}
\subsection{Scaling Law of MMathCoT-1M}
To better illustrate the effectiveness of MMathCoT-1M, we examine the scaling laws of SFT by training models on randomly selected samples representing various ratios of the full dataset.
\begin{table}[htbp] 
\centering
\caption{Scaling law validation on URSA-8B using different ratios of the MMathCoT-1M.} 
\label{tab:scaling_results} 
\resizebox{0.8\linewidth}{!}{
\begin{tabular}{@{}cccccc@{}} 
\toprule
\textbf{Ratio} & \textbf{MathVerse} & \textbf{MathVision} & \textbf{MathVista-GPS} & \textbf{WEMATH} & \textbf{DYNAMATH} \\
\midrule
1/4 & 34.7 & 20.5 & 68.5 & 43.5 & 36.6 \\
1/2 & 40.5 & 22.8 & 72.3 & 47.7 & 38.8 \\
3/4 & 42.0 & 26.7 & 77.9 & 50.9 & 42.2 \\
1   & \textbf{45.7} & \textbf{28.7} & \textbf{81.7} & \textbf{53.6} & \textbf{44.7} \\
\bottomrule
\end{tabular}}
\end{table}

As shown in Table~\ref{tab:scaling_results}, we can see that MMathCoT-1M clearly shows a training time scaling law, further validates the effectiveness of the synthesized data.

\subsection{Higher Upper Bound Taken from Stage \RN{1}}\label{app:upper_bound}

\begin{figure}[!t]
    \centering
    \includegraphics[width=1.0\linewidth]{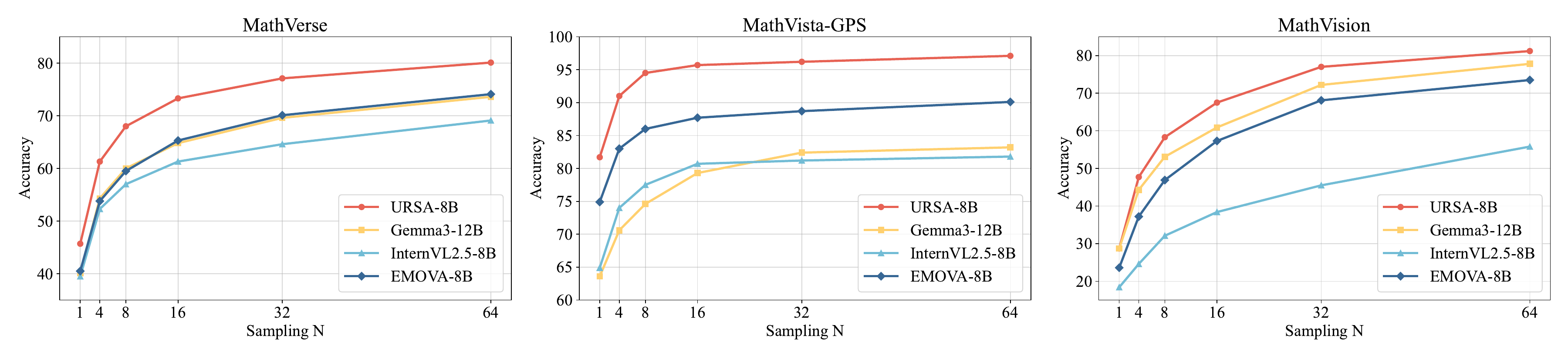}
    \caption{\textbf{Pass@N} evaluation on three benchmarks.}
    \label{fig:passn}
\end{figure}

In Stage \RN{1}, we obtain a more powerful base MLLM with enhanced reasoning capabilities through math-intensive vision-language alignment and instruction fine-tuning. Beyond the results in Table~\ref{tab:performance_opensource_highlight}, we explain why Stage \RN{1} can better serve subsequent experiments, focusing on test-time scaling and PRM applications. We select MathVerse, MathVision, and MathVista-GPS to observe the \textbf{pass@N} metric. As demonstrated in Figure~\ref{fig:passn}, we find that URSA-8B consistently outperforms current leading general MLLMs and math reasoning MLLMs. This indicates that while current trends favor RL-related techniques, the scaling law of supervised fine-tuning can still demonstrate its role in breaking through the base model's limitations. This naturally brings advantages in areas such as BoN evaluation and the proportion of valuable rollouts in online RL. First, URSA-8B's higher upper bound leads to richer and more reliable process label generation in Stage \RN{2}. Furthermore, since recent works claims that RL can only approach the optimal solution within its own exploration path~\citep{yue2025limit-of-rlvr,shao2024deepseekmath,zeng2024b}, Stage \RN{1} naturally expands the potential upper limit of the RL stage. This provides the most fundamental advantage to the performance of URSA-PS-GRPO-8B.

\subsection{Generalization Validation}\label{app:generalization}
To further vallidate the effectiveness of proposed MMathCoT-1M and PRM aided PS-GRPO. We select InternVL2.5-8B from the general-purpose MLLMs and Multimath from the math reasoning MLLMs for a generalization validation experiment. 
We do not conduct additional hyperparameter tuning but almost directly adopt the settings from Table~\ref{tab:hyperparameter}. The experiment on InternVL2.5-8B and Multimath are implemented on~\citet{meng2025mm} and TRL~\citep{vonwerra2022trl}. Given that these two models have already undergone sufficient alignment for general domains or specific vertical domains upon their release, we only carry out two stages of training: (\romannumeral1) MMathCoT-1M is used to enhance the base model’s mathematical reasoning capabilities; (\romannumeral2) URSA-8B-RM is involved in the PS-GRPO process. We present the results in Table~\ref{tab:generalization_progress}.
\begin{figure}[!t]
    \centering
    \caption{The progress of MMathCoT-1.1M and URSA-8B-RM aided PS-GRPO on InternVL2.5-8B and MultiMath.} 
    \label{tab:generalization_progress} 
    \vspace{-0.1in}
    \resizebox{0.95\textwidth}{!}{
        \begin{tabular}{@{}llllllll@{}}
            \toprule
            \multirow{2}{*}{\textbf{Model}} & \multirow{2}{*}{\textbf{Avg}} & \multicolumn{1}{l}{\textbf{MathVerse}} & \multicolumn{1}{l}{\textbf{MathVision}} & \multicolumn{1}{l}{\textbf{MathVista}} & \multicolumn{1}{l}{\textbf{WE-MATH}} & \multicolumn{1}{l}{\textbf{DYNAMATH}} & \multicolumn{1}{l}{\textbf{GeoQA}} \\
            & & \multicolumn{1}{l}{\small testmini} & \multicolumn{1}{l}{\small full set} & \multicolumn{1}{l}{\small gps} & \multicolumn{1}{l}{\small testmini} & \multicolumn{1}{l}{\small testmini} & \multicolumn{1}{l}{\small full set} \\
            \midrule
            InernVL2.5-8B & 45.2 & 39.5      & 19.7       & 64.9      & 44.7    & 40.5     & 61.6  \\ \midrule
            \ \ +\ MMathCoT-1.1M & 51.7 \difftext{6.5} & 43.3 \difftext{3.8} & 25.9 \difftext{6.2} & 77.9 \difftext{13.0} & 49.1 \difftext{4.4} & 45.3 \difftext{4.8} & 68.8 \difftext{7.2} \\
            \ \ +\ PS-GRPO      & \textbf{54.7} \difftext{9.5} & \textbf{47.5} \difftext{8.0} & \textbf{28.5} \difftext{8.8} & \textbf{80.1} \difftext{15.2} & \textbf{55.3} \difftext{10.6} & \textbf{45.9} \difftext{5.4} & \textbf{71.1} \difftext{9.5} \\
            \midrule 
            \midrule
            MultiMath & 43.1 & 27.7 & 16.3 & 66.8 & 42.2 & 37.9 & 67.7 \\ \midrule
            \ \ +\ MMathCoT-1.1M & 48.7 \difftext{5.6} & 36.9 \difftext{9.2} & 22.7 \difftext{6.4} & 74.4 \difftext{7.6} & 45.8 \difftext{3.6} & 40.4 \difftext{2.5} & 72.2 \difftext{4.5} \\
            \ \ +\ PS-GRPO      & \textbf{51.2} \difftext{8.1} & \textbf{39.7} \difftext{12.0} & \textbf{24.4} \difftext{8.1} & \textbf{77.7} \difftext{10.9} & \textbf{49.3} \difftext{7.1} & \textbf{42.6} \difftext{4.7} & \textbf{73.5} \difftext{5.8} \\
            \bottomrule
        \end{tabular}
    }
\end{figure}
The proposed MMathCoT-1M and PRM aided PS-GRPO demonstrate remarkable generalization capabilities across different models and benchmarks. When applied to InternVL2.5-8B and MultiMath, both models show significant performance improvements. For InternVL2.5-8B, adding MMathCoT-1M boosts the average score from 45.2 to 51.7, with even more significant gains when combined with PS-GRPO, reaching 54.7. Similarly, for MultiMath, the average score increases from 43.1 to 48.7 with MMathCoT-1M and further to 51.2 with PS-GRPO. These results highlight the effectiveness of our approach in enhancing mathematical reasoning capabilities across diverse models and tasks. The performance improvements are consistent across various benchmarks, including MathVerse, MathVision, MathVista, WE-MATH, DYNAMATH, and GeoQA, indicating that our methods are not only effective but also broadly applicable.
\subsection{Implementary Results on Other Benchmarks}
We provide supplementary results on WE-MATH, DYNAMATH and GeoQA when comparing BoN selection. As shown in table~\ref{tab:prm_comparison_new_datasets}, URSA-8B-RM remains an advantage with Self-consistency and InternVL2.5-8B ORM. When employing URSA-8B as reasoning model,  URSA-8B-RM outperforms Self-consistency with 4.6\%, 4.5\% and 2.7\% relative improvements in Best-of-8 performance.
\begin{table*}[!t]
  \centering
  \caption{Comparison of TTS with different models using BoN performance on WE-MATH, DYNAMATH, and GeoQA.}
  \label{tab:prm_comparison_new_datasets}
  \resizebox{0.95\textwidth}{!}{%
  \begin{tabular}{llcccccccccccc}
    \toprule
    \multirow{2}{*}{\textbf{Model}} & \multirow{2}{*}{\textbf{Method}} & \multicolumn{4}{c}{\textbf{WE-MATH}} & \multicolumn{4}{c}{\textbf{DYNAMATH}} & \multicolumn{4}{c}{\textbf{GeoQA}} \\
    \cmidrule(lr){3-6} \cmidrule(lr){7-10} \cmidrule(lr){11-14}
    & & N=4 & N=8 & N=16 & N=32 & N=4 & N=8 & N=16 & N=32 & N=4 & N=8 & N=16 & N=32 \\
    \midrule
    \multirow{3}{*}{URSA-8B}
    & Self-Consistency & 56.3 & 57.0 & 57.7 & 58.0 & 46.2 & 46.7 & 47.5 & 48.0 & 74.1 & 75.3 & 75.9 & 75.9 \\
    & InternVL2.5-8B ORM & 56.0 & 56.8 & 57.4 & 57.7 & 45.9 & 46.5 & 47.2 & 47.7 & 73.8 & 75.0 & 75.6 & 75.6 \\
    & URSA-8B-RM & \textbf{58.2} & \textbf{59.0} & \textbf{59.3} & \textbf{59.7} & \textbf{47.5} & \textbf{48.4} & \textbf{49.5} & \textbf{50.5} & \textbf{76.1} & \textbf{77.3} & \textbf{78.0} & \textbf{78.1} \\
    \midrule
    \multirow{3}{*}{AtomThink-EMOVA}
    & Self-Consistency & 51.7 & 52.4 & 52.9 & 53.6 & 42.3 & 43.0 & 43.7 & 44.0 & 65.7 & 66.5 & 66.6 & 66.8 \\
    & InternVL2.5-8B ORM & 51.5 & 52.2 & 52.7 & 53.3 & 42.1 & 42.8 & 43.5 & 43.7 & 65.5 & 66.3 & 66.4 & 66.6 \\
    & URSA-8B-RM & \textbf{53.7} & \textbf{54.5} & \textbf{55.0} & \textbf{55.8} & \textbf{44.1} & \textbf{44.9} & \textbf{45.6} & \textbf{46.0} & \textbf{67.9} & \textbf{68.8} & \textbf{69.0} & \textbf{69.3} \\
    \bottomrule
  \end{tabular}%
  }
\end{table*}
\section{Module Selection Criteria}\label{app:selection_criteria}
As for module selection, we primarily considered the choice of the vision encoder and the LLM backbone. 

\paragraph{Vision Encoder} To train a reasoning model with higher process sensibility and facilitate PRM training, we first conduct captioning tests on open-source models like DeepSeek-VL, Qwen2-VL, etc., using a manually selected dataset (approximately 80 examples). These examples primarily include function-related and geometry problems prone to visual confusion. We manually inspect the outputs of these open-source models and find that Qwen2-VL and LLaVA-OneVision performed poorly; even though their performance on standard benchmarks is good, they fail to ensure sufficiently accurate mathematical descriptions. However, DeepSeekVL's native hybrid vision tower design, integrating high- and low-resolution processing, subjectively exhibit better recognition accuracy. We speculate that this is due to QwenViT~\citep{bai2023qwen} being more heavily biased towards general multimodal tasks, resulting in less precise mathematical descriptions compared to simpler vision backbones. Therefore, we choose the SigLiP-L+SAM-B hybrid vision tower design.

\paragraph{LLM Backbone} Considering the open-source influence of the QwenLM-Series, we follow the choice of prior work such as MathPUMA~\citep{zhuang2025math} and Multimath~\citep{peng2024multimath} by using the QwenLM-Series backbone. However, we consider whether we could achieve higher performance by leveraging instruction models that has undergone unimodal math post-training, and thus compare Qwen2.5-7B-Instruct\footnote{\href{https://huggingface.co/Qwen/Qwen2.5-7B-Instruct}{https://huggingface.co/Qwen/Qwen2.5-7B-Instruct}} and Qwen2.5-Math-7B-Instruct\footnote{\href{https://huggingface.co/Qwen/Qwen2.5-Math-7B-Instruct}{https://huggingface.co/Qwen/Qwen2.5-Math-7B-Instruct}}. After completing the VL alignment stage, we conduct a small-scale comparative experiment on MMathCoT-1M, fine-tuning on ~50K examples. Finally, our results show that using Qwen2.5-Math-Instruct as the backbone yields an advantage of approximately 1 percentage point on MathVision and MathVerse. Therefore, we include Qwen2.5-Math-7B-Instruct as the LLM backbone for subsequent experiments.

\section{Ablation Studies}
\subsection{Effectiveness of Different Data Category}\label{app:abl1}
\begin{wrapfigure}{r}{0.5\textwidth}
    \vspace{-2ex} 
    \centering
    \includegraphics[width=1.0\linewidth]{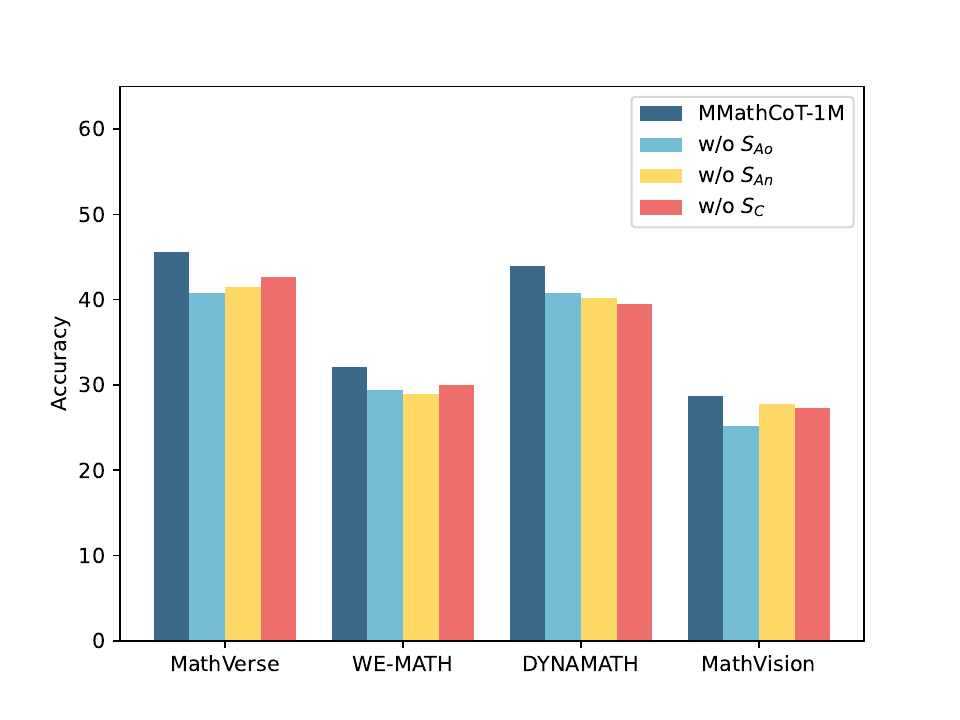}
    \vspace{-0.1in}
    \caption{Each synthesis strategy towards different type of source data works well.}
    \label{fig:data_ablation}
\end{wrapfigure}
In the first stage, we mainly synthesized large-scale multiclass CoT data. 

\begin{itemize}[left=1em] 
    \item \textbf{w/o }$\mathcal{S}_{Ao}$: In this variant, the answer-only data is reverted to its original format. This directly mimics the training mode used by models such as Math-LLaVA and Math-PUMA, which involves hybrid training on both direct answers ('fast thinking') and CoT thinking.
    \item \textbf{w/o }$\mathcal{S}_{An}$: This data will be replaced with its original organizational structure, where the analysis and final answer are provided in a free-form text format.
    \item \textbf{w/o }$\mathcal{S}_C$: This batch of data will be replaced with reasoning expressed in mathematical formal language, better reflecting symbolic and 'plan and reasoning' forms of reasoning.
\end{itemize}
The results are shown in Figure~\ref{fig:data_ablation}. Firstly, it is shown across all datasets that using the complete synthesized data achieves the best results, highlighting the role of MMathCoT-1M data. More specifically, we find: \romannumeral1) $\mathcal{S}_{Ao}$ demonstrates the greatest impact on MathVerse and MathVision, indicating that expanded CoT data is important for problems where absolute solution accuracy is pursued; \romannumeral2) However, on WE-MATH, the replacement of $\mathcal{S}_{An}$ leads to the most significant performance drop, suggesting that content rewriting better aligns with the end-to-end requirements posed by the WE-MATH benchmark, and mixing training with data lacking clear logical sequences may reduce hierarchical accuracy; \romannumeral3) The results on DYNAMATH indicate that rewriting and natural language formulation effectively enhance reasoning robustness from the perspective of textual diversity. This reveals that the thought pattern in textual form tends to maintain the stability of the thought process more effectively under scenarios involving image transformations.

\subsection{Selection of External Closed-source MLLM}  
In this section, we primarily present a comparison of metrics between Gemini-1.5-Flash-002\footnote{\href{https://deepmind.google/technologies/gemini/flash/}{https://deepmind.google/technologies/gemini/flash/}} and other popular MLLMs, as well as a comparison on partial training data. 
\begin{table}[htbp] 
    \centering 
    \caption{SFT Performance with 50K data synthesized by two closed-source MLLMs, respectively.} 
    \label{tab:sft_comparison_inline} 
    \resizebox{0.8\linewidth}{!}{%
    \begin{tabular}{l|ccc} 
    \toprule
    Model & MathVista-GPS & MathVerse & MathVision \\
    \midrule
    URSA-8B w/ GPT-4o   & 54.1  & 33.3  & 18.8   \\
    \midrule
    URSA-8B w/ Gemini-1.5-Flash-002   & 55.1  & 32.5  & 18.3   \\
    \bottomrule
    \end{tabular}}
\end{table}

\begin{itemize}[left=1em] 
    \item \textit{Metrics Comparison}: We compare the performance of Gemini-1.5-Flash-002, GPT-4o, and GPT-4o-mini on some math-related tasks, as shown in Table~\ref{tab:close_mllm}. We observe that Gemini-1.5-Flash-002 is a MLLM that performs well on both unimodal and multimodal math tasks, and GPT-4o does not have a significant advantage over it. This, to some extent, ensures the quality of the data synthesis.

\item \textit{SFT Performance}: To best illustrate the performance variations, we randomly sampled 50K data sources from MMathCoT-1M, applied three corresponding strategies using GPT-4o, and subsequently conducted SFT. The performance results are shown in Table~\ref{tab:sft_comparison_inline}. We observe that using GPT-4o did not provide a clear advantage. However, the construction of MMathCoT-1M and DualMath-1.1M involves approximately 2.7 million API calls. The output token cost of Gemini-1.5-flash-002 is the same as that of GPT-4o-mini and is one-thirty-third of that of GPT-4o\footnote{\href{https://docsbot.ai/models/gemini-1-5-flash-002}{https://docsbot.ai/models/gemini-1-5-flash-002}}. Therefore, Gemini-1.5-Flash-002 becomes a cost-effective choice.
\end{itemize}

However, we must say that if community researchers can afford the cost of accessing more powerful closed-source models, we expect the results to be even better.

\begin{table}[htbp] 
\centering 
\caption{Comparison of Model Performance on Math Benchmarks} 
\label{tab:close_mllm} 
\resizebox{0.7\linewidth}{!}{%
\begin{tabular}{l|ccccc} 
\toprule 
          \textbf{Model} & \textbf{Avg} & \textbf{MATH} & \textbf{MathVista} & \textbf{MathVerse} & \textbf{MathVision} \\ 
\midrule 
GPT-4o     & 55.4         & 76.6          & 63.8               & 50.8               & 30.4                \\
\midrule 
Gemini-1.5-Flash-002 & 53.6 & 79.9          & 58.4               & 49.4               & 26.3                \\
\midrule 
GPT-4o-mini & 48.0        & 70.2          & 56.7               & 42.3               & 22.8                \\
\bottomrule 
\end{tabular}
}
\end{table}

\newpage
\section{Implementation Details}\label{app:implementation_details}
\subsection{RL Data Curation}
\begin{wrapfigure}{r}{0.4\textwidth}
    \vspace{-2ex}
    \centering
    \caption{Statistics of RL data for vanilla GRPO and PS-GRPO.}
    \label{tab:rl_statistics}

    \vspace{-1ex} 

    \resizebox{0.9\linewidth}{!}{%
    \begin{tabular}{lr}
     \toprule
     \textbf{Statistic} & \textbf{Number} \\
     \midrule
      \textit{Total Prompts}&  \\
      ~- Total number &15.3K \\
    \textit{Data Source}& \\
      ~- MathV360K & 2.7K \\
      ~- Multimath-EN & 7.5K \\
      ~- MAVIS-Geo & 2.2K \\
      ~- MAVIS-MetaGen & 0.9K \\
      ~- Geo170K-QA & 2.1K \\
     \midrule
     \textit{Problem Category Statistics} \\ 
  ~- Plane Geometry & 4.1K~(26.8\%)  \\
  ~- Analytic Geometry & 1.9K~(12.4\%) \\
  ~- Solid Geometry & 1.0K~(6.5\%) \\
  ~- Algebra & 1.1K~(7.2\%) \\
  ~- Function & 2.9K~(19.0\%) \\
  ~- Chart & 1.9K~(12.4\%) \\
  ~- Textbook QA & 0.9K~(5.9\%) \\
  ~- Formula & 0.8K~(5.2\%) \\
  ~- Arithmetic & 0.7K~(4.6\%) \\
     \bottomrule
     \vspace{-3ex}
    \end{tabular}
    }
\end{wrapfigure}
After instruction fine-tuning on MMathCoT-1M, the overall accuracy did not exceed 50\%. Therefore, we believe it still has the potential to be directly utilized in the RL phase. We collect 20K data with a types mixture ratio similar to that of instruction fine-tuning and conduct a one-time static filtering before RL. Specifically, we use URSA-8B to perform 8 samplings on this 20K data, filtering out examples where all 8 sampling results are either incorrect or correct. This left approximately 15K+ data for training vanilla GRPO and PS-GRPO. We implement PS-GRPO using TRL~\citep{vonwerra2022trl,openr1}. The statistics of the RL data can be found in the table~\ref{tab:rl_statistics}.

\subsection{Parameters and Time Cost}
\vspace{-3pt}

In this section, we provide the specific parameter settings and time costs for the three stages. Our experiments are based on Python 3.10 and PyTorch 2.4.0+cu124. We use AdamW~\citep{diederik2014adam} as the optimizer. We use Fully Shared Data Parallel (FSDP)~\citep{DBLP:journals/pvldb/ZhaoGVLHXWSOSDB23} as the distributed training framework. Unless otherwise specified, experiments are conducted on 32$\times$ NVIDIA-H100-HBM3 GPUs by default. Additionally, we provide important parameters used in data construction. During the generation of positive and negative example pairs, we set the $temperature$ to 1.0, $n\_return\_sequences$ to 16, and $top\_p$ to 0.95. In the \textit{BinaryErrorLocating} phase, we set the $temperature$ to 0.3, $n\_return\_sequences$ to 16, and $top\_p$ to 0.95. 

We adapt the vLLM~\citep{kwon2023efficient} framework for the URSA-8B's architecture~(hybrid vision tower + MLP + Qwen2.5-math-Instruct is not originally supported by VLLM) and use it as an acceleration tool during the inference phase. During the data pair generation phase, we use 16$\times$ NVIDIA-H100-HBM3 GPUs for inference, which takes approximately 28 hours. In the~\textit{BinaryErrorLocating} phase, we also use 16$\times$ NVIDIA-H100-HBM3 GPUs for inference, taking about 20 hours.

The hyperparameter and time cost used in Stage \RN{1} and Stage \RN{2} are demonstrated in Table~\ref{tab:hyperparameter}. Since the parameters used in Stage \RN{3} are somewhat different, we list them separately in Table~\ref{tab:hyperparameter_rl}.
Recently, much work has provided numerous optimization tricks for GRPO, such as training-time dynamic sampling, clipping higher values, abandoning KL loss, etc~\citep{hu2025reasoner,yu2025dapo}. However, to independently verify the effectiveness of PRM-guided reward modeling, we have not added these tricks in either vanilla GRPO or PS-GRPO to ensure a fair and valid verification process. We only do \textbf{one-time} difficulty-based data selection before applying RL. 

\begin{table*}[htbp]
\centering
\caption{Hyperparameter setting and training time cost in Stage \RN{1} and \RN{2}.}
\label{tab:hyperparameter}
\resizebox{0.9\linewidth}{!}{
\begin{tabular}{l|c|c|c} 
\toprule
Hyperparameters \& Cost & VL-alignment & Instruction Fine-tuning & PRM Training\\
\midrule 
Learning Rate & 1e-4 & 1e-5 & 5e-6 \\
Epoch & 1 & 2 & 2 \\
Warm-up Ratio & 0.02 & 0.02 & 0.02 \\
Weight Decay & 0.02 & 0.01 & 0.02 \\
Batch Size & 64 & 128 & 128 \\
Trainable Parts & Aligner & Vision Encoder, Aligner, Base LLM & Base LLM \\
Data Size & 860K & 1.0M & 1.1M \\
Time Cost & $\sim$3.5h & $\sim$11h & $\sim$12h \\
\bottomrule
\end{tabular}
}
\end{table*}

\subsection{Benchmarks}\label{app:sec:benchmark}

In this section, we introduce the detailed subtasks and metrics of four used benchmarks to more precisely demonstrate the evaluation.

\paragraph{MathVerse} MathVerse~\citep{zhang2025mathverse} is a benchmark for testing the reasoning abilities of MLLMs when the information content in text and image modalities varies. Specifically, the models focus on performance in six scenarios: Text-Dominant~(TD), Text-Lite~(TL), Text-Only~(TO), Vision-Intensive~(VI), Vision-Dominant~(VD) and Vision-Only~(VO).
\begin{wrapfigure}{r}{0.35\linewidth}
\centering
\captionof{table}{Hyperparameter setting and training time cost in Stage \RN{3}.}
\label{tab:hyperparameter_rl}
\resizebox{0.9\linewidth}{!}{
\begin{tabular}{l|c} 
\toprule
Hyperparameters / Cost & Value\\
\midrule 
Epochs & 2 \\
Learning Rate & 2e-6 \\
Temperature & 1.0 \\
Rollout number per prompt & 8 \\
Prompt Max Length & 6048 \\
Output Max Length & 3072 \\
Precision & bf16 \\
Train Batch Size & 512 \\
KL Coefficient & 0.003 \\
Data size & 15K \\
Time Cost & $\sim$18h \\
\bottomrule
\end{tabular}
\vspace{-4pt}
}
\end{wrapfigure}
\paragraph{WE-MATH} WE-MATH~\citep{qiao2024we} is the first benchmark that decompose composite problems into sub-problems according to the required knowledge concepts. In figure~\ref{tab:results_on_WE-MATH_acc}, the actual content corresponding to the abbreviations is as follows. Mem: Measurement, PF: Plane Figures, SF: Solid Figures, TMF: Transformations and Motion of Figures, PD: Position and Direction, AL: Angles and Length, UCU: Understanding and Conversion of Units, CPF: Calculation of Plane Figures, UPF: Understanding of Plane Figures, CSF: Calculation of Solid Figures, USF: Understanding of Solid Figures, BTF: Basic Transformations of Figures, CCF: Cutting and Combining of Figures, Dir: Direction, Pos: Position, RoM: Route Map, CCP: Correspondence of Coordinates and Positions. 
\paragraph{DYNAMATH} DYNAMATH~\citep{zou2024dynamath} is a benchmark designed to evaluate the robustness of MLLMs in mathematical reasoning. 
Specifically, it includes tests across multiple dimensions, including Solid Geometry~(SG), Plane Geometry~(PG), Analytic Geometry~(AG), Algebra~(AL), Puzzle Test~(PT), Graph Theory~(GT), Arithmetic~(AR), Scientific Figure~(SF) and Statistics~(ST). It includes 501 seed questions and 5010 generated questions.
\paragraph{GeoQA} The GeoQA~\citep{chen2021geoqa} dataset is a specialized dataset designed for evaluating and training models in the field of geographic question answering. Its test set includes 734 samples.
\paragraph{MathVista} MathVista~\citep{lu2023mathvista} comprises a total of 5 subtasks: Geometry Problem Solving~(GPS), Math Word Problem~(MWP), Figure Question Answering~(FQA), Textbook Question Answering~(TQA) and Visual Question Answering~(VQA). Like the previous math reasoning works, our model training process does not overly focus on knowledge-intensive tasks~(such as VQA and FQA), hence we choose GPS as the primary task.
\paragraph{MathVision} MathVision~\citep{wang2024measuring} is a large-scale multimodal math reasoning dataset that broadens the disciplinary scope of the multimodal mathematics field. The test set contains 3,040 examples, covering 16 key competencies, and provides reliable testing performance. Specifically, The specific meanings of the various disciplinary indicators in Table~\ref{tab:mathvision} are listed as following. Alg: algebra, AnaG: analytic geometry, Ari: arithmetic, CombG: combinatorial geometry, Comb: combinatorics, Cnt: counting, DescG: descriptive geometry, GrphT: graph theory, Log: logic, Angle: metric geometry - angle, Area: metric geometry - area, Len: metric geometry-length, SolG: solid geometry, Stat: statistics, Topo: topology, TransG: transformation geometry.

\paragraph{Evaluation Criteria} Our comparison is based on the following criteria: First, we select the results from the official leaderboards of each benchmark. Second, we choose the results from the original papers or technical reports of each model. Finally, we conduct our own inference and evaluation using vLLM~\citep{kwon2023efficient}. Our evaluation adheres to the rules of the benchmarks themselves, which are as follows:

\begin{itemize}[left=1em] 
    \item \textbf{Rule-based Matching}: WEMATH, GeoQA.
    \item \textbf{LLM-as-a-Judge}: MathVista, MathVision, MathVerse, Dynamath.
\end{itemize}

\begin{figure}[!t]
    \centering
    \includegraphics[width=1.0\linewidth]{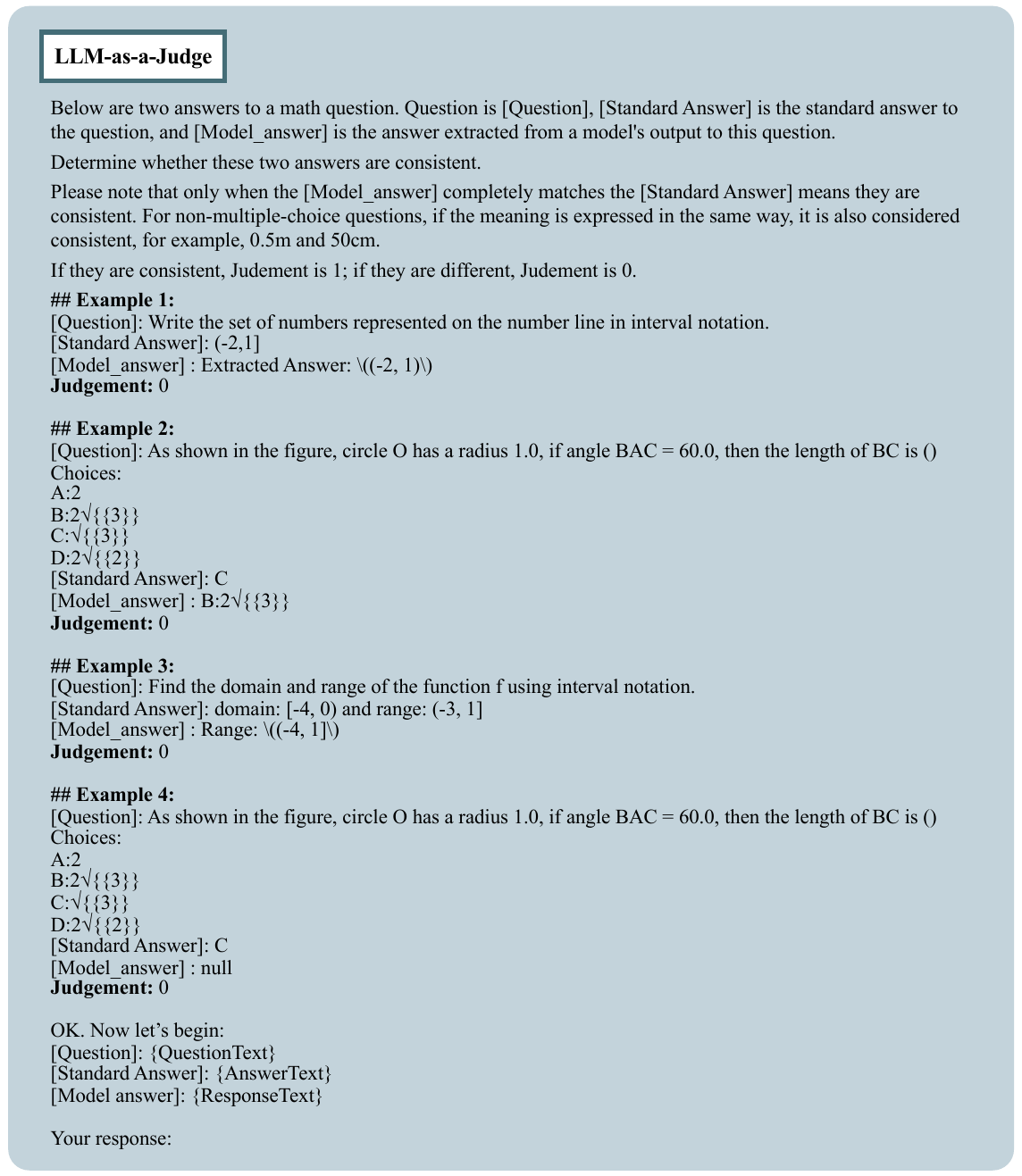}
    \vskip -0.1in
    \caption{LLM-as-a-Judge prompt used for answer matching.}
    \label{fig:llm_as_a_judge}
\end{figure}

The prompt for LLM-as-a-Judge is shown in Figure~\ref{fig:llm_as_a_judge}.

\subsection{Algorithm}
In this section, we place the specific process of BIE from Section~\ref{sec:misinterpretation} into Algorithm~\ref{alg:binary}. Specifically, the input is a solution that points to an incorrect answer. We set a per-step sampling hyperparameter $N_{\text{mid}}$. Initially, we set the start and end points of the search range to Step 1 and Step N, respectively. We first consider the $mc$ value of Step (1+N)//2. If it is positive, it indicates that the first totally erroneous step occurs in the latter half; otherwise, we look in the first half. This reduces the number of searches to $\mathcal O(\log N)$.
\begin{algorithm}[!t]   
   \caption{BinaryErrorLocating}
   \label{alg:binary}
\begin{algorithmic}[1]     
   \State {\bfseries Input:} Solution set $S=\{y_i|i=1,2,...,N\}$, mid step sampling num $N_{mid}$.
   \For{$i=1$ {\bfseries to} $N$}
       \State $l \leftarrow 0$, $r \leftarrow \text{StepLen}(y_i)$ \Comment{Define start and end index.}
       \While{$l < r$} 
           \State $mid \leftarrow \lfloor (l + r) / 2 \rfloor$ 
           \State MID\_RES $\leftarrow {\mathcal{F}}(\{y_{i,1}, y_{i,2}, ..., y_{i,mid}\}, N_{mid})$ \Comment{Sampling from mid step.}
           \If{\Call{Verify}{MID\_RES} does not contain True} \Comment{Judge sampling's final answer.}
               \State $r \leftarrow mid$
           \Else
               \State $l \leftarrow mid + 1$
           \EndIf
       \EndWhile
       \State $e_i \leftarrow \Call{WrongStepLabeling}{y_i, l}$ \Comment{Collect sequential labels of $y_i$.}
       \State ErrorLocatingSet.append\{$e_i$\} 
   \EndFor
   \State {\bfseries Output:} ErrorLocatingSet
\end{algorithmic}
\end{algorithm}

Besides, we introduce the process of PS-GRPO in Algorithm~\ref{alg:ps_grpo}. This process involves merging of outcome reward and process-as-outcome reward, and subsequent relative advantages calculation. 
\begin{algorithm}[!t]   
   \caption{PS-GRPO}
   \label{alg:ps_grpo}
\begin{algorithmic}[1]     
   \State {\bfseries Input:} Policy model $\pi_{\theta}$, Process Reward Model $\phi_{\theta}$, train dataset $D_{RL}=\{I_i,Q_i,Y_i\}_{i=1}^N$.
   \State ErrorLocatingSet $\leftarrow$ []
   \For{$\text{Epoch}=1\ to\ N$}
   \For{$\text{Batched data}\ \{I,Q,Y\}\ \ in\ D_{RL}$}
        \State Generate rollouts of $\{I, Q\}$: $\{c^j\}_{j=1}^M$$\sim \pi_{\theta}$
        \State Calculate process reward sets $S=\{(p_1,p_2,\cdots, p_{|c_j|})\sim \phi_{\theta}(c^j)|0 < j < M\}$
        \State Calculate reward $\{r^j\}_{i=1}^M$ by Equation~\ref{equ:drop_moment} and~\ref{equ:reward_modeling}.
        \State Calculate relative advantages $\{\hat A^j\}_{i=1}^M$ using Equation~\ref{equ:adv}.
        \State Update policy model $\pi_{\theta}$ using Equation~\ref{eq:grpo_objective_superscript}.
   \EndFor
   \EndFor
   \State {\bfseries Output:} Updated policy model $\pi_{\theta}$.
\end{algorithmic}
\end{algorithm}

\section{Prompt Design}\label{app:prompt_design}

\subsection{Prompt Utilized in MMathCoT-1M Synthesis}

In this section, we provide the specific prompts for three-module data synthesis. Additionally, Gemini-1.5-Flash is a model that is very sensitive to prompts and parameters in practical experience, and we will share detailed adjustment experiences.

\begin{figure}[!t]
    \centering
    \includegraphics[width=1.0\linewidth]{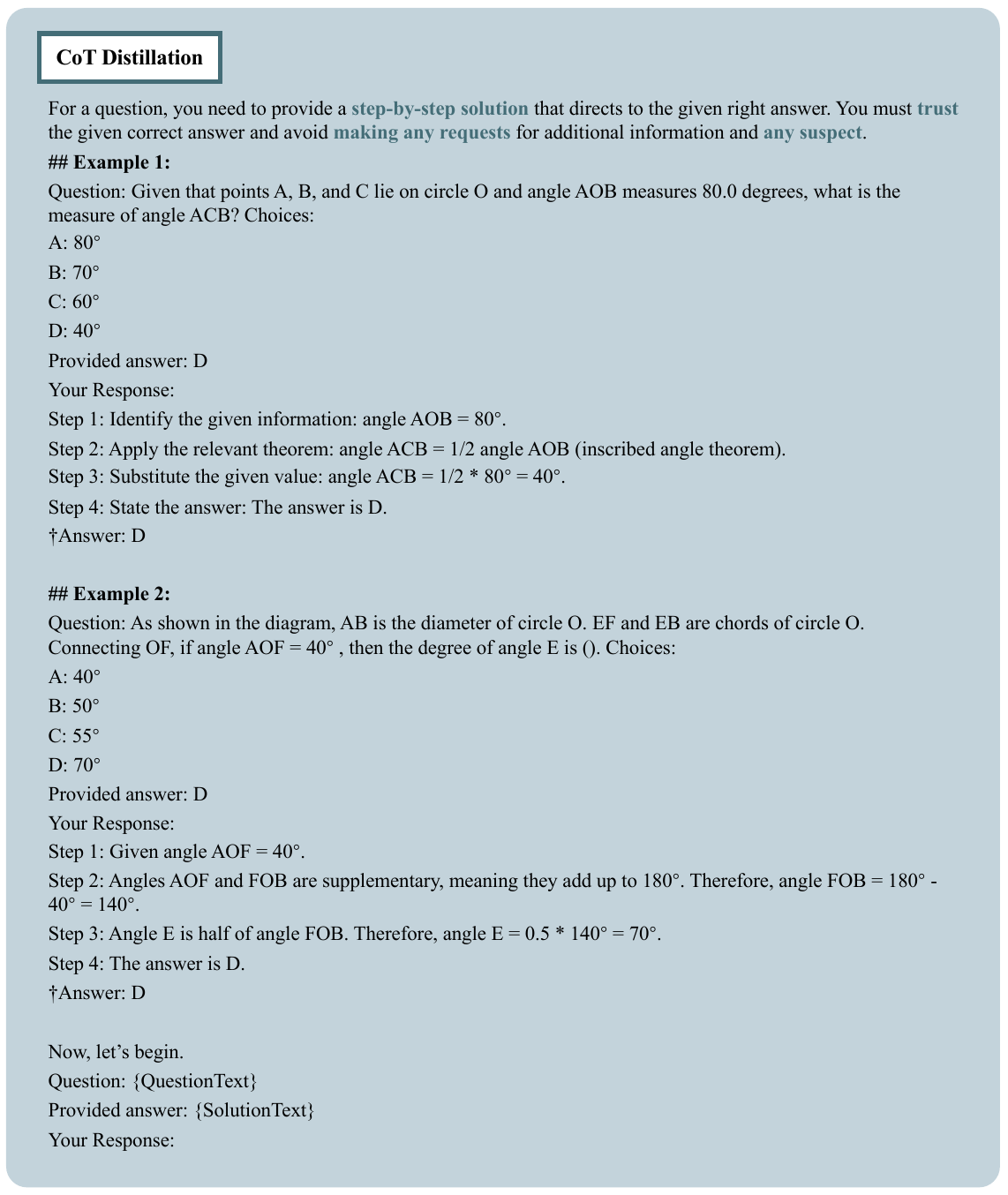}
    \vskip -0.1in
    \caption{CoT expansion prompt for answer-only data.}
    \label{fig:prompt_cot_dis}
\end{figure}

\paragraph{CoT Expansion} CoT expansion prompt for answer-only data source can be seen in Figure~\ref{fig:prompt_cot_dis}. We order the Gemini-1.5-flash to give a reasonable process directing to the ground-truth. After the execution, we find that the outputs is not so clear. The model sometimes will give trajectories that include \enquote{we must trust the answer} or \enquote{let me assume}. We identify these phrases as signals that the model can not solve the problem naturally and independently. We will filter these samples.

\begin{figure}[!t]
    \centering
    \includegraphics[width=1.0\linewidth]{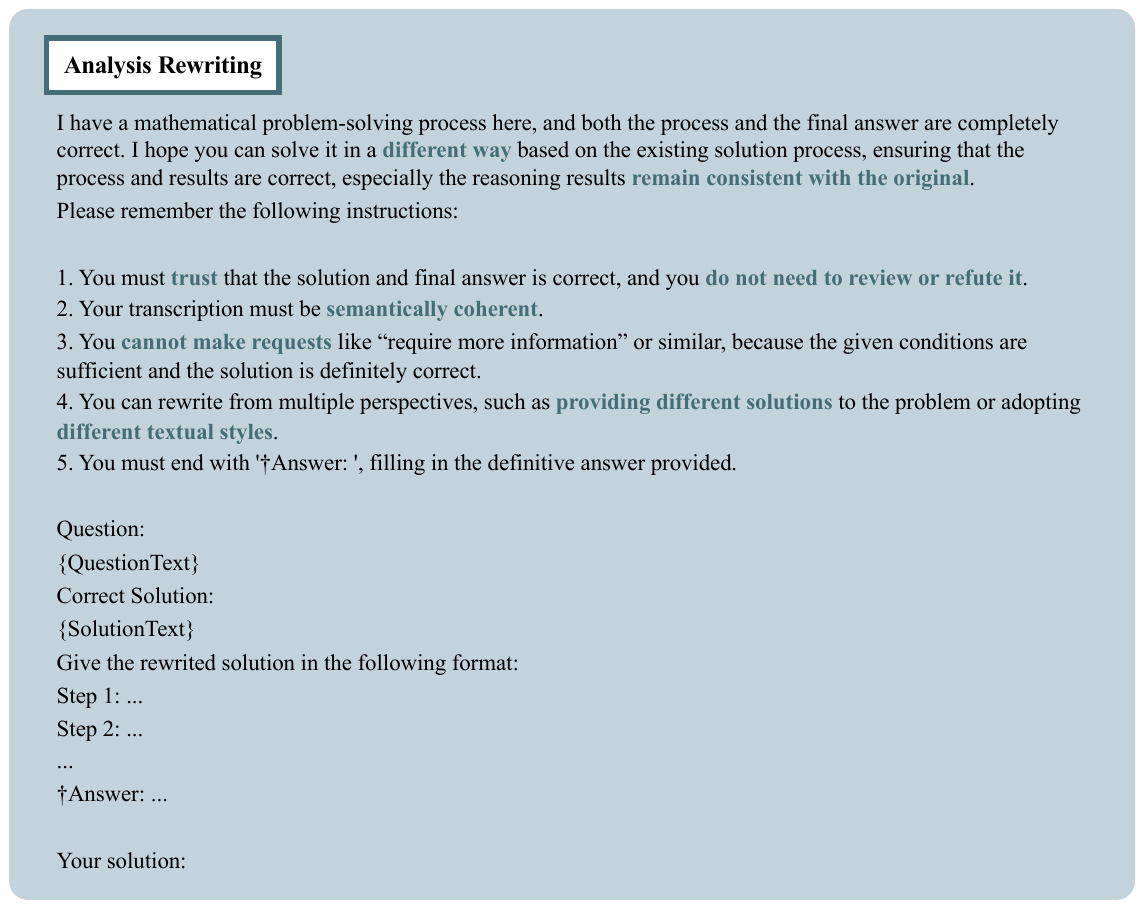}
    \vskip -0.1in
    \caption{Analysis rewriting prompt for analysis-formatted data.}
    \label{fig:prompt_rewriting}
\end{figure}

\paragraph{Analysis Rewriting} Rewriting prompt for analysis-formatted data synthesis is illustrated in Figure~\ref{fig:prompt_rewriting}. For solutions in an analytical format, we transform them into clear step-by-step format trajectories. During this process, Gemini-1.5-flash-002 does not exhibit significant questioning or make conditional requests. We improve data quality through reorganization and polishing of the language logic.

\begin{figure}[!t]
    \centering
    \includegraphics[width=1.0\linewidth]{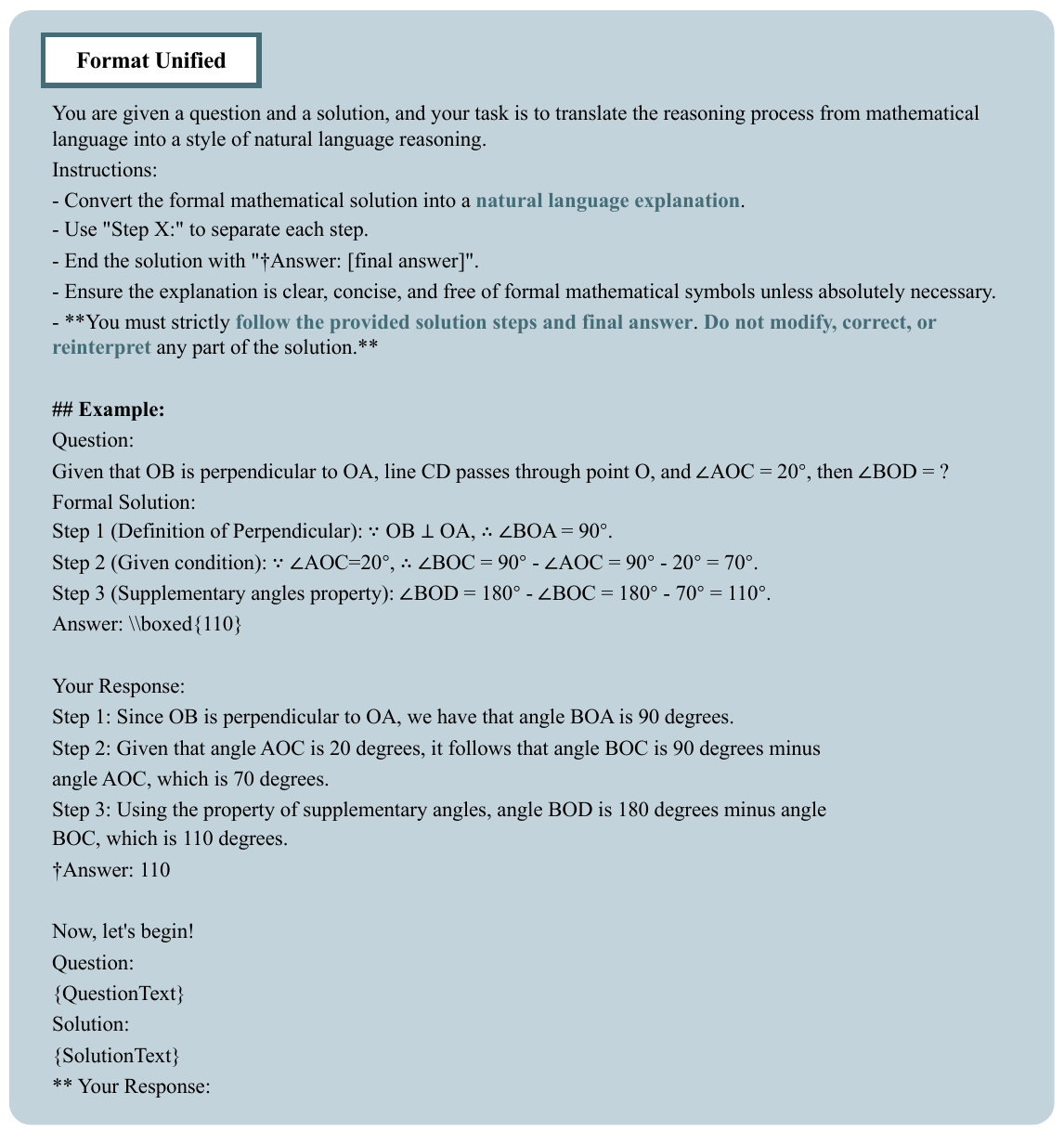}
    \vskip -0.1in
    \caption{Format unify prompt for mathematical and symbolic reasoning style data.}
    \label{fig:prompt_unify}
\end{figure}

\paragraph{Format Unified} By employing a unified format prompt shown in Figure~\ref{fig:prompt_unify} to modify the reasoning styles of plan-and-reasoning and symbolic approaches, we are able to extract a more natural language process aligned with the pre-training style. A single example is sufficient to elicit perceptually favorable responses.

\begin{figure}[!t]
    \centering
    \includegraphics[width=1.0\linewidth]{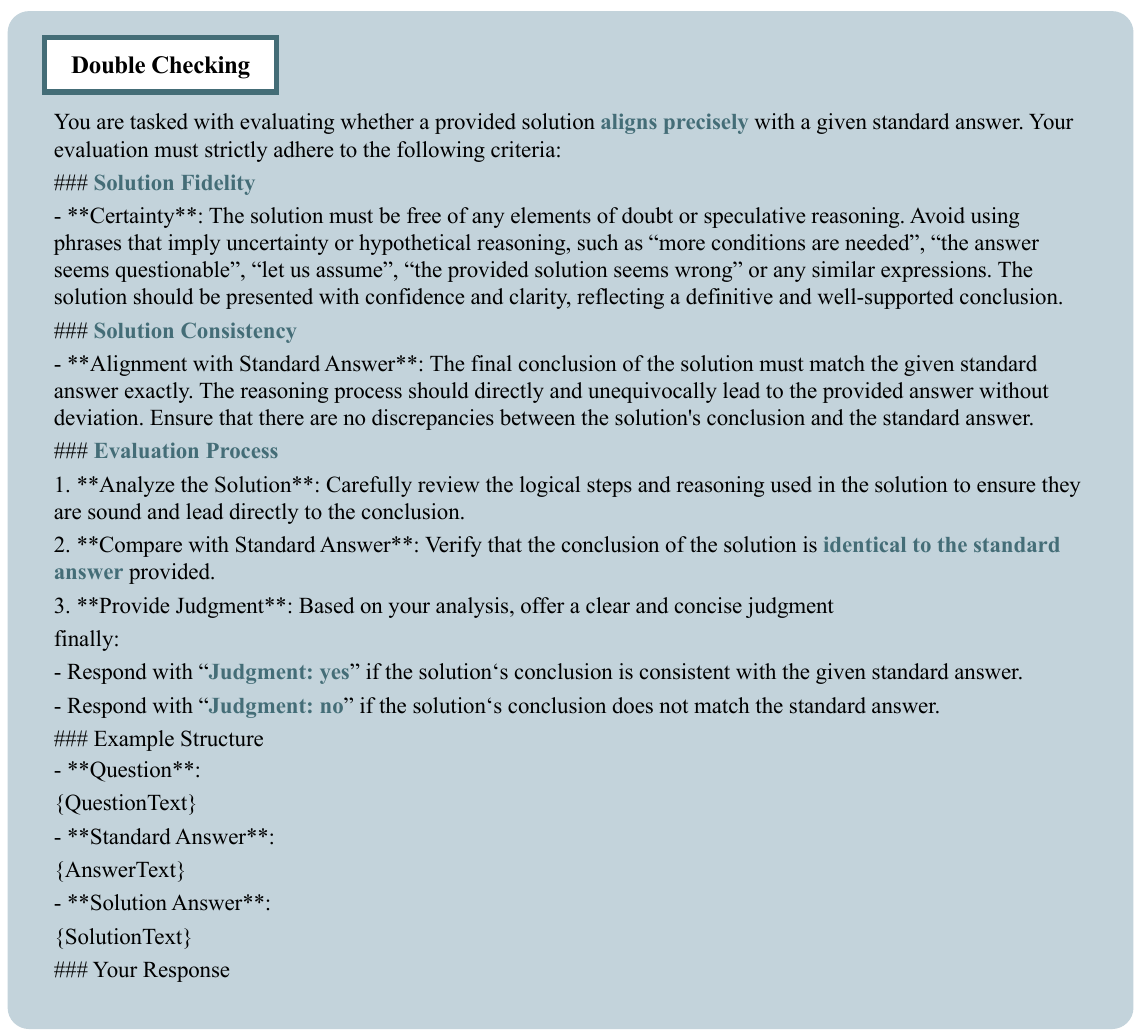}
    \vskip -0.1in
    \caption{Double-checking prompt for ensuring high-quality and appropriate trajectories in synthesized CoT reasoning data.}
    \label{fig:double_check}
\end{figure}

\paragraph{Double-checking} After completing the above three points, we apply an LLM-as-a-judge for double-checking the synthesized data, ensuring that the solutions do not contain unreasonable processes, such as untimely questioning, conditional requests, or reasoning loops. The specific prompt design is shown in Figure~\ref{fig:double_check}. After this layer of filtering, we obtain the final MMathCoT-1M.

\subsection{Prompt Utilized in DualMath-1.1M Synthesis}
\begin{figure}[!t]
    \centering
    \includegraphics[width=1.0\linewidth]{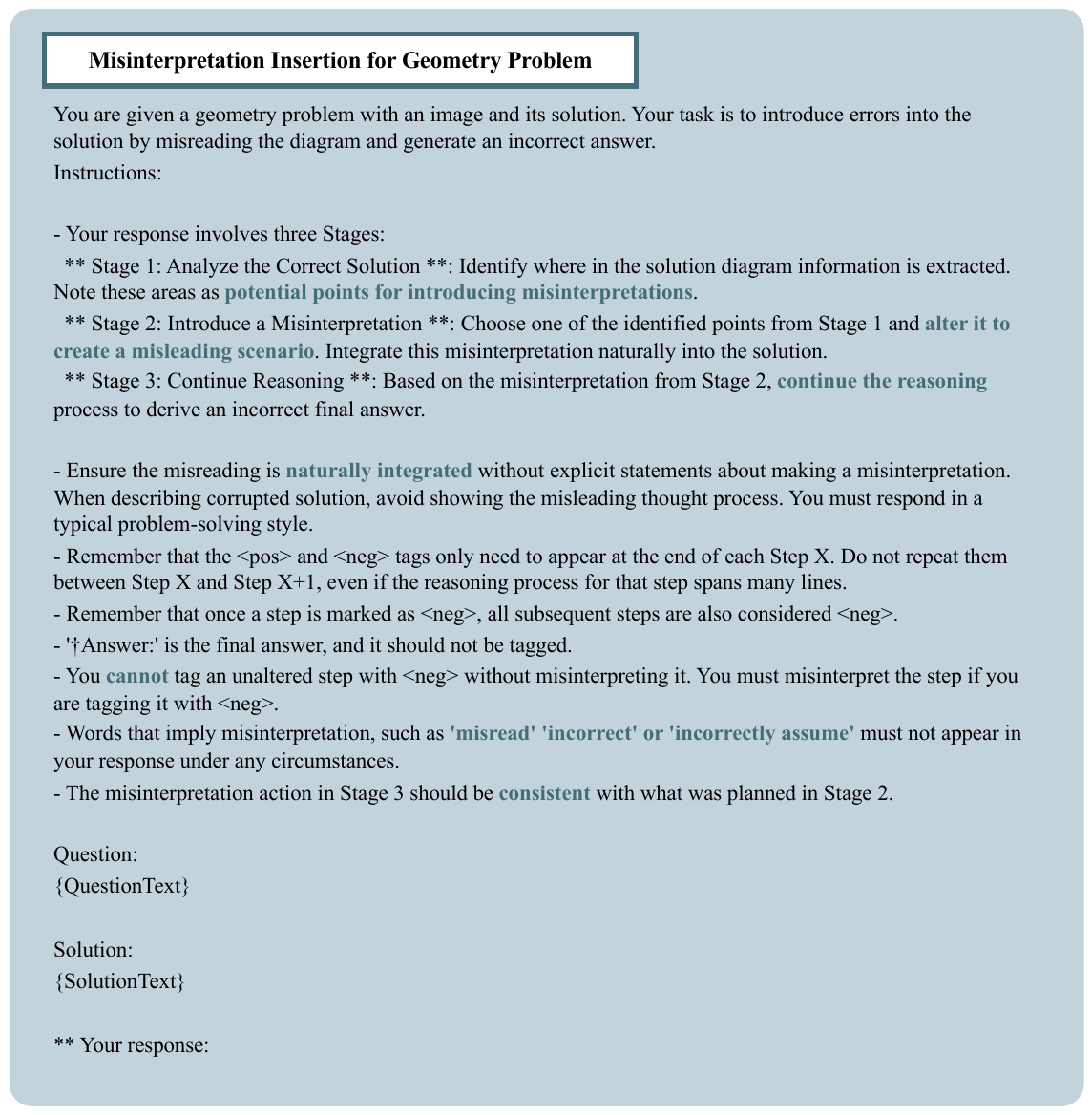}
    \vskip -0.1in
    \caption{Misinterpretation insertion for geometry-related problems.}
    \label{fig:mie}
\end{figure}

\begin{figure}[!t]
    \centering
    \includegraphics[width=1.0\linewidth]{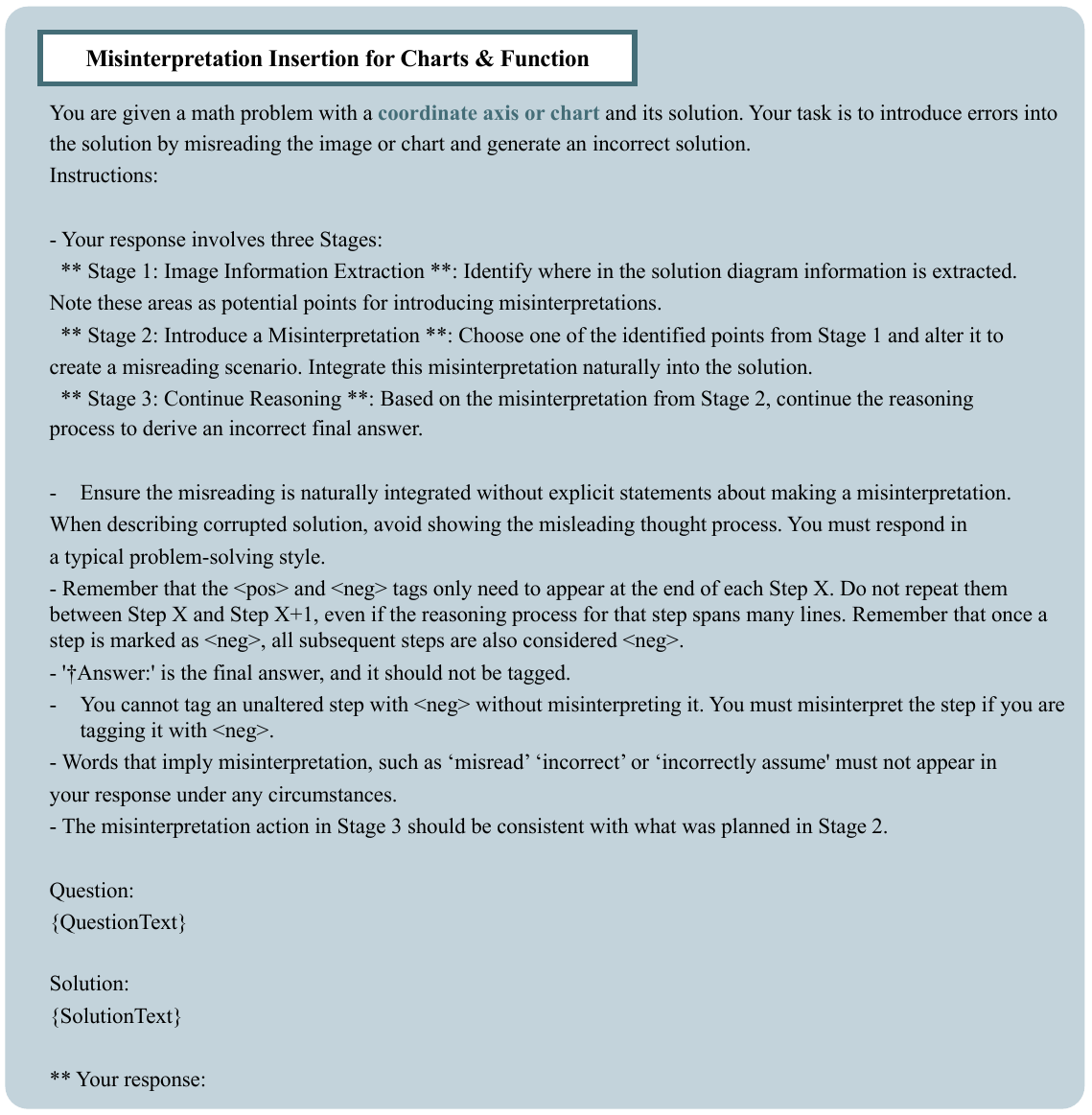}
    \vskip -0.1in
    \caption{Misinterpretation insertion for function and chart-related problems.}
    \label{fig:mie_1}
\end{figure}

In this section, we demonstrate the prompts used in MIE. 
\begin{itemize}[left=1em] 
    \item \textbf{Geometry Problem}: For geometry problem, we prompt the Gemini-1.5-Flash-002 to first identify key geometry features in the figure. We then order it to introduce a misinterpretation on these elements. Finally, use the wrong information to execute a misleading solution. The total design can be seen in Figure~\ref{fig:mie}.
    \item \textbf{Charts \& Function}: For ChartQA and math functions, we prompt Gemini-1.5-Flash-002 to first check the fine-grained data points. We then attempt to insert spatially similar data to induce a misinterpretation. This subsequently leads to incorrect solutions for automatic labeling.
    \item \textbf{LLM-as-a-Judge}: For chart reasoning and function problem, we execute similar process on Gemini-1.5-Flash-002. We place it in Figure~\ref{fig:mie_1}.
\end{itemize}

\section{Case Study}
\subsection{Showcase on Best-of-N Evaluation}

\begin{figure}[!t]
    \centering
    \includegraphics[width=1.0\linewidth]{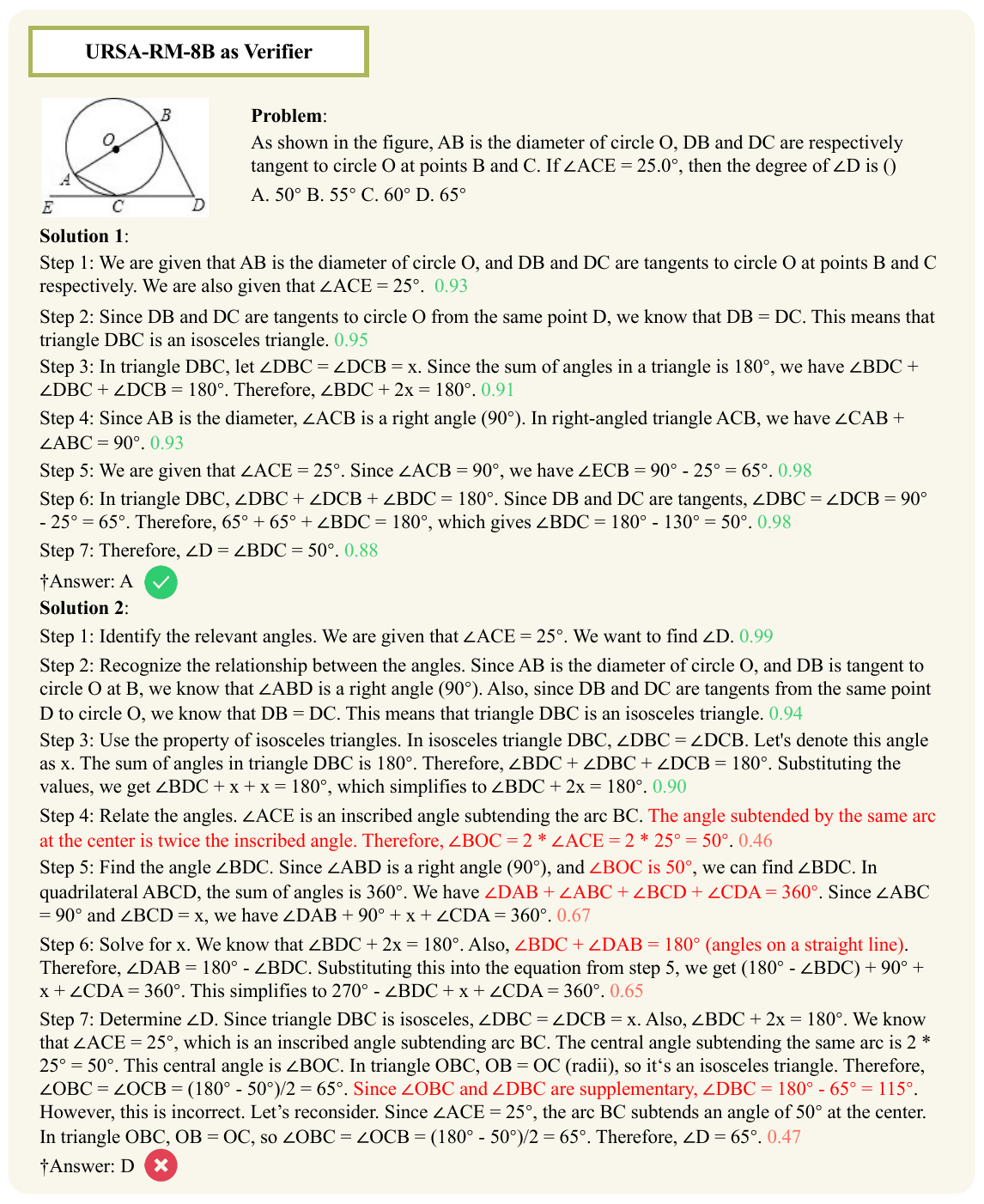}
    \caption{Case of URSA-8B-RM serving as a verifier in Best-of-N evaluation.}
    \label{fig:verifier_case}
\end{figure}

To more clearly illustrate the effectiveness of URSA-8B-RM in BoN evaluation, a case on MathVista-GPS is demonstrated (Figure~\ref{fig:verifier_case}). This case shows that URSA-8B-RM is sensitive to false theorem application and misunderstandings of angle-number relations. The good property not only enables URSA-8B-RM to perform well in BoN evaluation but also endows it with the potential to identify more valuable learning samples in online reinforcement learning.

\subsection{Process of Misinterpretation Insertion Engine}\label{app:sec:mie_case}
\begin{figure}[!t]
    \centering
    \includegraphics[width=1.0\linewidth]{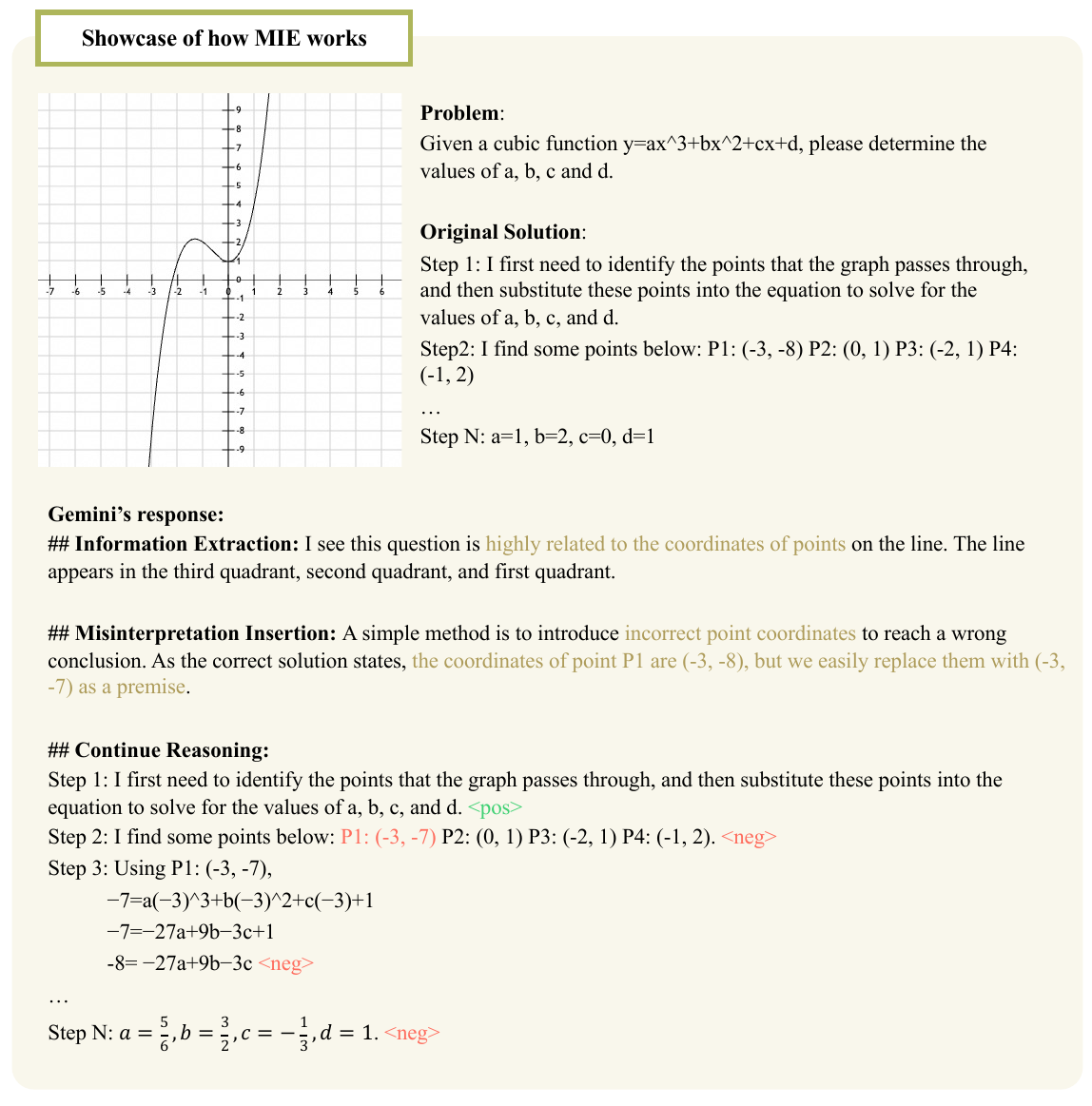}
    \caption{Case from MIE. We introduce specific step-level perception errors and continue false reasoning to construct a correctnesses-labeled solution.}
    \label{fig:mie_case}
\end{figure}

As shown in Figure~\ref{fig:mie_case}, MIE performs three main actions: First, it interprets the mathematical information in the image. Then, it replaces key information at a selected step. Finally, it continues reasoning based on the modified conditions.
\subsection{Failure Pattern in Process During GRPO}
\begin{figure}[!t]
    \centering
    \includegraphics[width=1.0\linewidth]{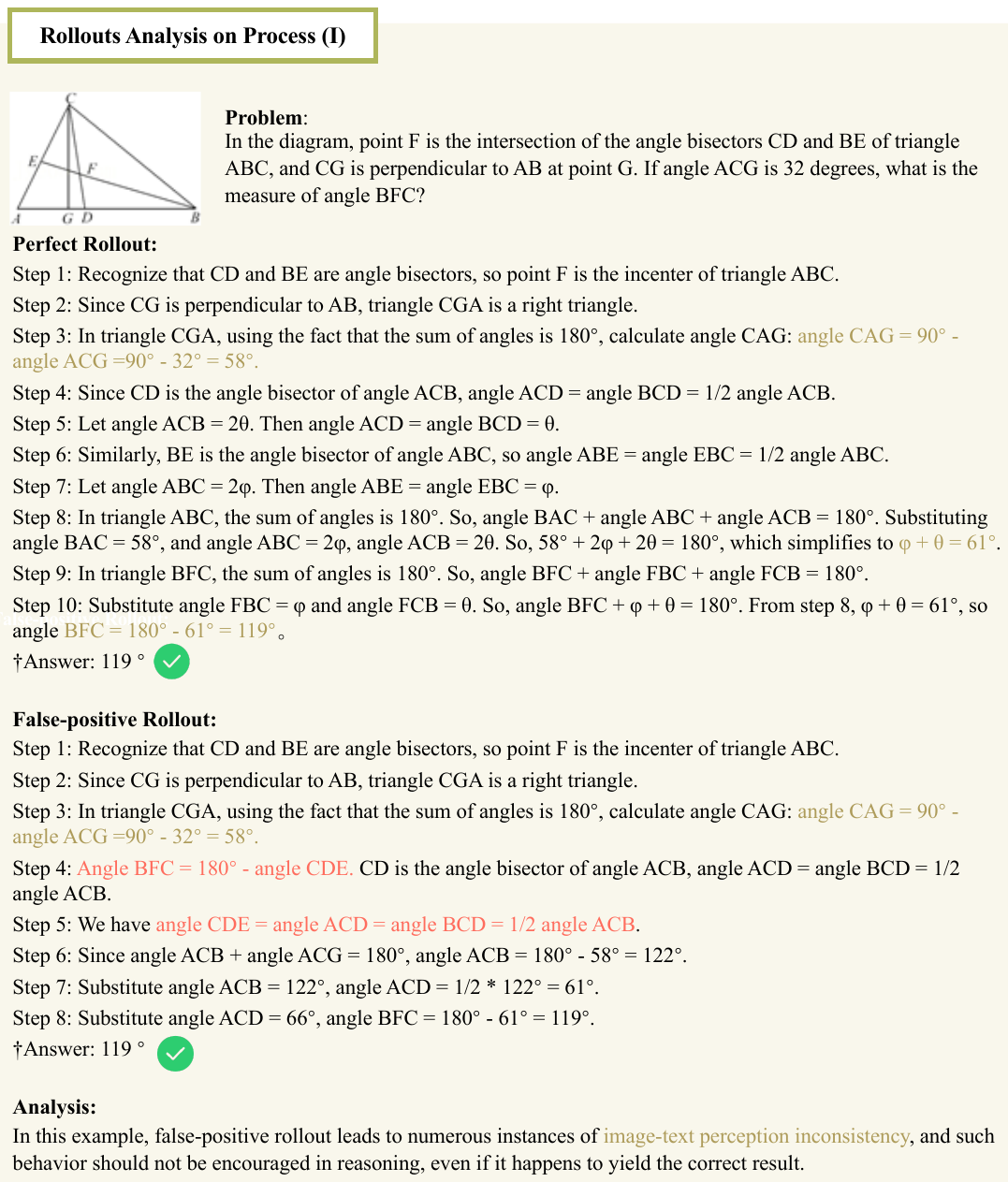}
    \caption{False positive rollout analysis \RN{1}.}
    \label{fig:rollout_case_1}
\end{figure}

\begin{figure}[!t]
    \centering
    \includegraphics[width=1.0\linewidth]{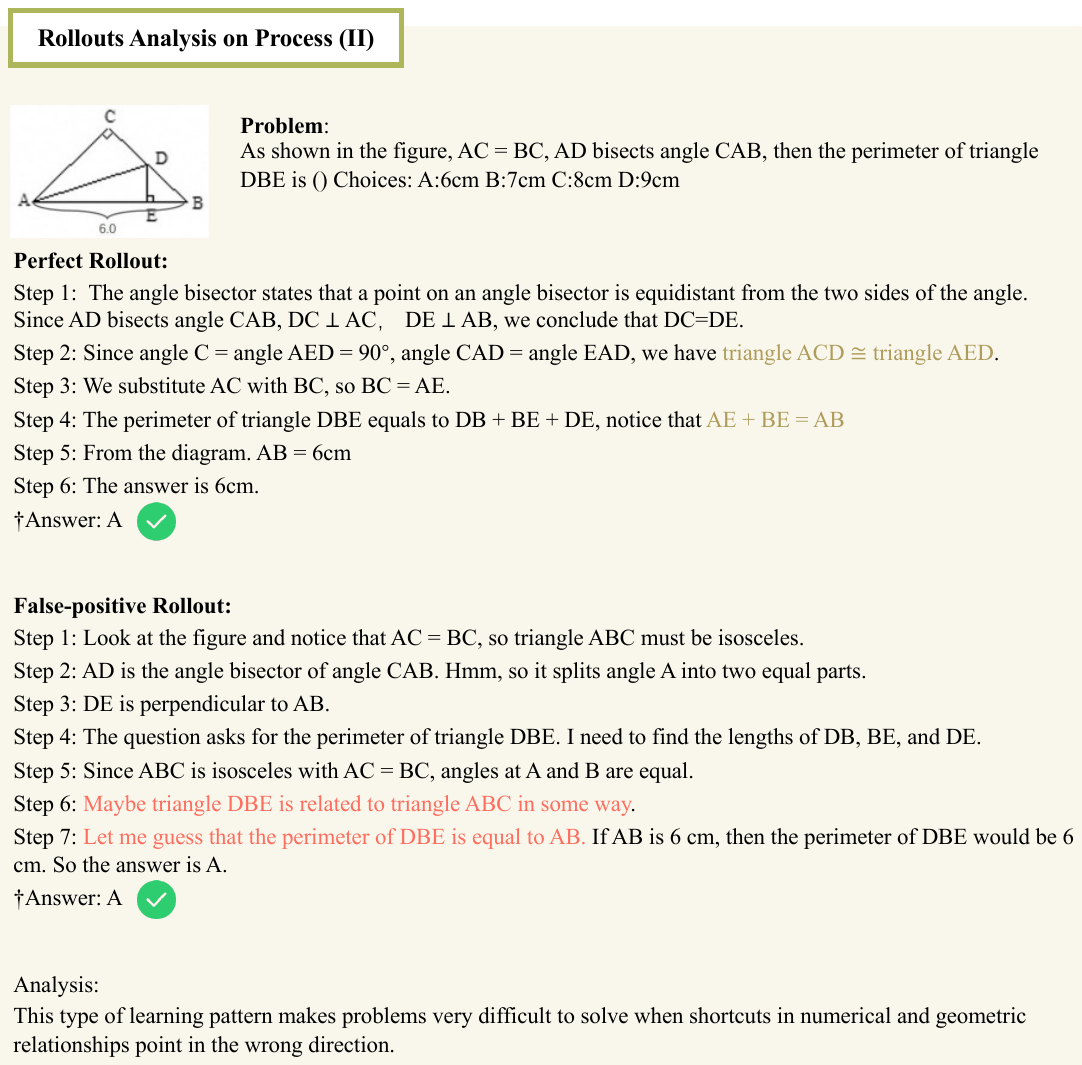}
    \caption{False positive rollout analysis \RN{2}.}
    \label{fig:rollout_case_2}
\end{figure}

In this section, we intuitively reveal through examples why PS-GRPO works effectively. We first introduce the concept of false-positive rollouts, which are rollouts that, despite reaching the correct answer, do not provide perfect intermediate actions to arrive at the solution. They can generally be divided into two categories: (\romannumeral1) the lack of visual condition alignment. Solutions in this category exhibit inconsistencies in reasoning regarding basic visual factors such as edge relationships, coordinate values, and theorem applications, revealing deficiencies in the pretraining phase, as shown in Figure~\ref{fig:rollout_case_1}. (\romannumeral2) the exploitation of shortcut patterns. These rollouts do not go through key steps but are guided directly to the correct answer after basic descriptions due to the high correlation between image features and problem-solving patterns during pretraining or SFT, as shown in Figure~\ref{fig:rollout_case_2}.
Therefore, PS-GRPO suppresses the advantageous direction brought by these erroneous actions through the sensitivity of the PRM in online RL for error identification. This leads to a more optimal paradigm that combines outcome rewards with process reward-based penalties.
\subsection{Cases on How Naive Process Reward Modeling Fails}
\begin{figure}[!t]
    \centering
    \includegraphics[width=1.0\linewidth]{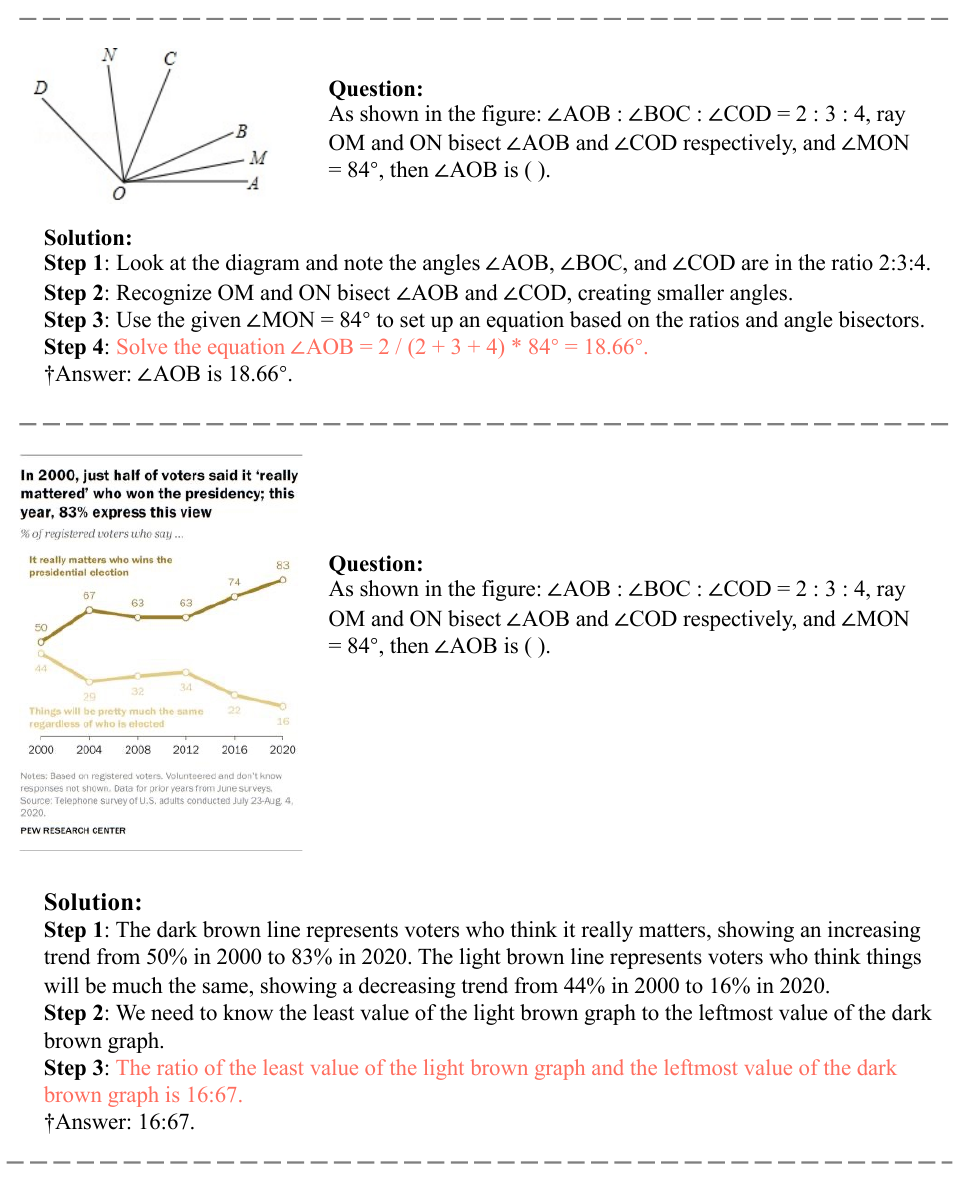}
    \caption{Bad case analysis on two process reward modeling variants.}
    \label{fig:bad_case}
\end{figure}

In this section, we elaborate on the two fundamental flaws of process reward guided RL mentioned in Section~\ref{sec:stage_3} and present some cases for illustration. In online RL, models can easily recognize the patterns for obtaining process rewards, leading to conservative analyses and concise responses as they sidestep PRM scrutiny.
As shown in Figure~\ref{fig:bad_case}, we have observed that the model often follows a distinct reasoning pattern. They initially read and analyze the given conditions comprehensively, but then make incorrect decisions based on this analysis, leading to wrong answers. This indicates that when explicitly modeling process rewards, models can easily focus on processes that seem "correct" in isolation. However, these processes may not be genuinely helpful for the final outcome and instead may lead the model to prioritize high process rewards over accuracy.

\end{document}